\documentclass[10pt,twocolumn,letterpaper]{article}

\usepackage{cvpr}              % To produce the CAMERA-READY version
\definecolor{cvprblue}{rgb}{0.21,0.49,0.74}
\usepackage[pagebackref,breaklinks,colorlinks,allcolors=cvprblue]{hyperref}
\usepackage[accsupp]{axessibility}
\usepackage{bm}
\usepackage{booktabs} 
\usepackage{multirow}
\usepackage{array}
\usepackage{colortbl}
\usepackage{dashrule}
\usepackage{arydshln}
\usepackage{capt-of} 
\usepackage{amsmath}
\usepackage{subcaption}
\usepackage{amsfonts}

\def\paperID{1472} % *** Enter the Paper ID here
\def\confName{CVPR}
\def\confYear{2026}

\title{SGAD-SLAM: Splatting Gaussians at Adjusted Depth for Better Radiance Fields in RGBD SLAM}

\author{Pengchong Hu ~~~~~ Zhizhong Han\\
Machine Perception Lab, Wayne State University, Detroit, USA\\
{\tt\small pchu@wayne.edu, h312h@wayne.edu}
}

\begin{document}
\maketitle

\begin{abstract}
3D Gaussian Splatting (3DGS) has made remarkable progress in RGBD SLAM. Current methods usually use 3D Gaussians or view-tied 3D Gaussians to represent radiance fields in tracking and mapping. However, these Gaussians are either too flexible or too limited in movements, resulting in slow convergence or limited rendering quality. To resolve this issue, we adopt pixel-aligned Gaussians but allow each Gaussian to adjust its position along its ray to maximize the rendering quality, even if Gaussians are simplified to improve system scalability. To speed up the tracking, we model the depth distribution around each pixel as a Gaussian distribution, and then use these distributions to align each frame to the 3D scene quickly. We report our evaluations on widely used benchmarks, justify our designs, and show advantages over the latest methods in view rendering, camera tracking, runtime, and storage complexity. Please see our project page for code and videos at \href{https://machineperceptionlab.github.io/SGAD-SLAM-Project}{https://machineperceptionlab.github.io/SGAD-SLAM-Project}.
\end{abstract}

\section{Introduction}
RGBD SLAM jointly estimates camera poses and geometry from an RGBD image sequence. It has been widely used in robotics, AR, and VR~\cite{keetha2024splatam,Zhu2021NICESLAM,wang2023coslam,Hu2023LNI-ADFP,9712211robot}. Traditional methods employ discrete 3D points to represent the geometry of scenes; however, these discrete representations do not represent continuous surfaces well or support novel view synthesis
More recent methods~\cite{Zhu2021NICESLAM,wang2023coslam,Hu2023LNI-ADFP,Sandström2023ICCVpointslam} employ continuous radiance fields to represent both the geometry and the appearance of scenes. They usually learn a representation called NeRF~\cite{mildenhall2020nerf}, a radiance field parameterized by a neural network, during mapping, and estimate camera poses during tracking, both of which are achieved by minimizing the rendering errors against the observed images. Although NeRF has proven to be an good representation in SLAM, the ray tracing-based rendering is very slow, especially in iterative optimization of mapping and tracking on each frame. This raises rendering efficiency as a challenge in rendering-based SLAM solutions.

3D Gaussian Splatting (3DGS)~\cite{kerbl3Dgaussians} has emerged as a promising alternative to overcome this challenge. By representing a radiance field using a set of explicit 3D Gaussian functions with attributes, 3DGS can render these Gaussians into images through a differentiable splatting operation, which significantly improves the rendering efficiency. With 3DGS, the latest SLAM methods~\cite{keetha2024splatam,ha2024rgbdgsicpslam,yugay2023gaussianslam,MatsukiCVPR2024_monogs,yan2023gs,zhu2024_loopsplat} learn 3D Gaussians and estimate camera poses by minimizing rendering errors against the observed images. Some of these methods allow Gaussians to move across the whole scene, but it is expensive to hold all Gaussians due to the limitation of the GPU memory, making it hard to scale up to large scenes. In contrast, some other methods~\cite{Hu2025VTGSSLAM} employ view-tied Gaussians, which are strictly anchored to fixed depth points; however, the strict constraint leads to a negative impact on rendering novel views. Therefore, how to represent a better radiance field for more accurate tracking and mapping in SLAM is still a challenge.

\begin{figure*}[t]
  \centering
% \vspace{-0.4in}
   \includegraphics[width=\linewidth]{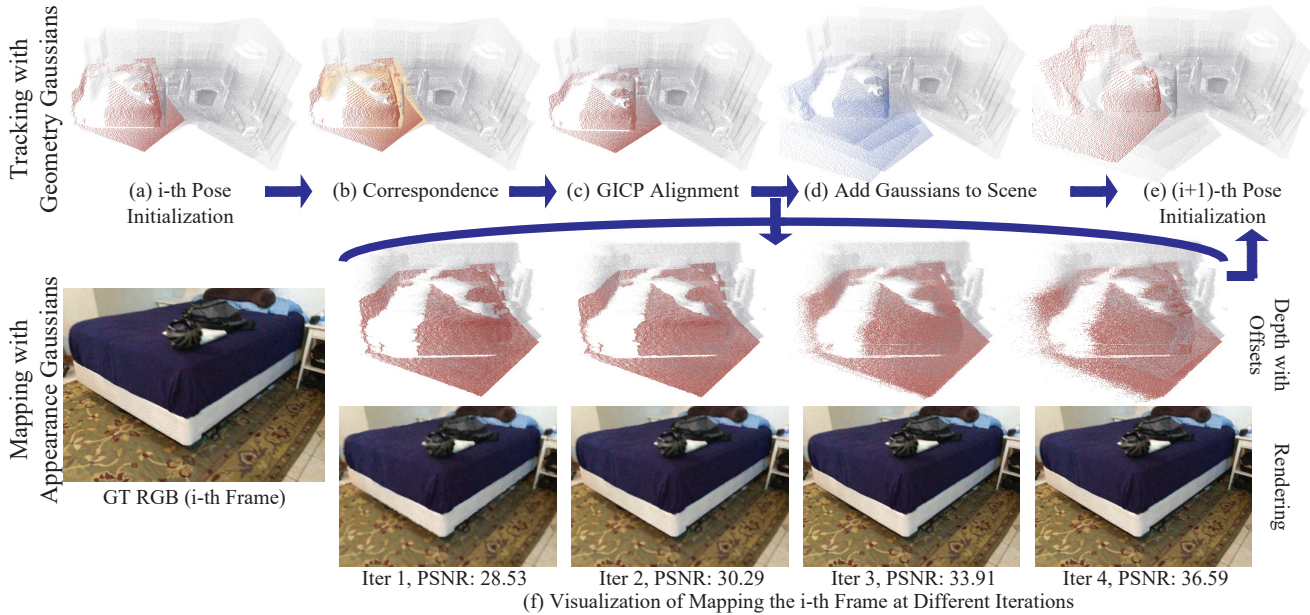}
   \vspace{-0.25in}

   \caption{Overview of our method. Our tracking strategy, illustrated in (a)-(e), employs geometry Gaussians to represent the scene structure, achieving robustness and efficiency. During mapping, we learn appearance Gaussians for rendering, where the depth offset at each pixel is progressively adjusted for better rendering, improving scalability and efficiency. We visualize the Gaussian centers for better readability.}
   \label{fig:overview}

\vspace{-0.1in}
\end{figure*}

To overcome these challenges, we propose an RGBD SLAM method with 3DGS based on better modeling of radiance fields, aiming for more scalable and efficient SLAM systems in large scenes. Our key idea is based on pixel-aligned Gaussians, but allows Gaussians to move along their rays, which not only improves the scalability but also fits better radiance fields, leading to more accurate tracking and mapping. Specifically, we associate a Gaussian with each pixel on each frame, and make Gaussians focus more on rendering the specific frame and its neighboring frames. This design allows us to merely maintain and optimize Gaussians associated with a few frames, but not all Gaussians in the scene, significantly improving the scalability of the SLAM system without a need to hold all Gaussians during the training. Moreover, we employ a simplified Gaussian to represent the scene, which also saves storage space and makes our method more memory efficient. To maximize the rendering quality that may be impacted by the simplified Gaussian modeling and the constraints on their movement and densification, we allow Gaussians to adjust their positions along their rays when learning Gaussian attributes during mapping. Furthermore, to speed up tracking, we introduce a novel method to estimate camera poses by aligning pixels on each frame to the scene in terms of geometry similarity, where we model the geometry around each pixel on a frame as a Gaussian distribution, aiming to approximate the local geometry around each pixel. We justify the effectiveness of each module and report evaluations through numerical and visual comparisons with the latest methods. Our main contributions are listed below.
% are listed below.

\begin{itemize}
\item We propose using pixel-aligned Gaussians at adjusted depth for better modeling of radiance fields in SLAM, improving the capability of mapping in large scenes and the rendering quality.
\item We introduce a novel tracking strategy based on geometry similarity in 3D, coupled with a rendering-based initialization, to significantly improve tracking efficiency.
\item We report state-of-the-art performance in tracking and mapping on the widely used benchmarks and show advantages over the latest 3DGS-based SLAM methods.
\end{itemize}

\section{Related Work}
\noindent\textbf{Multi-view Reconstruction. }Recently, due to the promising results in multi-view reconstruction~\cite{park2021nerfies,mueller2022instant,ruckert2021adop,yu_and_fridovichkeil2021plenoxels,wang2022go-surf,bozic2021transformerfusion,DBLP:journals/corr/abs-2209-15153,sun2021neucon,li2023rico,wang2023vggsfm,mast3rsfm,bai2025learningcompactlatentspace,udfstudio,zhang2026vrpudfunbiasedlearningunsigned,zhou2025ucanunsupervisedpointcloud,han2025sparsereconneuralimplicitsurface,noda2025learningbijectivesurfaceparameterization,zhang2025nerfpriorlearningneuralradiance,zhang2025monoinstanceenhancingmonocularpriors,jiang2024sensingsurfacepatchesvolume,sijia2023quantized,chen2024learninglocalpatternmodularization,zhang2024learning,chen2024sharpeningneuralimplicitfunctions,NeuralTPS,zhou2024zeroshotscenereconstructionsingle,chen2024inferringneuralsigneddistance,noda2024multipulldetailingsigneddistance} with neural implicit representations, many methods focus on incorporating more information beyond RGB images, such as depth~\cite{Yu2022MonoSDF,NeuralRGBDSurfaceReconstruction_2022_CVPR} and normals~\cite{wang2022neuris,guo2022manhattan,patel2024normalguideddetailpreservingneuralimplicit}, into the reconstruction pipeline as priors or supervision to infer more accurate and detailed geometry. Meanwhile, 3D Gaussians~\cite{kerbl3Dgaussians,3dgrt2024} has emerged as a novel scene representation, which is also widely used in multi-view reconstruction~\cite{zhang2024gspull,Huang2DGS2024,Yu2024GOF,wolf2024gs2mesh,fan2024trim,gslrm2024,charatan23pixelsplat,lin2024directlearningmeshappearance,chen2024pgsrplanarbasedgaussiansplatting,Liu2026shortersplatting,zhang2025materialrefgsreflectivegaussiansplatting,li2025gaussianudfinferringunsigneddistance,han2024binocularguided3dgaussiansplatting}. However, all of these methods rely on accurate camera poses that are usually obtained by COLMAP~\cite{schoenberger2016mvs}, which is different from approaches based on SLAM techniques.

\noindent\textbf{Dense Visual SLAM.}
While multi-view stereo (MVS)~\cite{schoenberger2016sfm,schoenberger2016mvs} can estimate dense depth maps and camera poses from multiple RGB images by leveraging multi-view consistency, recent visual SLAM methods~\cite{zhang2023goslam,tofslam,teigen2023rgbd,sandström2023uncleslam,sucar2021imap,wang2023coslam,Sandström2023ICCVpointslam,cp-slam,Hu2023LNI-ADFP,jiang2024queryquantizedneuralslam} integrate continuous implicit representations with the classical SLAM pipeline to achieve more accurate mapping performance, particularly in novel view synthesis. Using RGBD images as rendering supervision, these methods can learn neural radiance fields to obtain a continuous implicit representation of the entire scene. Additionally, some approaches incorporate depth priors~\cite{Hu2023LNI-ADFP}, segmentation priors~\cite{kong2023vmap,haghighi2023neural}, object-level priors~\cite{kong2023vmap}, or large reconstruction model priors~\cite{murai2024_mast3rslam,maggio2025vggtslam,slam3r} to further improve the tracking and mapping performance.

Given the superior efficiency and rendering quality of 3DGS~\cite{kerbl3Dgaussians}, recent methods~\cite{keetha2024splatam,MatsukiCVPR2024_monogs,hhuang2024photoslam,yan2023gs,yugay2023gaussianslam,sandstrom2024splatslam,murai2024_mast3rslam,wei2024gsfusion,szymanowicz24splatter,zhang2024gaussiancube} have integrated it into SLAM pipelines for differentiable rendering. However, to ensure color and geometric consistency across all frames, these approaches optimize a global 3D Gaussian map that represents the entire scene, which needs to be maintained in GPU memory at all times. Thus, these methods struggle to scale to extremely large scenes. To address the limitations of the global 3D Gaussian map, we propose optimizing pixel-aligned 3D Gaussians to represent only a portion of the scene. This design focuses more on specific frames and their neighboring views, which eliminates the need to store the entire scene's Gaussians in memory during training. Additionally, the pixel-aligned 3D Gaussians are allowed to move along rays, further enhancing rendering performance.

To achieve more accurate camera poses, many SLAM methods~\cite{liso2024loopyslam,zhu2024_loopsplat,bruns2024neuralgraphmapping,sandstrom2024splatslam} integrate loop closure. However, detecting loop closures among views typically relies on pre-trained priors and is highly sensitive to image quality. In contrast, our method directly aligns each frame to a global distribution to maximize their geometric similarity, achieving more precise and faster camera pose estimation without the need for pre-trained priors.

\noindent\textbf{Gaussian Alignment. }
Some works~\cite{xu2024grm,gslrm2024,gao2024meshbasedgaussiansplattingrealtime,luiten2023dynamic3dgs,Hu2025VTGSSLAM,zakharov2024gh} have 
% also 
explored aligning 3D Gaussians to various entities. However, in these approaches, Gaussians are either independent of camera positions or associated with many attributes. Specifically, VTGS-SLAM~\cite{Hu2025VTGSSLAM} anchors Gaussians directly to pixels without allowing movement along viewing rays, which limits rendering performance on neighboring views due to the reduced degrees of freedom in adjusting positions. Instead, our pixel-aligned Gaussians directly link Gaussian positions to camera poses while employing simplified attributes, and flexibly adjust their positions along the ray, enhancing the efficiency and scalability of our SLAM system.

\section{Method}
\noindent\textbf{Overview.} Our method, SGAD-SLAM, consists of a rendering-based mapping process that splats pixel-aligned Gaussians at adjusted depth to map the scene and a geometry similarity based tracking strategy, as shown in Fig.~\ref{fig:overview}. Given an RGBD image sequence including $I$ frames $\{V_i,D_i\}_{i=1}^I$, we aim to learn a set of pixel-aligned Gaussians $\{G_i\}_{i=1}^I$ to represent the geometry and appearance of the scene and also estimate the camera poses $\{p_i\}_{i=1}^I$ for each frame in the sequence. Additionally, we maintain a 3D point set to represent the scene for faster tracking, leading to another branch parallel to the mapping branch in Fig.~\ref{fig:overview}.

\begin{figure} %{r}{0.55\linewidth}
  \centering
% \vspace{-0.2in}
   \includegraphics[width=\linewidth]{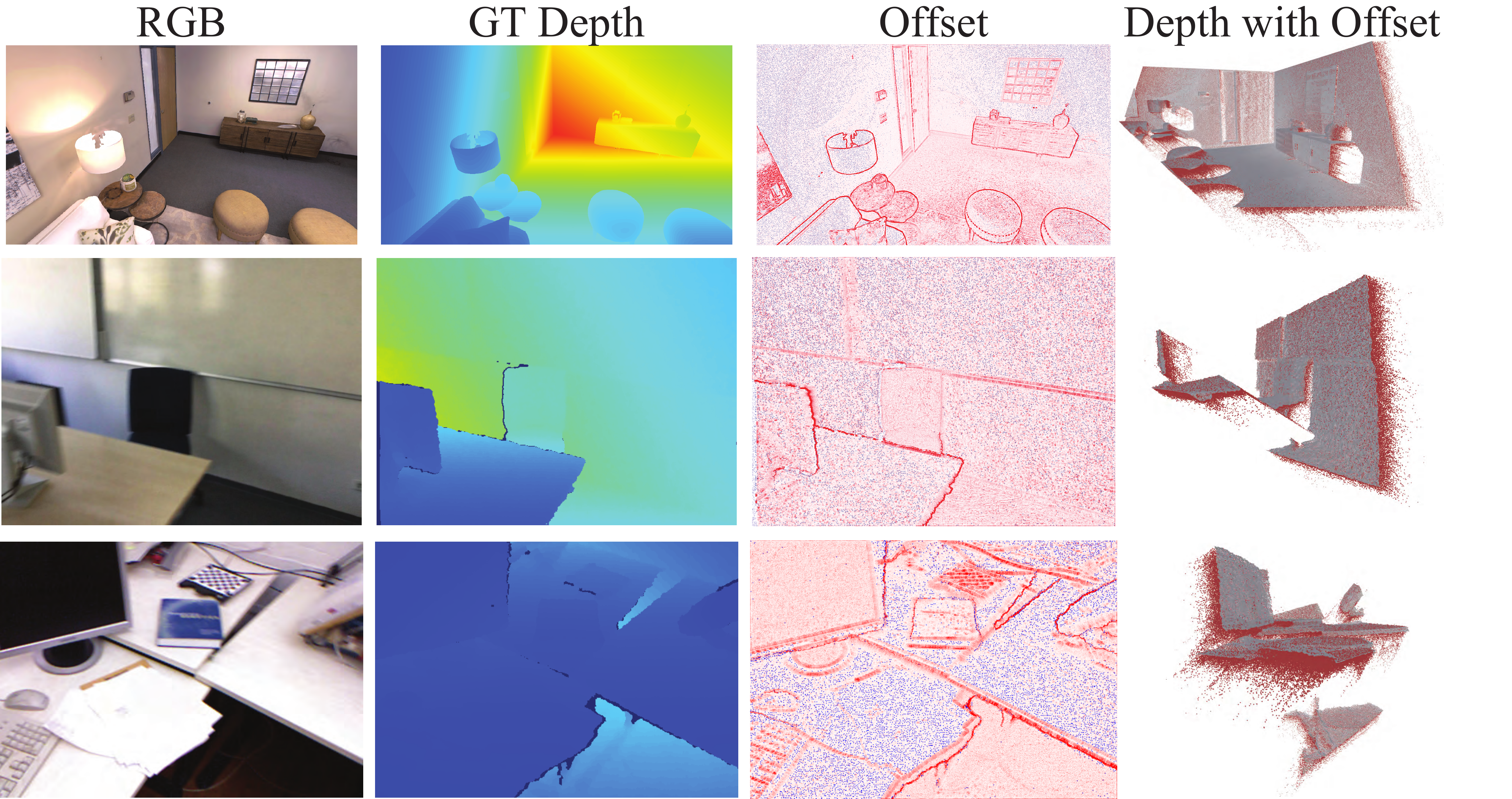}
   \vspace{-0.2in}
   \caption{Illustration of Offsets which place Gaussians either in front (blue) or behind (red) the depth points.}
   \label{fig:CameraEst}
\vspace{-0.05in}
\end{figure}
During mapping, we first initialize a set of 3D Gaussians $G_i=\{g_i^j\}_{j=1}^J$ on the $i$-th frame, and $g_i^j$ is associated to each pixel, where $i\in [1,I]$. We then learn the attributes of each Gaussian, including a depth offset, as illustrated in Fig.~\ref{fig:CameraEst}. The depth offset $\delta_i$ learned for the $i$-th frame indicates the position of the Gaussian along the ray that emits at the pixel from the camera center, as illustrated in Fig.~\ref{fig:Gaussians+Gaussianscale} (a).

During tracking, we optimize $p_i$ to maximize the geometry similarity between the local geometry distribution around pixels in the depth $D_i$ and a global geometry distribution representing the entire scene for better efficiency and robustness. Our tracking can start with a camera estimation by minimizing the rendering error using Gaussians on the previous frame, aiming for better robustness of tracking.

\begin{figure} %{r}{1.3\linewidth}
  \centering
% \vspace{-0.4in}        
   \includegraphics[width=\linewidth]{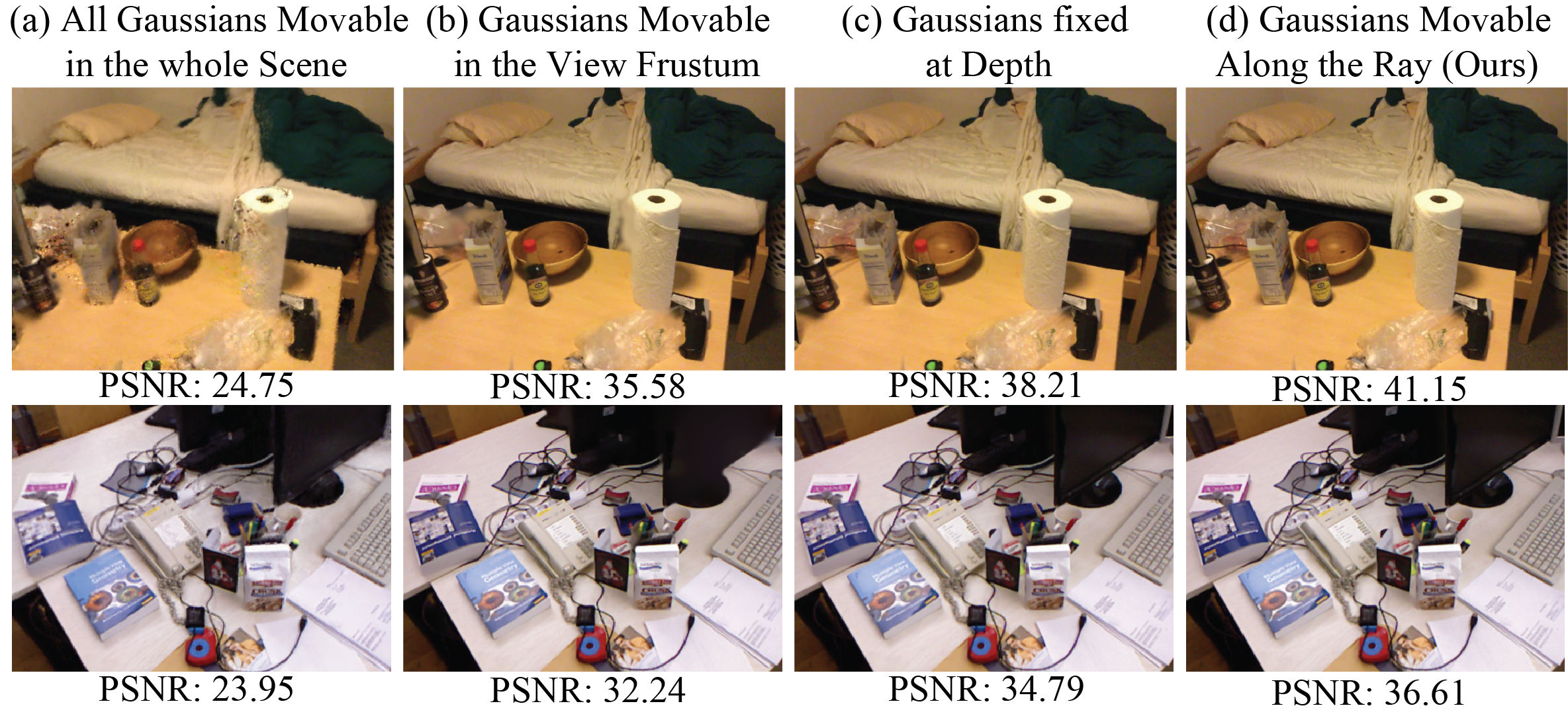}
   \vspace{-0.2in}
   \caption{Advantages of pixel-aligned Gaussians at adjusted depth over others in rendering comparison.}
   \label{fig:RenderComp}
\vspace{-0.15in}
\end{figure}

\subsection{Simplified Gaussian Representation}

To reduce the storage footprint of pixel-aligned Gaussians, we adopt a simplified spherical Gaussian representation with a depth offset ($\mathbb{R}^1$), in contrast to the ellipsoid Gaussians used in 3DGS~\cite{kerbl3Dgaussians}. Following VTGS-SLAM~\cite{Hu2025VTGSSLAM}, our representation retains only the color ($\mathbb{R}^{3}$), a single variance term as radius ($\mathbb{R}^1$), and opacity ($\mathbb{R}^1$), while omitting the 4D rotation, 3D position, and two additional variance terms of the ellipsoid Gaussians; however, we allow Gaussians to move along the ray. Moreover, we omit the local densification process used by other 3DGS-based SLAM methods.

% \vspace{-0.12in}
\subsection{Mapping}
\begin{figure}[b]
    \vspace{-0.2in}
    \centering

    % ---- Left image ----
    \begin{minipage}[t]{0.3\linewidth}
        \centering
        \includegraphics[width=\linewidth]{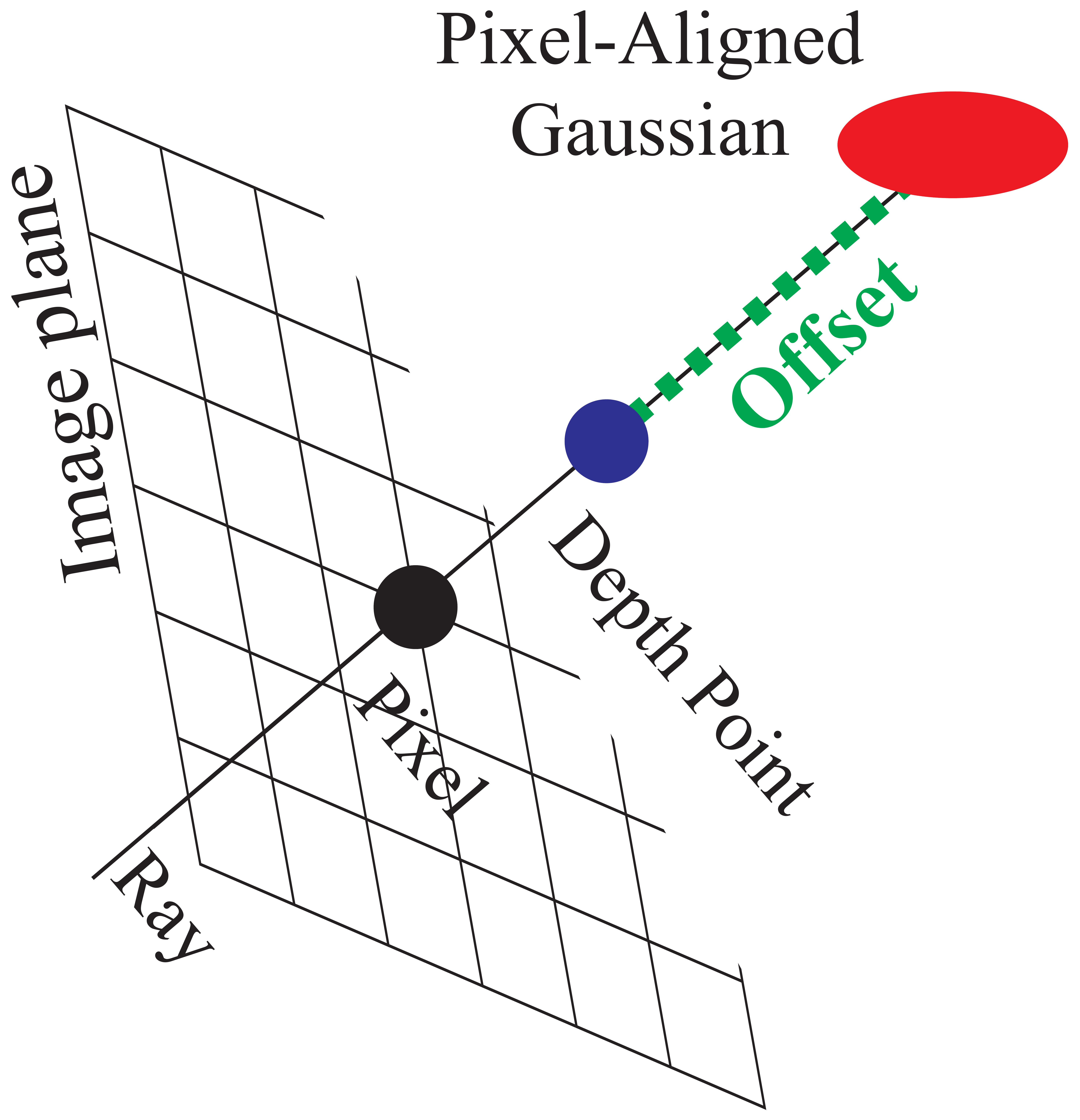}
        \subcaption{}
    \end{minipage}
    \hfill
    % ---- Right image ----
    \begin{minipage}[t]{0.67\linewidth}
        \centering
        \includegraphics[width=\linewidth]{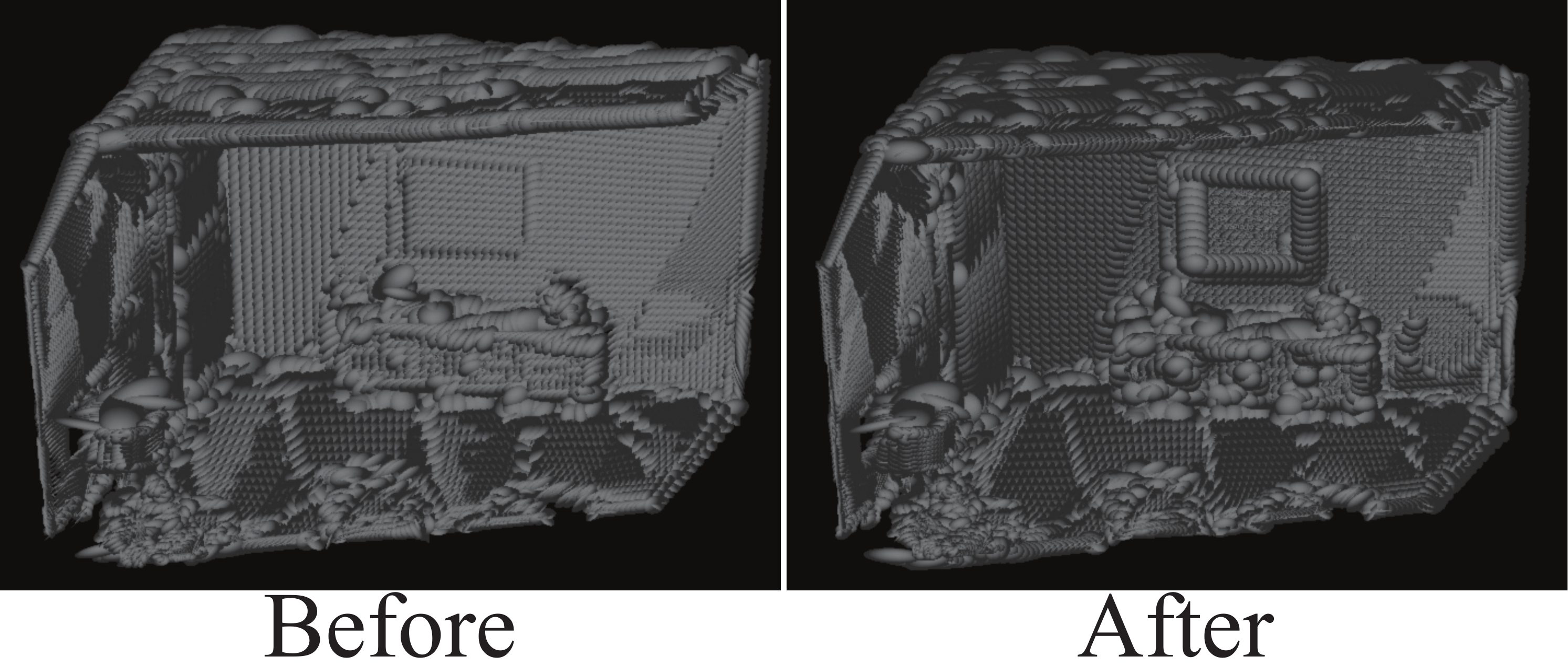}
        \subcaption{}
    \end{minipage}

    \vspace{-0.15in}
    \caption{(a) Pixel-aligned Gaussians at adjusted depth. (b) Illustration of Scale Normalization. Each Gaussian is shown with a minimum probability of 0.99.}
    \label{fig:Gaussians+Gaussianscale}
\end{figure}

% \vspace{-0.05in}
\noindent\textbf{Pixel-Aligned Gaussians at Adjusted Depth. }We align a Gaussian $g_i^j$ to the $j$-th pixel on the $i$-th depth map $D_i$, resulting in a set of Gaussians $G_i=\{g_i^j\}_{j=1}^J$. This design encourages Gaussians focus more on fitting the specific $i$-th frame and its neighboring frames. As illustrated in Fig.~\ref{fig:Gaussians+Gaussianscale} (a), we allow each Gaussian to move along the ray connecting the camera center to the associated pixel. Its position along the ray can be determined by $\tilde{d}_i^j=\tilde{D}_i(j)$, where $\tilde{D}_i=|D_i+\delta_i|$ is a depth map adjusted from $D_i$ with its depth offset $\delta_i$, as detailed in real examples in Fig.~\ref{fig:CameraEst}.

% Because of the adaptiveness and the dedication to specific pixels, 
Due to its adaptive, pixel-specific formulation, this offset significantly improves the rendering quality of pixel-aligned Gaussians, even under the constraints of simplified Gaussian modeling, restricted movement, and omitted densification, as shown in Fig.~\ref{fig:RenderComp}. We compare renderings with all Gaussians movable in the scene, the same number of Gaussians as ours but movable only in the view frustum, or the same number of Gaussians as ours located at GT depth. We can see that our pixel-aligned Gaussians at adjusted depth achieves the best rendering performance.

To initialize Gaussians using depth maps with missing values, we employ depth interpolation or render the missing depth regions using Gaussians from nearby previous views.

\noindent\textbf{Rendering RGB and Depth by Splatting. }We render pixel-aligned Gaussians via splatting, and optimize their attributes, including the depth offsets, by minimizing the rendering errors against RGB and depth observations. For a frame $\{V_i,D_i\}$, we render the associated Gaussians $G_i$ into RGB and depth images $\{V_k',D_k'\}$, i.e., $(V_k',D_k')=splat(G_i,p_k)$, where $k \in \{i, NN(i)\}$ and $NN(i)$ denotes the indices for the $NN$ neighboring frames of the $i$-th frame. We then minimize the rendering errors below,

\vskip -0.15in
\begin{equation}
\label{Eq:mappingrendering1}
\min_{G_i,\delta_i} \sum_{k} (\rho  ||V_k-V_k'||_1 + \tau L_{S}+ \sigma U_k ||D_k-D_k'||_1),
\end{equation}
% \vskip -0.12in

\noindent where $L_{S}$ is the SSIM loss between $V_k$ and $V_k'$, $U_k$ is a mask that filters out pixels without valid depth values, and $\rho$, $\tau$, and $\sigma$ are balance weights. Only Gaussians $G_i$ and depth offset $\delta_i$ in the current view are learnable, and all Gaussians and adjusted depth maps in the neighboring views are fixed. The optimization can maintain the appearance and geometry consistency of the Gaussians across the current frame and its neighbors.

% \vspace{-0.12in}
\subsection{Tracking}
% \vspace{-0.05in}
\noindent\textbf{Initialization. }We initialize the camera pose using the constant speed assumption unless otherwise specified. Meanwhile, to better handle scenarios with textureless appearance and large camera motion, our tracking can start from estimating cameras by minimizing the rendering errors against the current RGBD frame using the Gaussians in the previous frame, which improves our robustness in tracking.

\noindent\textbf{Matching a Frame to a Scene. }As illustrated in Fig.~\ref{fig:overview} (a)-(c), we estimate the camera pose $p_i$ of the $i$-th frame by aligning its depth to the scene based on geometric similarity. For a given depth map $D_i$, we represent the local geometry around a back-projected 3D point $pt_i^j$ as a Gaussian distribution, which centers at $d_i^j=D_i(j)$ and has a covariance matrix $c_i^j$ computed using its neighboring points $NN(pt_i^j)$. Specifically, for each frame, we uniformly select a set of 3D points $pt_i^j$ from the depth $D_i$ with a downsampling ratio $R$. Then, for each $pt_i^j$, we calculate its covariance matrix $c_i^j$ using its $K_c$ nearest neighbors $NN(pt_i^j)$. We then extract the scales and rotations from the covariance matrix using the SVD~\cite{Segal2009GeneralizedICP}. The resulting set of Gaussians centered at the selected points $pt_i^j$ on $D_i$ is denoted as $T_i$. 

Similarly, we maintain a global set of 3D Gaussians $T$ to represent the geometry of the scene. Unlike the pixel-aligned Gaussians used for appearance mapping, $T$ is progressively updated by incorporating non-overlapping 3D Gaussians from previous frames $\{T_1,...,T_{i-1}\}$. We then estimate $p_i$ by matching the Gaussians in $T_i$ to those in $T$.

\noindent\textbf{Generalized ICP for Matching.} We employ Generalized ICP (GICP)~\cite{Segal2009GeneralizedICP} for the matching process to improve tracking efficiency and robustness, as illustrated in Fig.~\ref{fig:overview} (b) and (c). Unlike standard ICP, GICP manages to maximize the overlap between source and target distributions, where each distribution is modeled as a set of Gaussians, i.e., $T_i$ and $T$. To enhance robustness, we apply GICP directly to depth points rather than 3D Gaussians~\cite{ha2024rgbdgsicpslam}. Furthermore, GICP provides a framework that supports efficient parallelization, which significantly speeds up the alignment in 3D.

Unlike GICP, which relies on the point-to-point distance, we first establish the correspondences between the Gaussians in $T_i$ and $T$ using the point-to-surface distance as the metric within each GICP iteration. Since we have the rotation matrix of each Gaussian via SVD, we regard the vector corresponding to the smallest scale as the normal vector of each Gaussian to calculate point-to-surface distances. 

Given these established correspondences between $T_i$ and $T$, GICP estimates the camera pose $p_i$ to maximize the overlap between Gaussians in $T_i$ and their correspondences in $T$. At each iteration, this is achieved by minimizing:

% \vskip -0.15in
% \abovedisplayskip -0.2in
\begin{equation}
\label{Eq:poseopt}
\min_{p_i} \ \sum_{a\in T_i,b\in T} e'(c_b+p_i\times c_a\times  p_i')e,
\end{equation}
% \vskip -0.1in

\noindent where $a$ and $b$ are corresponding Gaussians in $T_i$ and $T$, $e$ is the alignment error in each iteration, $c_a$ and $c_b$ are covariance matrices, and $'$ indicates the transpose operation. GICP iterates this optimization process until convergence.

\begin{table*}[t] %{r}{0.5\linewidth}

\begin{minipage}[t]{0.65\linewidth}
\caption{Rendering performance comparisons in PSNR $\uparrow$, SSIM $\uparrow$, and LPIPS $\downarrow$ on 3 datasets.} 
\setlength{\tabcolsep}{0.65mm}
	\centering
    \small
	\resizebox{\linewidth}{!}{
   \begin{tabular}{l|ccc|ccc|ccc}
    \toprule
    % \hline
    \textbf{Dataset} & \multicolumn{3}{c}{\textbf{\emph{Replica}}~\cite{replica}} & \multicolumn{3}{c}{\textbf{\emph{TUM}}~\cite{6385773ATERMSE_tumrgbd}} & \multicolumn{3}{c}{\textbf{\emph{ScanNet}}~\cite{scannet}} \\
     % \midrule
     \hline
      Method  &  PSNR $\uparrow$ &  SSIM $\uparrow$ &LPIPS $\downarrow$ &  PSNR $\uparrow$ &  SSIM $\uparrow$ &LPIPS $\downarrow$ &  PSNR $\uparrow$ &  SSIM $\uparrow$ &LPIPS $\downarrow$\\
    % \midrule
    \hline
   \rowcolor{gray!20}
\multicolumn{10}{l}{\textit{Neural Implicit Fields}} \\
    NICE-SLAM~\cite{Zhu2021NICESLAM} &24.42& 0.809 & 0.233& 14.86& 0.614 & 0.441 & 17.54 & 0.621 & 0.548  \\
    Vox-Fusion~\cite{Yang_Li_Zhai_Ming_Liu_Zhang_2022_voxfusion} &24.41 & 0.801 & 0.236 &16.46 & 0.677 &0.471 & 18.17 & 0.673 & 0.504 \\
   ESLAM~\cite{johari-et-al-2023-ESLAM} & 28.06 & 0.923 &0.245 & 15.26 &0.478 & 0.569&  15.29& 0.658 &  0.488 \\
     Point-SLAM~\cite{Sandström2023ICCVpointslam}  & 35.17 &0.975  & 0.124&  16.62 &  0.696 &  0.526 & 19.82&  0.751& 0.514\\
     Loopy-SLAM$*$~\cite{liso2024loopyslam} &  35.47 &  0.981  &  0.109 & 12.94 &  0.489 & 0.645 &15.23 & 0.629 & 0.671 \\
     % \midrule
     % \midrule
     \hline
     \hline
\rowcolor{gray!20}
\multicolumn{10}{l}{\textit{3D Gaussian Splatting}} \\
    SplaTAM~\cite{keetha2024splatam} & 34.11 & 0.970 & 0.100 &  22.80 &  0.893& 0.178& 19.14 &  0.716 &  0.358 \\
    Gaussian-SLAM~\cite{yugay2023gaussianslam} & 42.08 & 0.996 & 0.018 & 25.05& 0.929 & 0.168& 27.70 & 0.923 & 0.248 \\
    VTGS-SLAM~\cite{Hu2025VTGSSLAM} & 43.34 & 0.996 & \textbf{0.012} & 30.20 & 0.972 & 0.062 & 31.10 & 0.961 & 0.108 \\
    GS-ICP SLAM~\cite{ha2024rgbdgsicpslam} & 38.83 & 0.975 & 0.041 & 20.72 & 0.768 & 0.218 & - & - & - \\
    LoopSplat$*$~\cite{zhu2024_loopsplat} & 36.63 & 0.985 & 0.112 & 22.72 & 0.873 & 0.259 & 24.92 & 0.845 & 0.425 \\
    % \midrule
    % \midrule
    \hline
    \hline
   Ours & \textbf{44.87} & \textbf{0.998} & 0.021 & \textbf{38.60} & \textbf{0.997} & \textbf{0.012} & \textbf{42.31} & \textbf{0.997} & \textbf{0.049} \\
    % \bottomrule
    \hline
    \end{tabular}
    }
    % \vspace{-0.1in}
% \caption{Rendering performance comparison in PSNR $\uparrow$, SSIM $\uparrow$, and LPIPS $\downarrow$ on 3 Datasets.} 
% $*$ methods relying on pre-trained data-driven priors.}
% \vspace{-0.25in}
\label{tab:rendering}
\end{minipage}
\hfill
\begin{minipage}[t]{0.33\linewidth}
\caption{Reconstruction results in Depth L1 $[\mathrm{cm}]\downarrow$ and F1 $[\mathrm{\%}]\uparrow$ on Replica~\cite{replica}.}
\setlength{\tabcolsep}{1mm}
	\centering
    \small
	\resizebox{0.7\linewidth}{!}{
   \begin{tabular}{l|cc}
    \toprule
    % \hline
    % \textbf{Dataset} & \multicolumn{2}{c}{\textbf{\emph{Replica}}} \\
     % \midrule
     % \hline
      Method  &  L1 $\downarrow$ &  F1 $\uparrow$ \\
    % \midrule
    \hline
   \rowcolor{gray!20}
\multicolumn{3}{l}{\textit{Neural Implicit Fields}} \\
    NICE-SLAM~\cite{Zhu2021NICESLAM} & 2.97 & 43.9 \\
    Vox-Fusion~\cite{Yang_Li_Zhai_Ming_Liu_Zhang_2022_voxfusion} & 2.46 & 52.2 \\
   ESLAM~\cite{johari-et-al-2023-ESLAM} & 1.18 & 79.1  \\
   Co-SLAM~\cite{wang2023coslam} & 2.59 & 69.7 \\
   Point-SLAM~\cite{Sandström2023ICCVpointslam}  & 0.44 & 89.8  \\
     Loopy-SLAM$*$~\cite{liso2024loopyslam} &  0.35 &  90.8   \\
     % \midrule
     % \midrule
     \hline
     \hline
\rowcolor{gray!20}
\multicolumn{3}{l}{\textit{3D Gaussian Splatting}} \\
    SplaTAM~\cite{keetha2024splatam} & 0.72 & 86.1  \\
    GS-SLAM~\cite{yan2023gs} & 1.16 & 70.2 \\
    Gaussian-SLAM~\cite{yugay2023gaussianslam} & 0.68 & 88.9  \\
    VTGS-SLAM~\cite{Hu2025VTGSSLAM} & 0.51 & 90.4 \\
    LoopSplat$*$~\cite{zhu2024_loopsplat} & 0.53 & 90.0 \\
    \hline
    \hline
   Ours & \textbf{0.30} & \textbf{90.9} \\
    % \bottomrule
    \hline
    \end{tabular}
    }
% \caption{Reconstruction results in Depth L1 $[\mathrm{cm}]\downarrow$ and F1 $[\mathrm{\%}]\uparrow$ on Replica.}
% $*$ methods using pre-trained data-driven priors.}
% \vspace{-0.2in}
\label{Tab:reconreplicaperscene}
\end{minipage}
\vspace{-0.1in}
\end{table*}

\noindent\textbf{Scale Normalization. }One change we make here is to normalize the scale of each Gaussian, as shown in Fig.~\ref{fig:Gaussians+Gaussianscale} (b). Similar to~\cite{ha2024rgbdgsicpslam}, this change aims to minimize the impact of depth range variations across different frames, ensuring that all geometric matching is conducted at a consistent scale.

\noindent\textbf{Update Gaussian Set $T$. }After tracking the $i$-th frame, we add some Gaussians from $T_i$ into $T$ with the estimated pose $p_i$ and get the updated $T$ ready to track the next frame, as shown in Fig.~\ref{fig:overview} (d). We do not add Gaussians in $T_i$ that overlap with existing ones in $T$ to reduce redundancy.

\noindent\textbf{Analysis. }By modeling the local geometry around each point as a Gaussian distribution, we maintain a very compact set of Gaussians to represent the current scene that we have scanned and conduct a much more efficient tracking operation than the state-of-the-art methods.

\section{Experiments and Analysis}

\subsection{Experimental Setup}

\noindent\textbf{Implementation Details.}
To initialize Gaussians from ground truth depth images, we first inpaint~\cite{Navier-stokesinpainting} the missing depth values using the neighboring pixels. With the completed depth, we can initialize our pixel-aligned Gaussians to acquire better rendering performance, since the Gaussians are allowed to move along the ray. Further details are provided in the supplementary material.

\begin{figure*}[t]
  \centering
% \vspace{-0.2in}
   \includegraphics[width=\linewidth]{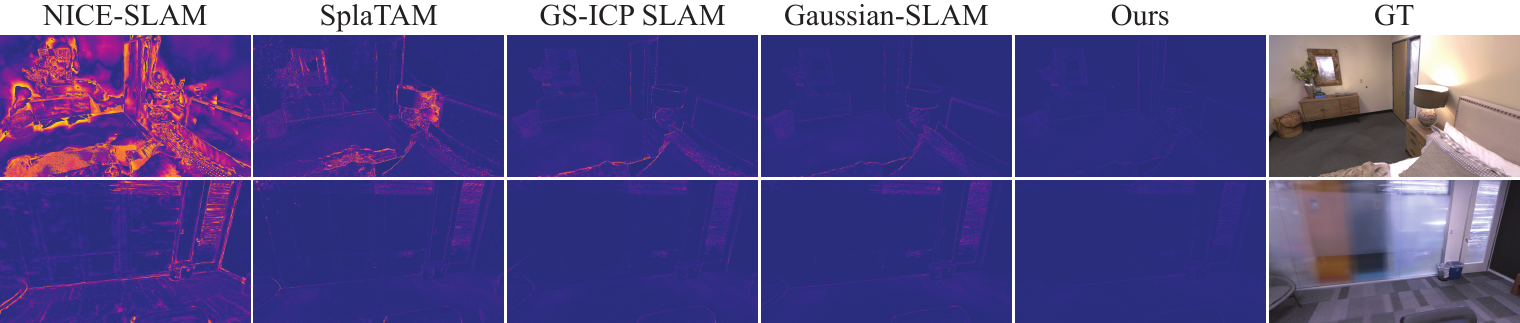}
   \vspace{-0.2in}
   \caption{Error map comparisons in rendering on Replica~\cite{replica}.}
   % ~\cite{replica}. }
   % Please find more comparisons in our video and supplementary materials.}
   \label{fig:mapping_replica}
\vspace{-0.05in}
\end{figure*}

\begin{figure*}[t]
  \centering
% \vspace{-0.1in}
   \includegraphics[width=\linewidth]{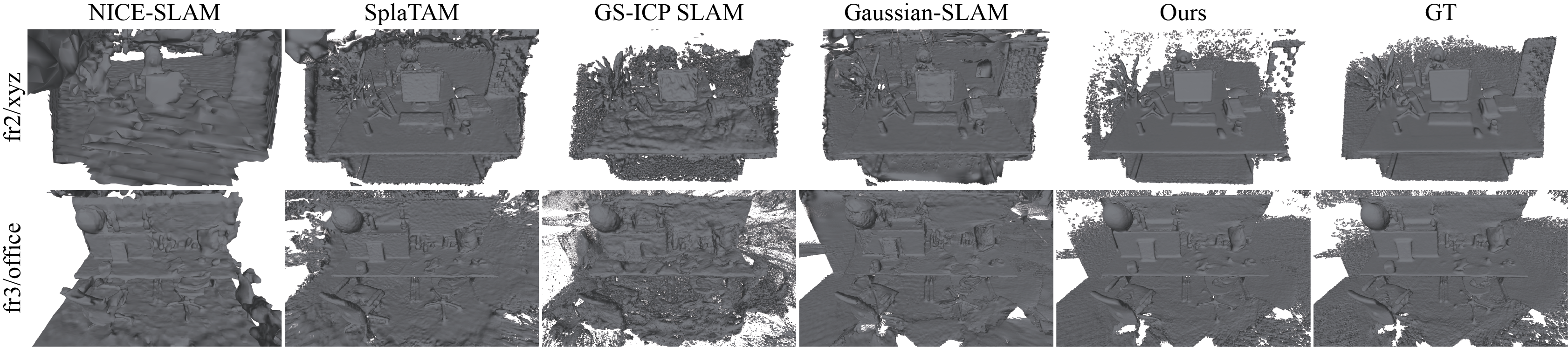}
   \vspace{-0.2in}
   \caption{Reconstruction comparisons on TUM-RGBD~\cite{6385773ATERMSE_tumrgbd}.}
   % \caption{Visual comparisons in reconstruction on TUM-RGBD~\cite{6385773ATERMSE_tumrgbd}.}
   % Please see more comparisons in our supplementary materials.}
   \label{fig:meshvis}
\vspace{-0.15in}
\end{figure*}

\noindent\textbf{Datasets and Metrics.}
We employ several widely used benchmarks in evaluations, including Replica~\cite{replica}, TUM-RGBD~\cite{6385773ATERMSE_tumrgbd}, ScanNet~\cite{scannet}, and ScanNet++~\cite{yeshwanthliu2023scannetpp}.

To evaluate the tracking performance, we employ the ATE RMSE $[\mathrm{cm}]$~\cite{6385773ATERMSE_tumrgbd}. To assess the rendering performance, we measure PSNR, SSIM~\cite{1284395_ssim}, and LPIPS~\cite{zhang2018unreasonableeffectivenessdeepfeatures_lpips}. Additionally, we reconstruct meshes of the scene using the Marching Cubes~\cite{Lorensen87marchingcubes}, following the approach in~\cite{Sandström2023ICCVpointslam}. The reconstruction quality is evaluated using the F1-score and depth L1. Please refer to our supplementary material for detailed descriptions of the datasets, evaluation metrics, and per-scene results across all benchmarks.
We primarily report the average results in the following tables. Note that $*$ indicates methods relying on pre-trained data-driven priors.

\noindent\textbf{Baselines.}
We compare our method, SGAD-SLAM, with the latest RGBD SLAM approaches, including NeRF-based RGBD SLAM methods: NICE-SLAM~\cite{Zhu2021NICESLAM}, Vox-Fusion~\cite{Yang_Li_Zhai_Ming_Liu_Zhang_2022_voxfusion}, ESLAM~\cite{johari-et-al-2023-ESLAM}, DF-Prior~\cite{Hu2023LNI-ADFP}, Co-SLAM~\cite{wang2023coslam}, Point-SLAM~\cite{Sandström2023ICCVpointslam}, and Loopy-SLAM~\cite{liso2024loopyslam}; as well as 3DGS-based RGBD SLAM methods: SplaTAM~\cite{keetha2024splatam}, 
% MonoGS~\cite{MatsukiCVPR2024_monogs}, 
GS-SLAM~\cite{yan2023gs}, Gaussian SLAM~\cite{yugay2023gaussianslam}, VTGS-SLAM~\cite{Hu2025VTGSSLAM}, GS-ICP SLAM~\cite{ha2024rgbdgsicpslam}, LoopSplat~\cite{zhu2024_loopsplat}, and CG-SLAM~\cite{hu2024cg}. Note that Point-SLAM~\cite{Sandström2023ICCVpointslam} requires ground truth depth
images as input to guide sampling during rendering, which is an unfair advantage over other NeRF-based methods. Additionally, some SLAM methods that incorporate pose graph optimization, such as Loopy-SLAM~\cite{liso2024loopyslam}, LoopSplat~\cite{zhu2024_loopsplat}, and CG-SLAM~\cite{hu2024cg}, leverage data-driven priors (e.g., pre-trained NetVLAD models~\cite{Arandjelovic16}) for loop closure detection and visibility checks. While these methods often report higher tracking accuracy, their reliance on pre-trained priors creates an unfair experimental setting compared to most SLAM methods that do not utilize such priors.

\begin{table} %{r}{0.5\linewidth}
% \vspace{-0.23in}
\caption{Tracking results in ATE RMSE $\downarrow [\mathrm{cm}]$ on Replica.}
% $*$ methods relying on pre-trained data-driven priors. $\dagger$ method using only RGB as input.}
% \vskip 0.05in
\setlength{\tabcolsep}{2pt}
  \centering
  \resizebox{\linewidth}{!}{
\begin{tabular}{lccccccccc}
\toprule
Method & \texttt{Rm0} & \texttt{Rm1} & \texttt{Rm2} & \texttt{Off0} & \texttt{Off1} & \texttt{Off2} & \texttt{Off3} & \texttt{Off4} & Avg. \\
\midrule
\rowcolor{gray!20}
\multicolumn{10}{l}{\textit{Neural Implicit Fields}} \\
NICE-SLAM~\cite{Zhu2021NICESLAM} & 1.69 & 2.04 & 1.55 & 0.99 & 0.90 & 1.39 & 3.97 & 3.08 & 1.95 \\
DF-Prior~\cite{Hu2023LNI-ADFP} & 1.39 & 1.55 & 2.60 & 1.09 & 1.23 & 1.61 & 3.61 & 1.42 & 1.81 \\
Vox-Fusion~\cite{Yang_Li_Zhai_Ming_Liu_Zhang_2022_voxfusion} & 0.27 & 1.33 & 0.47 & 0.70 & 1.11 & 0.46 & 0.26 & 0.58 & 0.65 \\
ESLAM~\cite{johari-et-al-2023-ESLAM} & 0.71 & 0.70 & 0.52 & 0.57 & 0.55 & 0.58 & 0.72 & 0.63 & 0.63 \\
Point-SLAM~\cite{Sandström2023ICCVpointslam} & 0.61 & 0.41 & 0.37 & 0.38 & 0.48 & 0.54 & 0.72 & 0.63 & 0.52 \\
\hdashline
Loopy-SLAM$*$~\cite{liso2024loopyslam} & 0.24 & 0.24 & 0.28 & 0.26 & 0.40 & 0.29 & 0.22 & 0.35 & 0.29 \\
\midrule
\midrule
\rowcolor{gray!20}
\multicolumn{10}{l}{\textit{3D Gaussian Splatting}} \\
SplaTAM~\cite{keetha2024splatam} & 0.31 & 0.40 & 0.29 & 0.47 & 0.27 & 0.29 & 0.32 & 0.55 & 0.36 \\
GS-SLAM~\cite{yan2023gs} & 0.48 & 0.53 & 0.33 & 0.52 & 0.41 & 0.59 & 0.46 & 0.70 & 0.50 \\
Gaussian-SLAM~\cite{yugay2023gaussianslam} & 0.29 & 0.29 & 0.22 & 0.37 & 0.23 & 0.41 & 0.30 & 0.35 & 0.31 \\
VTGS-SLAM~\cite{Hu2025VTGSSLAM} & 0.22 & 0.26 & 0.19 & 0.28 & 0.26 & 0.34 & 0.25 & 0.43 & 0.28 \\
GS-ICP SLAM~\cite{ha2024rgbdgsicpslam} & \textbf{0.15} & \textbf{0.16} & 0.11 & 0.18 & \textbf{0.12} & 0.17 & \textbf{0.16} & 0.21 & \textbf{0.16} \\
\hdashline
LoopSplat$*$~\cite{zhu2024_loopsplat} & 0.28 & 0.22 & 0.17 & 0.22 & 0.16 & 0.49 & 0.20 & 0.30 & 0.26 \\
CG-SLAM$*$~\cite{hu2024cg} & 0.29 & 0.27 & 0.25 & 0.33 & 0.14 & 0.28 & 0.31 & 0.29 & 0.27 \\
\midrule
\midrule
Ours & \textbf{0.15} & 0.17 & \textbf{0.10} & \textbf{0.16} & \textbf{0.12} & \textbf{0.16} & 0.25 & \textbf{0.20} & \textbf{0.16} \\

\bottomrule
\end{tabular}}
% \vspace{-0.1in}
% \caption{Tracking performance comparisons in ATE RMSE $\downarrow [\mathrm{cm}]$ on Replica~\cite{replica}. $*$ methods relying on pre-trained data-driven priors. $\dagger$ the method using only RGB as input.}
\label{Tab:camreplicaperscene}
\vspace{-0.21in}
\end{table}

\subsection{Evaluations}

\noindent\textbf{Replica. }We report our tracking results in Tab.~\ref{Tab:camreplicaperscene}, comparing our approach against both NeRF-based and 3DGS-based methods. We achieve the best tracking performance in 6 out of 8 scenes and the best average accuracy. Our method also shows improvements over methods that employ data-driven priors for loop detection with additional pose graph optimization, such as LoopSplat~\cite{zhu2024_loopsplat}, CG-SLAM~\cite{hu2024cg}, and Loopy-SLAM~\cite{liso2024loopyslam}.

Mapping evaluation results are presented in Tab.~\ref{tab:rendering}. Due to our pixel-aligned Gaussians with depth offsets, we can represent colors at each pixel more accurately, which shows significant improvements over the latest methods across all three metrics. We highlight our rendering accuracy in comparisons of error maps on rendered images in Fig.~\ref{fig:mapping_replica}, where we produce the minimum rendering errors.

Moreover, our method achieved the highest 3D reconstruction accuracy in Tab.~\ref{Tab:reconreplicaperscene}, where we follow previous methods to compare the accuracy of the depth rendered from reconstructed surfaces. Please find the visual comparison of reconstruction in our supplementary material.

\begin{table} %{r}{0.5\linewidth}
% \vskip -0.25in
\caption{Tracking results in ATE RMSE $\downarrow [\mathrm{cm}]$ on TUM-RGBD.}
% \vskip 0.05in
  \centering
  \resizebox{\linewidth}{!}{
\begin{tabular}{lcccc}
\toprule
Method & \texttt{fr1/desk} & \texttt{fr2/xyz} & \texttt{fr3/office} & Avg. \\
\midrule
\rowcolor{gray!20}
\multicolumn{5}{l}{\textit{Neural Implicit Fields}} \\
NICE-SLAM~\cite{Zhu2021NICESLAM} & 4.3 & 31.7 & 3.9 & 13.3 \\
Vox-Fusion~\cite{Yang_Li_Zhai_Ming_Liu_Zhang_2022_voxfusion} & 3.5 & 1.5 & 26.0 & 10.3 \\
Point-SLAM~\cite{Sandström2023ICCVpointslam} & 4.3 & 1.3 & 3.5 & 3.0 \\
\hdashline
Loopy-SLAM$*$~\cite{liso2024loopyslam} & 3.8 & 1.6 & 3.4 & 2.9 \\
\midrule
\midrule
\rowcolor{gray!20}
\multicolumn{5}{l}{\textit{3D Gaussian Splatting}} \\
SplaTAM~\cite{keetha2024splatam} & 3.4 & \textbf{1.2} & 5.2 & 3.3 \\ 
GS-SLAM~\cite{yan2023gs} & 3.3 & 1.3 & 6.6 & 3.7 \\
Gaussian-SLAM~\cite{yugay2023gaussianslam} & 2.6 & 1.3 & 4.6 & 2.9 \\
VTGS-SLAM~\cite{Hu2025VTGSSLAM} & 2.4 & 1.1 & 4.4 & 2.6 \\
GS-ICP SLAM~\cite{ha2024rgbdgsicpslam} & 2.7 & 1.8 & 2.7 & 2.4 \\
\hdashline
LoopSplat$*$~\cite{zhu2024_loopsplat} & 2.1 & 1.6 & 3.2 & 2.3 \\
CG-SLAM$*$~\cite{hu2024cg} & 2.4 & 1.2 & 2.5 & 2.0 \\
\midrule
\midrule
% Ours & \textbf{2.3} & 1.7 & \textbf{2.0} & \textbf{2.0} \\ % no rendering init
Ours & \textbf{2.2} & 1.7 & \textbf{2.0} & \textbf{2.0} \\ % rendering init
\bottomrule
\end{tabular}}
% \vspace{-0.1in}
% \caption{Tracking performance in ATE RMSE $\downarrow [\mathrm{cm}]$ on TUM-RGBD~\cite{6385773ATERMSE_tumrgbd}. $*$ methods using pre-trained data-driven priors.}
\vspace{-0.2in}
\label{Tab:camtumperscene}
\end{table}

\begin{figure*}[t]
  \centering
% \vspace{-0.1in}
   \includegraphics[width=\linewidth]{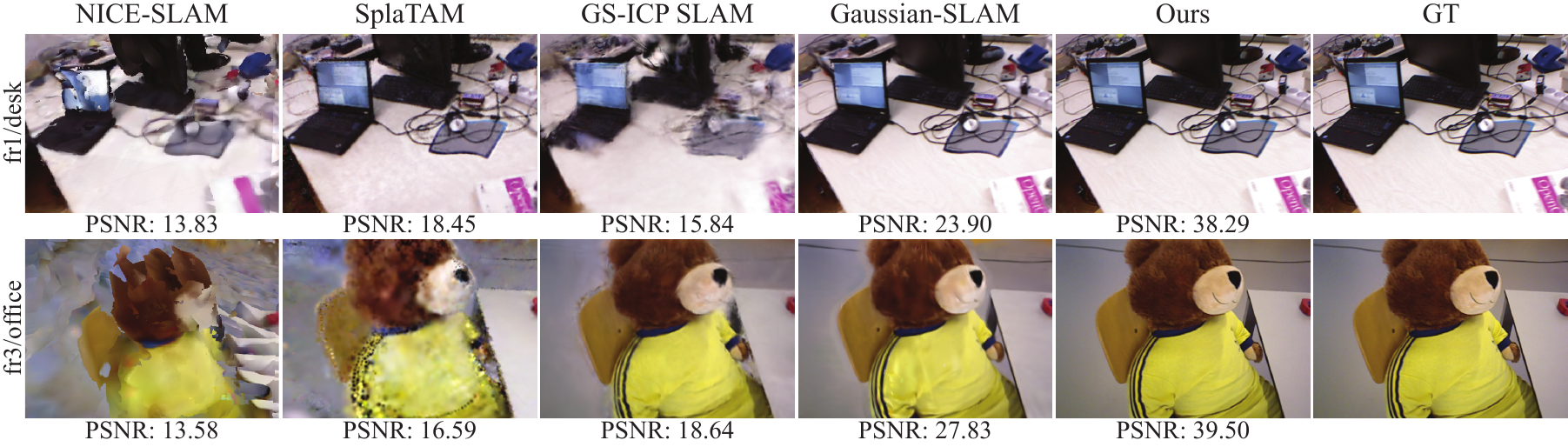}
   \vspace{-0.25in}
   \caption{Rendering comparisons on TUM-RGBD~\cite{6385773ATERMSE_tumrgbd}.}
   % Please see our video and supplementary materials for more comparisons.}
   \label{fig:mapping_tum}
\vspace{-0.1in}
\end{figure*}

\noindent\textbf{TUM-RGBD. }We report tracking comparisons with NeRF-based and 3DGS-based methods in Tab.~\ref{Tab:camtumperscene}. We not only achieve the best performance in average accuracy but also produce the best results in 2 out of 3 scenes. We also produce the best mapping performance as shown in Tab.~\ref{tab:rendering}, which significantly outperforms the other recent methods in all three metrics. The visual comparisons in rendering in Fig.~\ref{fig:mapping_tum} show our high fidelity rendering, indicating significant improvements in terms of PSNR and visual effect. The best rendering among the latest rendering-based methods justifies the effectiveness of our Gaussian representations, even if we use simplified Gaussians and also impose constraints in movements and densification. Additionally, we visualize the reconstruction by fusing the rendered depth with the estimated camera poses in Fig.~\ref{fig:meshvis}, which shows that we can recover the scene more accurately.

\begin{figure}[t] %{r}{0.5\linewidth}
  \centering
% \vspace{-0.1in}
   \includegraphics[width=\linewidth]{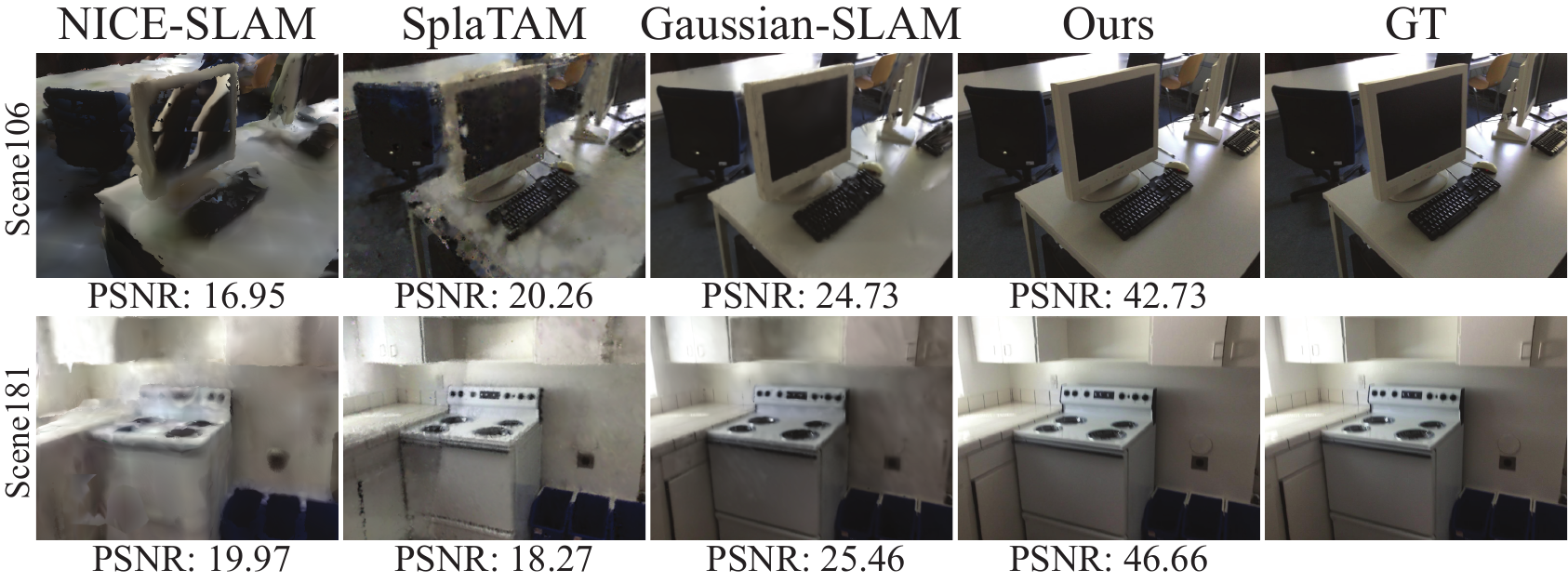}
   \vspace{-0.25in}
   \caption{Rendering comparisons on ScanNet~\cite{scannet}.}
   % Please see our video and supplementary materials for more comparisons.}
   \label{fig:mapping_scannet}
\vspace{-0.2in}
\end{figure}

\noindent\textbf{ScanNet. }We report numerical comparisons in tracking in Tab.~\ref{Tab:camscannetperscene}. Our method achieves the best in 5 out of 6 scenes, and is comparable to the best average result obtained by methods with loop closure. These comparisons show that our method can work well with real scanning data that have challenging illumination and large depth variations. Moreover, we report our mapping performance in Tab.~\ref{tab:rendering}. GS-ICP SLAM~\cite{ha2024rgbdgsicpslam} does not report its results on ScanNet, and we were unable to produce plausible results with its code as well. Our method can significantly outperform the NeRF-based and 3DGS-based methods in mapping. More detailed evaluations can be found in rendering comparisons in Fig.~\ref{fig:mapping_scannet} and reconstruction comparisons in the supplementary material, highlighting our advantages.

\begin{table}[t]
\centering
\begin{minipage}[t]{\linewidth}
\caption{Tracking results in ATE RMSE $\downarrow [\mathrm{cm}]$ on ScanNet.}
% \vskip 0.05in
\setlength{\tabcolsep}{2pt}
  \centering
  \resizebox{\linewidth}{!}{
\begin{tabular}{lccccccc}
\toprule
Method & \texttt{0000} & \texttt{0059} & \texttt{0106} & \texttt{0169} & \texttt{0181} & \texttt{0207} & Avg. \\
\midrule
\rowcolor{gray!20}
\multicolumn{8}{l}{\textit{Neural Implicit Fields}} \\
NICE-SLAM~\cite{Zhu2021NICESLAM} & 12.0 & 14.0 & 7.9 & 10.9 & 13.4 & 6.2 & 10.7 \\
Vox-Fusion~\cite{Yang_Li_Zhai_Ming_Liu_Zhang_2022_voxfusion} & 68.8 & 24.2 & 8.4 & 27.3 & 23.3 & 9.4 & 26.9 \\
Point-SLAM~\cite{Sandström2023ICCVpointslam} & \textbf{10.2} & 7.8 & 8.7 & 22.2 & 14.8 & 9.5 & 12.2 \\
\hdashline
Loopy-SLAM$*$~\cite{liso2024loopyslam} & 4.2 & 7.5 & 8.3 & 7.5 & 10.6 & 7.9 & 7.7 \\
\midrule
\midrule
\rowcolor{gray!20}
\multicolumn{8}{l}{\textit{3D Gaussian Splatting}} \\
SplaTAM~\cite{keetha2024splatam} & 12.8 & 10.1 & 17.7 & 12.1 & 11.1 & 7.5 & 11.9  \\ 
Gaussian-SLAM~\cite{yugay2023gaussianslam} & 24.8 & 8.6 & 11.3 & 14.6 & 18.7 & 14.4 & 15.4 \\
% GS-ICP SLAM~\cite{ha2024rgbdgsicpslam}  &  &  &  &  &  &  &  \\
VTGS-SLAM~\cite{Hu2025VTGSSLAM} & 17.8 & 8.7 & 11.8 & 10.5 & 10.6 & 8.6 & 11.3\\
\hdashline
LoopSplat$*$~\cite{zhu2024_loopsplat} & 6.2 & 7.1 & 7.4 & 10.6 & 8.5 & 6.6 & 7.7 \\
CG-SLAM$*$~\cite{hu2024cg} & 7.1 & 7.5 & 8.9 & 8.2 & 11.6 & 5.3 & 8.1 \\
\midrule
\midrule
Ours & 11.9 & \textbf{6.4} & \textbf{5.3} & \textbf{8.5} & \textbf{10.3} & \textbf{4.7} & \textbf{7.9} \\

\bottomrule
\end{tabular}}
% \vspace{-0.2in}
% \caption{Tracking performance in ATE RMSE $\downarrow [\mathrm{cm}]$ on ScanNet~\cite{scannet}. $*$ methods using pre-trained data-driven priors.}
\label{Tab:camscannetperscene}

\end{minipage}
\hfill
\vspace{0.005in}
\begin{minipage}[t]{\linewidth}
\caption{Tracking results in ATE RMSE $\downarrow [\mathrm{cm}]$ on ScanNet++.}
% \vskip 0.05in
\setlength{\tabcolsep}{2pt}
  \centering
  \resizebox{\linewidth}{!}{
\begin{tabular}{lcccccc}
\toprule
Method & \texttt{a} & \texttt{b} & \texttt{c} & \texttt{d} & \texttt{e} & Avg. \\
\midrule
\rowcolor{gray!20}
\multicolumn{7}{l}{\textit{Neural Implicit Fields}} \\
Point-SLAM~\cite{Sandström2023ICCVpointslam} & 246.16 & 632.99 & 830.79 & 271.42 & 574.86 & 511.24 \\
ESLAM~\cite{johari-et-al-2023-ESLAM} & 25.15 & 2.15 & 27.02 & 20.89 & 35.47 & 22.14 \\
\hdashline
Loopy-SLAM$*$~\cite{liso2024loopyslam} & - & - & 25.16 & 234.25 & 81.48 & 113.63 \\
\midrule
\midrule
\rowcolor{gray!20}
\multicolumn{7}{l}{\textit{3D Gaussian Splatting}} \\
SplaTAM~\cite{keetha2024splatam} & 1.50 & \textbf{0.57} & 0.31 & 443.10 & 1.58 & 89.41  \\ 
Gaussian-SLAM~\cite{yugay2023gaussianslam} & 1.37 & 5.97 & 2.70 & 2.35 & 1.02 & 2.68 \\
VTGS-SLAM~\cite{Hu2025VTGSSLAM} & 2.8 & 1.5 & 1.0 & 1.2 & 1.3 & 1.6 \\
\hdashline
LoopSplat$*$~\cite{zhu2024_loopsplat} & 1.14 & 3.16 & 3.16 & 1.68 & 0.91 & 2.05 \\
\midrule
\midrule
Ours(w/o Initialization) & 5.57 & 16.7 & 1.7 & 4.5 & 4.2 & 6.5 \\
Ours & \textbf{0.80} & 0.71 & \textbf{0.05} & \textbf{0.63} & \textbf{0.74} & \textbf{0.59} \\

\bottomrule
\end{tabular}}
% \vspace{-0.2in}
% \caption{Tracking performance in ATE RMSE $\downarrow [\mathrm{cm}]$ on ScanNet++~\cite{yeshwanthliu2023scannetpp}. $*$ methods relying on pre-trained data-driven priors.}
% \vspace{-0.1in}
\label{Tab:camscannetppperscene}
\end{minipage}
\vspace{-0.05in}
\end{table}

\begin{table} %{r}{0.5\linewidth}
% \vspace{-0.3in}
\caption{Rendering comparisons in PSNR $\uparrow$ on ScanNet++~\cite{yeshwanthliu2023scannetpp}.}
% \vskip 0.05in
\setlength{\tabcolsep}{0.7mm}
\small
  \centering
  % \vspace{-0.15in}
  \resizebox{\linewidth}{!}{
\begin{tabular}{lcccccc}
\toprule
Method & SplaTAM~\cite{keetha2024splatam} & \begin{tabular}[x]{@{}c@{}}Gaussian-\\SLAM~\cite{yugay2023gaussianslam}\end{tabular} & \begin{tabular}[x]{@{}c@{}}VTGS-\\SLAM~\cite{Hu2025VTGSSLAM}\end{tabular}  & \begin{tabular}[x]{@{}c@{}}Loop\\Splat$*$~\cite{zhu2024_loopsplat}\end{tabular} & Ours \\
\midrule
Training views & 26.71 & 30.38 & 32.22 & 30.20 & \textbf{36.73}  \\
Novel views & 17.82 & 21.27 & 21.46 & 21.30 & \textbf{22.32}  \\
\bottomrule
\end{tabular}
}
% \vspace{-0.2in}
% \caption{Rendering performance comparison in PSNR $\uparrow$ on ScanNet++~\cite{yeshwanthliu2023scannetpp}. $*$ methods using pre-trained data-driven priors.}
\vspace{-0.15in}
\label{Tab:renderscannetpp}
\end{table}

\noindent\textbf{ScanNet++. }Since scenes in ScanNet++ have large and sudden camera motions between consecutive frames, instead of using constant speed initialization for each frame, we employ RGBD odometry~\cite{colorptreg_odo} to initialize a camera pose, which is then roughly optimized by minimizing errors of the rendering with Gaussians in the previous frame for several iterations. Then we use the optimized camera pose as an initialization for our geometric matching process.

\begin{figure}[t]
    \centering
    % \vspace{-0.1in}

    \centering
    \includegraphics[width=\linewidth]{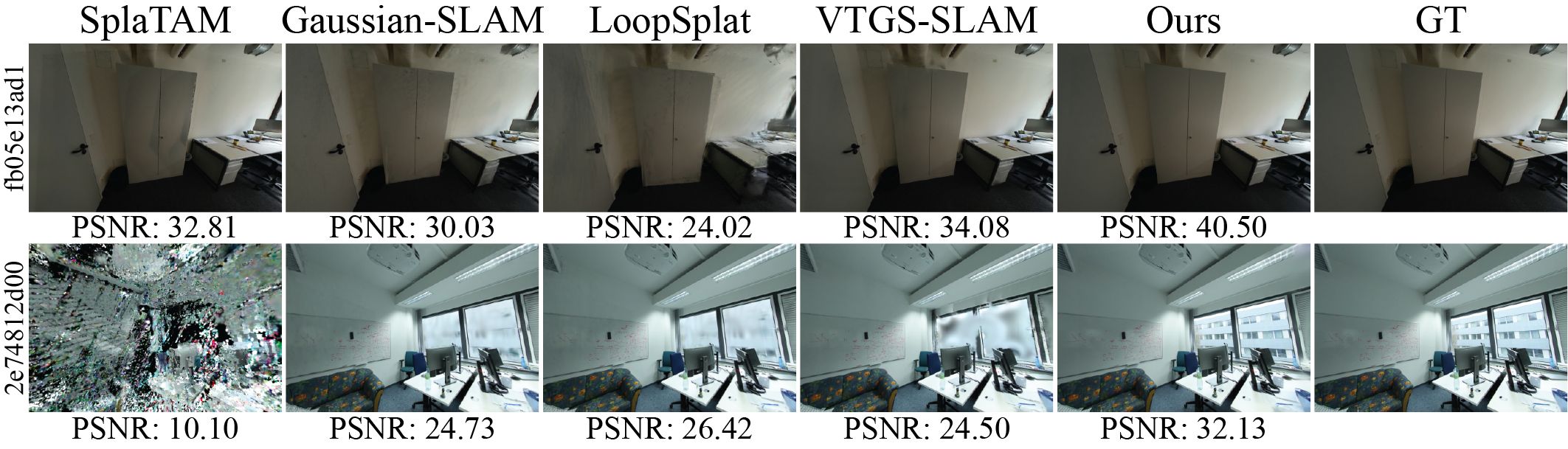}
    \vspace{-0.25in}
    \caption{Training view rendering comparisons on ScanNet++.}
    \vspace{-0.1in}
    \label{fig:mapping_scannetpp_train_nvs_a}
\end{figure}

\begin{figure}[t]
        \centering
        \includegraphics[width=\linewidth]{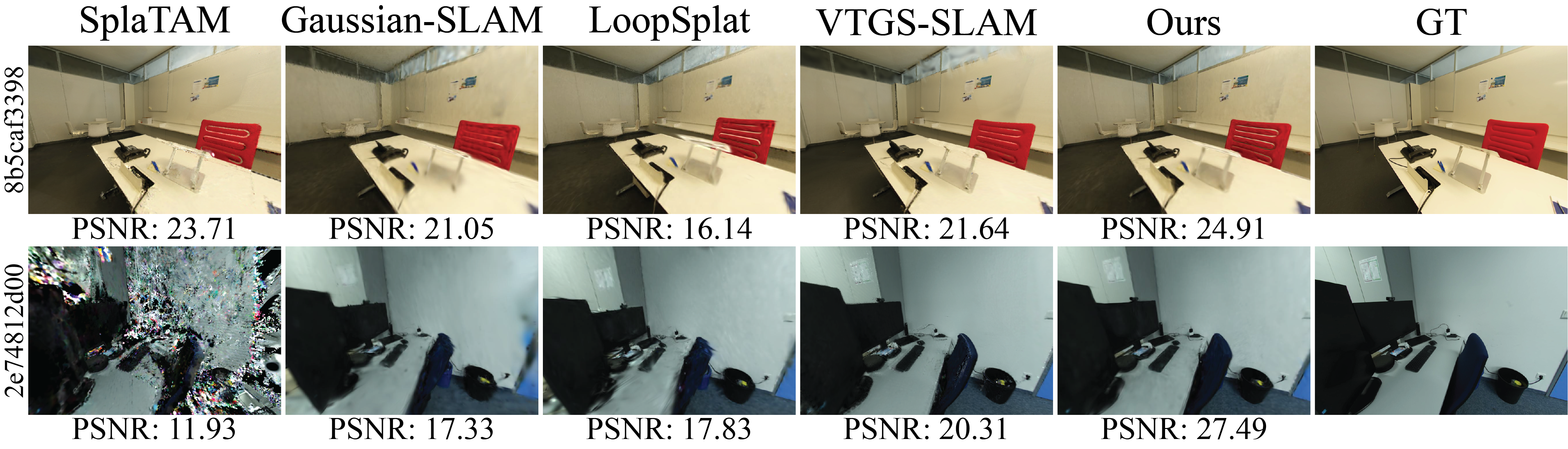}
\vspace{-0.25in}
    \caption{Novel view rendering comparisons on ScanNet++.}
    \vspace{-0.15in}
    \label{fig:mapping_scannetpp_train_nvs_b}
    % \vspace{-0.1in}
\end{figure}

We report our tracking performance in Tab.~\ref{Tab:camscannetppperscene}, which shows our robust performance on these challenging scenes. We also report the tracking performance with our initialization poses obtained by rendering, which highlights the significant improvements introduced by our tracking strategy. Moreover, we evaluate the mapping performance in Tab.~\ref{Tab:renderscannetpp} and Fig.~\ref{fig:mapping_scannetpp_train_nvs_a}. The comparisons highlight our advantages in rendering over state-of-the-art rendering-based SLAM methods. Furthermore, we evaluate novel view synthesis in Tab.~\ref{Tab:renderscannetpp} and Fig.~\ref{fig:mapping_scannetpp_train_nvs_b}, which demonstrate that we can also synthesize more plausible novel views.

\begin{table}[b!] %{r}{0.5\linewidth}
\vspace{-0.15in}
\caption{Runtime and Memory Usage on Replica. Ours$^\dagger$: Parallel on 8 GPUs. NUM1: Total Number of Gaussians. NUM2: Max Number of Learnable Gaussians.}
% \vskip 0.05in
\setlength{\tabcolsep}{0.7mm}
\centering
\small
  \resizebox{\linewidth}{!}{
\begin{tabular}{lccccc}
\toprule
 % & \begin{tabular}[x]{@{}c@{}}Tracking\\/Frame(s)$\downarrow$\end{tabular} & \begin{tabular}[x]{@{}c@{}}Mapping\\/Frame(s)$\downarrow$ \end{tabular} & \begin{tabular}[x]{@{}c@{}} Total\\/Frame(s)$\downarrow$\end{tabular} & \begin{tabular}[x]{@{}c@{}} Total Num of\\Gaussians\end{tabular} & \begin{tabular}[x]{@{}c@{}} Max Num of Le-\\arnable Gaussians\end{tabular}\\
 & \begin{tabular}[x]{@{}c@{}}Tracking\\/Frame(s)$\downarrow$\end{tabular} & \begin{tabular}[x]{@{}c@{}}Mapping\\/Frame(s)$\downarrow$ \end{tabular} & \begin{tabular}[x]{@{}c@{}} Total\\/Frame(s)$\downarrow$\end{tabular} & \begin{tabular}[x]{@{}c@{}} NUM1\end{tabular} & \begin{tabular}[x]{@{}c@{}} NUM2 \end{tabular}\\
\midrule
NICE-SLAM~\cite{Zhu2021NICESLAM} & 1.06 & 1.15 & 2.21 & - & - \\
Point-SLAM~\cite{Sandström2023ICCVpointslam} & 1.11 & 3.52 & 4.63 & - & - \\
SplaTAM~\cite{keetha2024splatam} & 2.70 & 4.89 & 7.59 & 5832$K$ & 5832$K$ \\
Gaussian-SLAM~\cite{yugay2023gaussianslam} & 0.83 & 0.93 & 1.76 & 32592$K$ & 1983$K$ \\
GS-ICP SLAM~\cite{ha2024rgbdgsicpslam} & 0.03 & 1.02 & 1.05 & 1544$K$ & 1544$K$ \\
\midrule
Ours & 0.01 & 0.89 & 0.90 & 326400$K$ & 816$K$ \\
Ours$^\dagger$ & \textbf{0.01} & \textbf{0.15} & \textbf{0.16} & 326400$K$ & 6528$K$ \\
\bottomrule
\end{tabular}
}
% \vspace{-0.15in}
% \caption{Runtime and Memory Usage on Replica. Ours$*$: Parallel on 8 GPUs. NUM1: Total Number of Gaussians. NUM2: Max Number of Learnable Gaussians.}
% \caption{Runtime and Memory Usage on Replica~\cite{replica}. }
% \vspace{-0.26in}
% \vspace{-0.25in}
% %Total GS: the total number of Gaussians representing a whole, Max GS: the maximum number of Gaussians in memory.
% }
\label{Tab:memoryandtime}
\end{table}

\begin{table}[t] %{r}{0.5\linewidth}
% \vspace{-0.23in}
\caption{Ablation study on Gaussian modeling on TUM-RGBD~\cite{6385773ATERMSE_tumrgbd}. Mode1: All Gaussians Movable in the Whole Scene; Mode2: Gaussians Movable in the View Frustum; Mode3: Gaussians Fixed at Depth; Ours: Gaussians Movable Along the Ray.}
% \vskip 0.05in
% \vspace{-0.05in}
\setlength{\tabcolsep}{1mm}
\small
\centering
  % \resizebox{\linewidth}{!}{
\begin{tabular}{lcccc}
\toprule
 % & \begin{tabular}[x]{@{}c@{}}All Gaussians Movable\\in the Whole Scene\end{tabular} & \begin{tabular}[x]{@{}c@{}}Gaussians Movable\\in the View Frustum\end{tabular} & \begin{tabular}[x]{@{}c@{}}Gaussians Fixed\\at Depth\end{tabular} & \begin{tabular}[x]{@{}c@{}}Gaussians Movable\\Along the Ray (Ours)\end{tabular} \\
 & Mode1 & Mode2 & Mode3 & Ours \\
\midrule
Num of Gaussians & 4450$K$ & 205$K$ & 205$K$ & 205$K$ \\ 
Num of attributes & 14 & 14 & 5 & 6 \\ 
\midrule
PSNR$\uparrow$ & 20.90 & 37.67 & 37.16 & \textbf{38.60} \\
SSIM$\uparrow$ & 0.771 & 0.995 & 0.995 & \textbf{0.997} \\
LPIPS$\downarrow$ & 0.216 & 0.017 & 0.019 & \textbf{0.012} \\
\bottomrule
\end{tabular}
% }
% \vspace{-0.2in}
% \caption{Ablation study on Gaussian modeling in rendering on TUM-RGBD. Mode1: All Gaussians Movable in the Whole Scene, Mode2: Gaussians Movable in the View Frustum, Mode3: Gaussians Fixed at Depth, Mode4: Gaussians Movable Along the Ray (Ours).}
% \vspace{-0.05in}

\label{Tab:ablation_gaussianmodle}
\end{table}

\subsection{Ablation Studies and Analysis}
\noindent\textbf{Pixel-aligned Gaussians. }We highlight our Gaussian representations in rendering each frame by replacing them with the ones in the original 3DGS. Tab.~\ref{Tab:ablation_gaussianmodle} reports comparisons with many more Gaussians that are movable in the whole scene, the same number of Gaussians as ours that are only movable in each frame, and the same number of Gaussians as ours but fixed at the depth points. The comparisons indicate that our pixel-aligned Gaussians placed at the adjusted depth can significantly improve the rendering performance and reduce the storage due to fewer attributes for Gaussians.

\begin{table}[t] %{r}{0.5\linewidth}
% \vspace{-0.25in}
\caption{Ablation study on robustness to additional noise in rendering and tracking on \texttt{fr3/office} in TUM-RGBD~\cite{6385773ATERMSE_tumrgbd}.}
% \vskip 0.05in
% \vspace{-0.05in}
% \setlength{\tabcolsep}{0.3mm}
\small
\centering
  \resizebox{0.85\linewidth}{!}{
\begin{tabular}{lccccc}
\toprule
 & \begin{tabular}[x]{@{}c@{}}10$\%$ \end{tabular} & \begin{tabular}[x]{@{}c@{}}20$\%$ \end{tabular} & \begin{tabular}[x]{@{}c@{}}30$\%$ \end{tabular} & \begin{tabular}[x]{@{}c@{}}40$\%$ \end{tabular} & \begin{tabular}[x]{@{}c@{}} Ours\end{tabular}\\
\midrule
PSNR$\uparrow$ & 39.86 & 39.86 & 39.86 & 39.86 & \textbf{39.86} \\
SSIM$\uparrow$ & 0.997 & 0.997 & 0.997 & 0.997 &  \textbf{0.997} \\
LPIPS$\downarrow$ & 0.013 & 0.013 & 0.013 & 0.013 & \textbf{0.012}  \\
ATE RMSE$\downarrow$ & 2.1 & 2.1 & 2.2 & 2.2 & \textbf{2.0} \\
\bottomrule
\end{tabular}
}
% \vspace{-0.15in}
% \caption{Ablation study on robustness to additional noise in rendering and tracking on fr3/office in TUM-RGBD.}
% \vspace{-0.05in}
\label{Tab:ablation_noisedepth}
\end{table}

\noindent\textbf{Robustness. }Although we report our results on real data corrupted with noise, we further evaluate our robustness to additional noise in Tab.~\ref{Tab:ablation_noisedepth}. We add these additional noises to the depth values of some pixels that are randomly sampled, such as $10\%$-$40\%$, and evaluate the tracking and mapping performance. Due to the learnable depth offset, our rendering does not get impacted at all. Since we employ Gaussian distributions to represent the geometry around each depth point and perform tracking by aligning these distributions, our tracking is also very robust to noise.

\begin{table}
% \vspace{-0.25in}
\caption{Ablation study on metrics for correspondence and scale regularization on TUM-RGBD. Ours: Point2surf \& Ellipse scale.}
% \vspace{-0.05in}
% \vskip 0.05in
% \setlength{\tabcolsep}{2pt}
    % \renewcommand{\arraystretch}{1.2}
\centering
  \resizebox{\linewidth}{!}{
\begin{tabular}{lccccc}
\toprule
 % & Point2point w/o scale(GICP) & Point2point w/ scale & Point2surf w/o scale & Point2surf w/ Plane scale & Ours  \\
 & \begin{tabular}[x]{@{}c@{}}Point2point\\w/o scale(GICP)\end{tabular} 
 & \begin{tabular}[x]{@{}c@{}}Point2point\\w/ Ellipse scale\end{tabular} 
 & \begin{tabular}[x]{@{}c@{}}Point2surf\\w/o scale\end{tabular} 
 & \begin{tabular}[x]{@{}c@{}}Point2surf\\w/ Plane scale\end{tabular} 
 & Ours \\
\midrule
ATE RMSE$\downarrow$ & 106.1 & 17.0 & 104.6 & 27.0 & \textbf{2.0} \\
\bottomrule
\end{tabular}}
% \vspace{-0.1in}
% \caption{Ablation study on metrics for correspondence and scale regularization on TUM-RGBD~\cite{6385773ATERMSE_tumrgbd}. Ours employed Point2surf \& Ellipse scale.}
\vspace{-0.1in}
\label{Tab:ablation_p2p_scale}
\end{table}

\noindent\textbf{Scale Normalization. }As shown in Tab.~\ref{Tab:ablation_p2p_scale}, our method achieves more robust tracking through three key improvements over vanilla GICP: geometry distribution modeling, scale normalization, and the point-to-surface distance for point matching. In contrast, recent methods such as GS-ICP SLAM~\cite{ha2024rgbdgsicpslam} and G2S-ICP SLAM~\cite{pak2025g2sicpslamgeometryawaregaussian} align appearance Gaussians rather than raw depth points. Our ablations demonstrate that the tracking degrades when using noise-sensitive point-to-point distances, removing scale normalization, or applying the flat-plane scale normalization in~\cite{Segal2009GeneralizedICP}.

\noindent\textbf{Time and Storage Complexity. }Tab.~\ref{Tab:memoryandtime} shows our advantages in scalability and efficiency. We can learn the most Gaussians to cover a scene, and it is also the fastest to produce the best rendering. Meanwhile, we only need to maintain and optimize a small number of Gaussians on the current frame. If we map the scene with 8 GPUs in parallel, where we only use Gaussians associated with each frame in mapping, we can process the scene even faster.

\section{Conclusion}
We propose SGAD-SLAM, a scalable and highly efficient 3DGS-based RGBD SLAM system based on better radiance field modeling by splatting pixel-aligned Gaussians at adjusted depth. Our method can represent the scene accurately and improve the rendering quality as well while successfully eliminating the need to maintain and optimize all Gaussians in the scene during frame rendering. This design not only significantly improves our capability to handle large scenes that require a large number of Gaussians to cover, but also speeds up the rendering by removing the reliance on keyframes as rendering targets. Beyond the rendering, we also introduce a novel tracking strategy that significantly improves the accuracy and the efficiency. We model the geometry around each depth point as a Gaussian distribution and estimate camera poses by aligning the distributions on the frame to the scene based on the geometry similarity. Our extensive evaluations on widely used benchmarks justify these designs and demonstrate our advantages over recent state-of-the-art methods in terms of accuracy, scalability, runtime, and storage complexity.

\section*{Acknowledgements}
This project was partially supported by an NVIDIA academic award and a Richard Barber research award.

{
    \small
    \bibliographystyle{ieeenat_fullname_cvpr2026}
    \bibliography{papers}

@ARTICLE{Lorensen87marchingcubes,
    author = {William E. Lorensen and Harvey E. Cline},
    title = {Marching cubes: A high resolution {3D} surface construction algorithm},
    journal = {Computer Graphics},
    year = {1987},
    volume = {21},
    number = {4},
    pages = {163--169}
}

@inproceedings{mildenhall2020nerf,
  title={{NeRF}: Representing Scenes as Neural Radiance Fields for View Synthesis},
  author={Ben Mildenhall and Pratul P. Srinivasan and Matthew Tancik and Jonathan T. Barron and Ravi Ramamoorthi and Ren Ng},
  year={2020},
  booktitle={European Conference on Computer Vision},
}

@InProceedings{Zhu2021NICESLAM,
      author    = {Zhu, Zihan and Peng, Songyou and Larsson, Viktor and Xu, Weiwei and Bao, Hujun and Cui, Zhaopeng and Oswald, Martin R. and Pollefeys, Marc},
      title     = {NICE-SLAM: Neural Implicit Scalable Encoding for SLAM},
      booktitle = {{IEEE} Conference on Computer Vision and Pattern Recognition},
      year      = {2022}
  }

@inproceedings{yu_and_fridovichkeil2021plenoxels,
      title={Plenoxels: Radiance Fields without Neural Networks},
      author={{Sara Fridovich-Keil and Alex Yu} and Matthew Tancik and Qinhong Chen and Benjamin Recht and Angjoo Kanazawa},
      year={2022},
      booktitle={{IEEE} Conference on Computer Vision and Pattern Recognition},
}

@article{mueller2022instant,
    title = {Instant Neural Graphics Primitives with a Multiresolution Hash Encoding},
    author = {Thomas M\"uller and Alex Evans and Christoph Schied and Alexander Keller},
    journal = {arXiv:2201.05989},
    year = {2022},
}

@article{ruckert2021adop,
  title={Adop: Approximate differentiable one-pixel point rendering},
  author={R{\"u}ckert, Darius and Franke, Linus and Stamminger, Marc},
  journal={arXiv:2110.06635},
  year={2021}
}

@InProceedings{NeuralRGBDSurfaceReconstruction_2022_CVPR,
    author    = {Azinovi\'c, Dejan and Martin-Brualla, Ricardo and Goldman, Dan B and Nie{\ss}ner, Matthias and Thies, Justus},
    title     = {Neural RGB-D Surface Reconstruction},
    booktitle = {IEEE Conference on Computer Vision and Pattern Recognition},
    year      = {2022},
    pages     = {6290-6301}
}

@article{Yu2022MonoSDF,
  author    = {Yu, Zehao and Peng, Songyou and Niemeyer, Michael and Sattler, Torsten and Geiger, Andreas},
  title     = {{MonoSDF}: Exploring Monocular Geometric Cues for Neural Implicit Surface Reconstruction},
  journal={ArXiv},
  year={2022},
  volume={abs/2022.00665}
}

@InProceedings{wang2022neuris, 
   Author = {Wang, Jiepeng and Wang, Peng and Long, Xiaoxiao and Theobalt, Christian and Komura, Taku and Liu, Lingjie and Wang, Wenping},
   Title = {{NeuRIS}: Neural Reconstruction of Indoor Scenes Using Normal Priors},
   BookTitle = {European Conference on Computer Vision},
   year = {2022},
}

@inproceedings{guo2022manhattan,
  title={Neural 3D Scene Reconstruction with the Manhattan-world Assumption},
  author={Guo, Haoyu and Peng, Sida and Lin, Haotong and Wang, Qianqian and Zhang, Guofeng and Bao, Hujun and Zhou, Xiaowei},
  booktitle={{IEEE} Conference on Computer Vision and Pattern Recognition},
  year={2022}
}

@article{park2021nerfies,
  author    = {Park, Keunhong and Sinha, Utkarsh and Barron, Jonathan T. and Bouaziz, Sofien and Goldman, Dan B and Seitz, Steven M. and Martin-Brualla, Ricardo},
  title     = {Nerfies: Deformable Neural Radiance Fields},
  journal   = {{IEEE} International Conference on Computer Vision},
  year      = {2021},
}

@inproceedings{schoenberger2016sfm,
    author={Sch\"{o}nberger, Johannes Lutz and Frahm, Jan-Michael},
    title={Structure-from-Motion Revisited},
    booktitle={{IEEE} Conference on Computer Vision and Pattern Recognition},
    year={2016},
}

@inproceedings{schoenberger2016mvs,
    author={Sch\"{o}nberger, Johannes Lutz and Zheng, Enliang and Pollefeys, Marc and Frahm, Jan-Michael},
    title={Pixelwise View Selection for Unstructured Multi-View Stereo},
    booktitle={European Conference on Computer Vision},
    year={2016},
}

@article{dai2017bundlefusion,
  title={BundleFusion: Real-time Globally Consistent 3D Reconstruction using On-the-fly Surface Re-integration},
  author={Dai, Angela and Nie{\ss}ner, Matthias and Zoll{\"o}fer, Michael and Izadi, Shahram and Theobalt, Christian},
  journal={ACM Transactions on Graphics},
  year={2017}
}

@article{replica,
  author    = {Julian Straub and
               Thomas Whelan and
               Lingni Ma and
               Yufan Chen and
               Erik Wijmans and
               Simon Green and
               Jakob J. Engel and
               Raul Mur{-}Artal and
               Carl Ren and
               Shobhit Verma and
               Anton Clarkson and
               Mingfei Yan and
               Brian Budge and
               Yajie Yan and
               Xiaqing Pan and
               June Yon and
               Yuyang Zou and
               Kimberly Leon and
               Nigel Carter and
               Jesus Briales and
               Tyler Gillingham and
               Elias Mueggler and
               Luis Pesqueira and
               Manolis Savva and
               Dhruv Batra and
               Hauke M. Strasdat and
               Renzo De Nardi and
               Michael Goesele and
               Steven Lovegrove and
               Richard A. Newcombe},
  title     = {The Replica Dataset: {A} Digital Replica of Indoor Spaces},
  journal   = {CoRR},
  volume    = {abs/1906.05797},
  year      = {2019},
}

@article{scannet,
  author    = {Angela Dai and
               Angel X. Chang and
               Manolis Savva and
               Maciej Halber and
               Thomas A. Funkhouser and
               Matthias Nie{\ss}ner},
  title     = {ScanNet: Richly-annotated 3D Reconstructions of Indoor Scenes},
  journal   = {CoRR},
  volume    = {abs/1702.04405},
  year      = {2017}
}

@inproceedings{wang2022go-surf,
  author={Wang, Jingwen and Bleja, Tymoteusz and Agapito, Lourdes},
  booktitle={International Conference on 3D Vision},
  title={GO-Surf: Neural Feature Grid Optimization for Fast, High-Fidelity RGB-D Surface
  Reconstruction},
  year={2022},
}

@article{sun2021neucon,
  title={{NeuralRecon}: Real-Time Coherent {3D} Reconstruction from Monocular Video},
  author={Sun, Jiaming and Xie, Yiming and Chen, Linghao and Zhou, Xiaowei and Bao, Hujun},
  journal={{IEEE} Conference on Computer Vision and Pattern Recognition},
  year={2021}
}

@article{bozic2021transformerfusion,
title={TransformerFusion: Monocular RGB Scene Reconstruction using Transformers},
author={Bozic, Aljaz and Palafox, Pablo and Thies, Justus and Dai, Angela and Niessner, Matthias},
journal={Advances in Neural Information Processing Systems},
year={2021}}

@article{DBLP:journals/corr/abs-2209-15153,
  author       = {Zi{-}Xin Zou and
                  Shi{-}Sheng Huang and
                  Yan{-}Pei Cao and
                  Tai{-}Jiang Mu and
                  Ying Shan and
                  Hongbo Fu},
  title        = {MonoNeuralFusion: Online Monocular Neural 3D Reconstruction with Geometric Priors},
  journal      = {CoRR},
  volume       = {abs/2209.15153},
  year         = {2022},
}

@inproceedings{sucar2021imap,
  title={iMAP: Implicit mapping and positioning in real-time},
  author={Sucar, Edgar and Liu, Shikun and Ortiz, Joseph and Davison, Andrew J},
  booktitle={Proceedings of the IEEE/CVF International Conference on Computer Vision},
  pages={6229--6238},
  year={2021}
}

@article{kong2023vmap,
  title={vMAP: Vectorised Object Mapping for Neural Field SLAM},
  author={Kong, Xin and Liu, Shikun and Taher, Marwan and Davison, Andrew J},
  journal={arXiv preprint arXiv:2302.01838},
  year={2023}
}

@misc{wang2023coslam,
      title={Co-SLAM: Joint Coordinate and Sparse Parametric Encodings for Neural Real-Time SLAM}, 
      author={Hengyi Wang and Jingwen Wang and Lourdes Agapito},
      year={2023},
      eprint={2304.14377},
      archivePrefix={arXiv},
}

@misc{haghighi2023neural,
      title={Neural Implicit Dense Semantic SLAM}, 
      author={Yasaman Haghighi and Suryansh Kumar and Jean-Philippe Thiran and Luc Van Gool},
      year={2023},
      eprint={2304.14560},
      archivePrefix={arXiv},
      primaryClass={cs.CV}
}

@INPROCEEDINGS{6385773ATERMSE_tumrgbd,
  author={Sturm, Jürgen and Engelhard, Nikolas and Endres, Felix and Burgard, Wolfram and Cremers, Daniel},
  booktitle={2012 IEEE/RSJ International Conference on Intelligent Robots and Systems}, 
  title={A benchmark for the evaluation of RGB-D SLAM systems}, 
  year={2012},
  volume={},
  number={},
  pages={573-580},
  doi={10.1109/IROS.2012.6385773}}

@inproceedings{Yang_Li_Zhai_Ming_Liu_Zhang_2022_voxfusion,  
 title={Vox-Fusion: Dense Tracking and Mapping with Voxel-based Neural Implicit Representation}, 
 url={http://dx.doi.org/10.1109/ismar55827.2022.00066}, 
 DOI={10.1109/ISMAR55827.2022.00066}, 
 booktitle={2022 IEEE International Symposium on Mixed and Augmented Reality (ISMAR)}, 
 author={Yang, Xingrui and Li, Hai and Zhai, Hongjia and Ming, Yuhang and Liu, Yuqian and Zhang, Guofeng}, 
 year={2022}, 
 month={Dec}, 
 language={en-US} 
 }

@inproceedings{sijia2023quantized,
    title={Coordinate Quantized Neural Implicit Representations for Multi-view 3D Reconstruction},
    author={Sijia Jiang and Jing Hua and Zhizhong Han},
    booktitle={{IEEE} International Conference on Computer Vision},
    year={2023}
}

@inproceedings{NeuralTPS,
  author = {Chao Chen and Zhizhong Han and Yu-Shen Liu},
  title = {Unsupervised Inference of Signed Distance Functions from Single Sparse Point Clouds without Learning Priors},
  booktitle = {Proceedings of the IEEE/CVF Conference on Computer Vision and Pattern Recognition (CVPR)},
  year = {2023},
}

@inproceedings{zhang2023goslam,
    title     = {GO-SLAM: Global Optimization for Consistent 3D Instant Reconstruction},
    author    = {Zhang, Youmin and Tosi, Fabio and Mattoccia, Stefano and Poggi, Matteo},
    booktitle = {Proceedings of the IEEE/CVF International Conference on Computer Vision (ICCV)},
    month     = {October},
    year      = {2023}
  }

@inproceedings{tofslam,
    title={Multi-Modal Neural Radiance Field for Monocular Dense SLAM with a Light-Weight ToF Sensor},
    author={Liu Xinyang and Li Yijin and Teng Yanbin and Bao Hujun and Zhang Guofeng and Zhang Yinda and Cui Zhaopeng},
    booktitle={International Conference on Computer Vision (ICCV)},
    year={2023}
    }

@misc{teigen2023rgbd,
      title={RGB-D Mapping and Tracking in a Plenoxel Radiance Field}, 
      author={Andreas L. Teigen and Yeonsoo Park and Annette Stahl and Rudolf Mester},
      year={2023},
      eprint={2307.03404},
      archivePrefix={arXiv},
      primaryClass={cs.CV}
}

@misc{sandström2023uncleslam,
      title={UncLe-SLAM: Uncertainty Learning for Dense Neural SLAM}, 
      author={Erik Sandström and Kevin Ta and Luc Van Gool and Martin R. Oswald},
      year={2023},
      eprint={2306.11048},
      archivePrefix={arXiv},
      primaryClass={cs.CV}
}

@inproceedings{Hu2023LNI-ADFP,
      title = {Learning Neural Implicit through Volume Rendering with Attentive Depth Fusion Priors},
      author = {Hu, Pengchong and Han, Zhizhong},
      booktitle = {Advances in Neural Information Processing Systems (NeurIPS)},
      year = {2023}
    }

@misc{li2023rico,
      title={RICO: Regularizing the Unobservable for Indoor Compositional Reconstruction}, 
      author={Zizhang Li and Xiaoyang Lyu and Yuanyuan Ding and Mengmeng Wang and Yiyi Liao and Yong Liu},
      year={2023},
      eprint={2303.08605},
      archivePrefix={arXiv},
      primaryClass={cs.CV}
}

@inproceedings{keetha2024splatam,
        title={SplaTAM: Splat, Track \& Map 3D Gaussians for Dense RGB-D SLAM},
        author={Keetha, Nikhil and Karhade, Jay and Jatavallabhula, Krishna Murthy and Yang, Gengshan and Scherer, Sebastian and Ramanan, Deva and Luiten, Jonathon},
        booktitle={Proceedings of the IEEE/CVF Conference on Computer Vision and Pattern Recognition},
        year={2024}
}

@article{MatsukiCVPR2024_monogs,
  title={{G}aussian {S}platting {SLAM}},
  author={Hidenobu Matsuki and Riku Murai and Paul H. J. Kelly and Andrew J. Davison},
  booktitle={Proceedings of the IEEE/CVF Conference on Computer Vision and Pattern Recognition},
  year={2024}
}

@inproceedings{hhuang2024photoslam,
	title = {Photo-SLAM: Real-time Simultaneous Localization and Photorealistic Mapping for Monocular, Stereo, and RGB-D Cameras},
	author = {Huang, Huajian and Li, Longwei and Cheng Hui and Yeung, Sai-Kit},
	booktitle = {Proceedings of the IEEE/CVF Conference on Computer Vision and Pattern Recognition},
	year = {2024}
}

@inproceedings{yan2023gs,
  author    = {Yan, Chi and Qu, Delin and Xu, Dan and Zhao, Bin and Wang, Zhigang and Wang, Dong and Li, Xuelong},
  title     = {GS-SLAM: Dense Visual SLAM with 3D Gaussian Splatting},
  booktitle = {CVPR},
  year      ={2024},
}

@misc{yugay2023gaussianslam,
      title={Gaussian-SLAM: Photo-realistic Dense SLAM with Gaussian Splatting}, 
      author={Vladimir Yugay and Yue Li and Theo Gevers and Martin R. Oswald},
      year={2023},
      eprint={2312.10070},
      archivePrefix={arXiv},
      primaryClass={cs.CV}
}

@article{sandstrom2024splatslam,
  title={Splat-SLAM: Globally Optimized RGB-only SLAM with 3D Gaussians},
  author={Sandstr{\"o}m, Erik and Tateno, Keisuke and Oechsle, Michael and Niemeyer, Michael and Van Gool, Luc and Oswald, Martin R and Tombari, Federico},
  journal={arXiv preprint arXiv:2405.16544},
  year={2024}
}

@inproceedings{liso2024loopyslam,
  title={Loopy-slam: Dense neural slam with loop closures},
  author={Liso, Lorenzo and Sandstr{\"o}m, Erik and Yugay, Vladimir and Van Gool, Luc and Oswald, Martin R},
  booktitle={Proceedings of the IEEE/CVF Conference on Computer Vision and Pattern Recognition},
  pages={20363--20373},
  year={2024}
}

@article{bruns2024neuralgraphmapping,
  title={Neural Graph Mapping for Dense SLAM with Efficient Loop Closure},
  author={Bruns, Leonard and Zhang, Jun and Jensfelt, Patric},
  journal={arXiv preprint arXiv:2405.03633},
  year={2024}
}

@misc{zhu2024_loopsplat,
      title={LoopSplat: Loop Closure by Registering 3D Gaussian Splats}, 
      author={Liyuan Zhu and Yue Li and Erik Sandström and Shengyu Huang and Konrad Schindler and Iro Armeni},
      year={2024},
      eprint={2408.10154},
      archivePrefix={arXiv},
      primaryClass={cs.CV}
}

@inproceedings{zhang2024gspull,
  title = {Neural Signed Distance Function Inference through Splatting 3D Gaussians Pulled on Zero-Level Set},
  author = {Wenyuan Zhang and Yu-Shen Liu and Zhizhong Han},
  booktitle = {Advances in Neural Information Processing Systems},
  year = {2024},
}

@misc{patel2024normalguideddetailpreservingneuralimplicit,
      title={Normal-guided Detail-Preserving Neural Implicit Functions for High-Fidelity 3D Surface Reconstruction}, 
      author={Aarya Patel and Hamid Laga and Ojaswa Sharma},
      year={2024},
      eprint={2406.04861},
      archivePrefix={arXiv},
      primaryClass={cs.CV},
      url={https://arxiv.org/abs/2406.04861}, 
}

@misc{xu2024grm,
      title={GRM: Large Gaussian Reconstruction Model for Efficient 3D Reconstruction and Generation}, 
      author={Xu Yinghao and Shi Zifan and Yifan Wang and Chen Hansheng and Yang Ceyuan and Peng Sida and Shen Yujun and Wetzstein Gordon},
      year={2024},
      eprint={ 2403.14621},
      archivePrefix={arXiv},
      primaryClass={cs.CV}
}

@article{gslrm2024,
    author={Zhang, Kai and Bi, Sai and Tan, Hao and Xiangli, Yuanbo and Zhao, Nanxuan 
      and Sunkavalli, Kalyan and Xu, Zexiang},
    title     = {GS-LRM: Large Reconstruction Model for 3D Gaussian Splatting},
    journal   = {European Conference on Computer Vision},
    year      = {2024},
}

@misc{gao2024meshbasedgaussiansplattingrealtime,
      title={Mesh-based Gaussian Splatting for Real-time Large-scale Deformation}, 
      author={Lin Gao and Jie Yang and Bo-Tao Zhang and Jia-Mu Sun and Yu-Jie Yuan and Hongbo Fu and Yu-Kun Lai},
      year={2024},
      eprint={2402.04796},
      archivePrefix={arXiv},
      primaryClass={cs.GR},
      url={https://arxiv.org/abs/2402.04796}, 
}

@inproceedings{luiten2023dynamic3dgs,
  title={Dynamic 3D Gaussians: Tracking by Persistent Dynamic View Synthesis},
  author={Luiten, Jonathon and Kopanas, Georgios and Leibe, Bastian and Ramanan, Deva},
  booktitle={3DV},
  year={2024}
}

@article{zakharov2024gh,
    title={Human Hair Reconstruction with Strand-Aligned 3D Gaussians},
    author={Zakharov, Egor and Sklyarova, Vanessa and Black, Michael J and Nam, Giljoo and Thies, Justus and Hilliges, Otmar},
    journal={ArXiv},
    month={Sep}, 
    year={2024} 
}

@Article{kerbl3Dgaussians,
      author       = {Kerbl, Bernhard and Kopanas, Georgios and Leimk{\"u}hler, Thomas and Drettakis, George},
      title        = {3D Gaussian Splatting for Real-Time Radiance Field Rendering},
      journal      = {ACM Transactions on Graphics},
      number       = {4},
      volume       = {42},
      month        = {July},
      year         = {2023},
      url          = {https://repo-sam.inria.fr/fungraph/3d-gaussian-splatting/}
}

@article{3dgrt2024,
    author = {Nicolas Moenne-Loccoz and Ashkan Mirzaei and Or Perel and Riccardo de Lutio and Janick Martinez Esturo and Gavriel State and Sanja Fidler and Nicholas Sharp and Zan Gojcic},
    title = {3D Gaussian Ray Tracing: Fast Tracing of Particle Scenes},
    journal = {ACM Transactions on Graphics and SIGGRAPH Asia},
    year = {2024},
}

@inproceedings{Huang2DGS2024,
    title={2D Gaussian Splatting for Geometrically Accurate Radiance Fields},
    author={Huang, Binbin and Yu, Zehao and Chen, Anpei and Geiger, Andreas and Gao, Shenghua},
    publisher = {Association for Computing Machinery},
    booktitle = {SIGGRAPH 2024 Conference Papers},
    year      = {2024},
    doi       = {10.1145/3641519.3657428}
}

@article{Yu2024GOF,
  author    = {Yu, Zehao and Sattler, Torsten and Geiger, Andreas},
  title     = {Gaussian Opacity Fields: Efficient Adaptive Surface Reconstruction in Unbounded Scenes},
  journal   = {ACM Transactions on Graphics},
  year      = {2024},
}

@article{wolf2024gs2mesh,
      title={GS2Mesh: Surface Reconstruction from Gaussian Splatting via Novel Stereo Views},
      author={Wolf, Yaniv and Bracha, Amit and Kimmel, Ron},
      journal={arXiv preprint arXiv:2404.01810},
      year={2024}
    }

@misc{chen2024pgsrplanarbasedgaussiansplatting,
      title={PGSR: Planar-based Gaussian Splatting for Efficient and High-Fidelity Surface Reconstruction}, 
      author={Danpeng Chen and Hai Li and Weicai Ye and Yifan Wang and Weijian Xie and Shangjin Zhai and Nan Wang and Haomin Liu and Hujun Bao and Guofeng Zhang},
      year={2024},
      eprint={2406.06521},
      archivePrefix={arXiv},
      primaryClass={cs.CV},
      url={https://arxiv.org/abs/2406.06521}, 
}

@article{fan2024trim,
  author    = {Fan, Lue and Yang, Yuxue and Li, Minxing and Li, Hongsheng and Zhang, Zhaoxiang},
  title     = {Trim 3D Gaussian Splatting for Accurate Geometry Representation},
  journal   = {arXiv preprint arXiv:2406.07499},
  year      = {2024},
}

@misc{lin2024directlearningmeshappearance,
      title={Direct Learning of Mesh and Appearance via 3D Gaussian Splatting}, 
      author={Ancheng Lin and Jun Li},
      year={2024},
      eprint={2405.06945},
      archivePrefix={arXiv},
      primaryClass={cs.CV},
      url={https://arxiv.org/abs/2405.06945}, 
}

@inproceedings{charatan23pixelsplat,
      title={pixelSplat: 3D Gaussian Splats from Image Pairs for Scalable Generalizable 3D Reconstruction},
      author={David Charatan and Sizhe Li and Andrea Tagliasacchi and Vincent Sitzmann},
      year={2023},
      booktitle={arXiv},
}

@inproceedings{Sandström2023ICCVpointslam,
  author    = {Sandström, Erik and Li, Yue and Van Gool, Luc and R. Oswald, Martin},
  title     = {Point-SLAM: Dense Neural Point Cloud-based SLAM},
  booktitle = {Proceedings of the IEEE/CVF International Conference on Computer Vision (ICCV)},
  year      = {2023}
}

@inproceedings{johari-et-al-2023-ESLAM,
  author = {Johari, M. M. and Carta, C. and Fleuret, F.},
  title = {{ESLAM}: Efficient Dense SLAM System Based on Hybrid Representation of Signed Distance Fields},
  booktitle = {Proceedings of the IEEE international conference on Computer Vision and Pattern Recognition (CVPR)},
  year = {2023},
  type = {Highlight}
}

@ARTICLE{1284395_ssim,
  author={Zhou Wang and Bovik, A.C. and Sheikh, H.R. and Simoncelli, E.P.},
  journal={IEEE Transactions on Image Processing}, 
  title={Image quality assessment: from error visibility to structural similarity}, 
  year={2004},
  volume={13},
  number={4},
  pages={600-612},
  doi={10.1109/TIP.2003.819861}}

@misc{zhang2018unreasonableeffectivenessdeepfeatures_lpips,
      title={The Unreasonable Effectiveness of Deep Features as a Perceptual Metric}, 
      author={Richard Zhang and Phillip Isola and Alexei A. Efros and Eli Shechtman and Oliver Wang},
      year={2018},
      eprint={1801.03924},
      archivePrefix={arXiv},
      primaryClass={cs.CV},
      url={https://arxiv.org/abs/1801.03924}, 
}

@inproceedings{yeshwanthliu2023scannetpp,
  title={ScanNet++: A High-Fidelity Dataset of 3D Indoor Scenes},
  author={Yeshwanth, Chandan and Liu, Yueh-Cheng and Nie{\ss}ner, Matthias and Dai, Angela},
  booktitle = {Proceedings of the International Conference on Computer Vision ({ICCV})},
  year={2023}
}

@inproceedings{cp-slam,
    title={CP-SLAM: Collaborative Neural Point-based SLAM System},
    author={Jiarui Hu and Mao Mao and Hujun Bao and Guofeng Zhang and Zhaopeng Cui},
    booktitle={Neural Information Processing Systems (NeurIPS)},
    year={2023}
}

@ARTICLE{9712211robot,
  author={Adamkiewicz, Michal and Chen, Timothy and Caccavale, Adam and Gardner, Rachel and Culbertson, Preston and Bohg, Jeannette and Schwager, Mac},
  journal={IEEE Robotics and Automation Letters}, 
  title={Vision-Only Robot Navigation in a Neural Radiance World}, 
  year={2022},
  volume={7},
  number={2},
  pages={4606-4613},
}

@InProceedings{Arandjelovic16,
  author       = "Arandjelovi\'c, R. and Gronat, P. and Torii, A. and Pajdla, T. and Sivic, J.",
  title        = "{NetVLAD}: {CNN} architecture for weakly supervised place recognition",
  booktitle    = "IEEE Conference on Computer Vision and Pattern Recognition",
  year         = "2016",
}

@article{hu2024cg,
    title={CG-SLAM: Efficient Dense RGB-D SLAM in a Consistent Uncertainty-aware 3D Gaussian Field},
    author={Hu, Jiarui and Chen, Xianhao and Feng, Boyin and Li, Guanglin and Yang, Liangjing and Bao, Hujun and Zhang, Guofeng and Cui, Zhaopeng},
    journal={arXiv preprint arXiv:2403.16095},
    year={2024}
}

@INPROCEEDINGS{colorptreg_odo,
  author={Park, Jaesik and Zhou, Qian-Yi and Koltun, Vladlen},
  booktitle={2017 IEEE International Conference on Computer Vision (ICCV)}, 
  title={Colored Point Cloud Registration Revisited}, 
  year={2017},
  volume={},
  number={},
  pages={143-152},
  doi={10.1109/ICCV.2017.25}}

@misc{li2024sgsslamsemanticgaussiansplatting,
      title={SGS-SLAM: Semantic Gaussian Splatting For Neural Dense SLAM}, 
      author={Mingrui Li and Shuhong Liu and Heng Zhou and Guohao Zhu and Na Cheng and Tianchen Deng and Hongyu Wang},
      year={2024},
      eprint={2402.03246},
      archivePrefix={arXiv},
      primaryClass={cs.CV},
      url={https://arxiv.org/abs/2402.03246}, 
}

@misc{wei2024gsfusiononlinergbdmapping,
    title={GSFusion: Online RGB-D Mapping Where Gaussian Splatting Meets TSDF Fusion}, 
    author={Jiaxin Wei and Stefan Leutenegger},
    year={2024},
    eprint={2408.12677},
    archivePrefix={arXiv},
    primaryClass={cs.CV},
    url={https://arxiv.org/abs/2408.12677}, 
  }

@misc{ha2024rgbdgsicpslam,
      title={RGBD GS-ICP SLAM}, 
      author={Seongbo Ha and Jiung Yeon and Hyeonwoo Yu},
      year={2024},
      eprint={2403.12550},
      archivePrefix={arXiv},
      primaryClass={cs.CV},
      url={https://arxiv.org/abs/2403.12550}, 
}

@article{slam3r,
  title={SLAM3R: Real-Time Dense Scene Reconstruction from Monocular RGB Videos},
  author={Liu, Yuzheng and Dong, Siyan and Wang, Shuzhe and Yang, Yanchao and Fan, Qingnan and Chen, Baoquan},
  journal={arXiv preprint arXiv:2412.09401},
  year={2024}
}

@misc{mast3rsfm,
      title={MASt3R-SfM: a Fully-Integrated Solution for Unconstrained Structure-from-Motion}, 
      author={Bardienus Duisterhof and Lojze Zust and Philippe Weinzaepfel and Vincent Leroy and Yohann Cabon and Jerome Revaud},
      year={2024},
      eprint={2409.19152},
      archivePrefix={arXiv},
      primaryClass={cs.CV},
      url={https://arxiv.org/abs/2409.19152}, 
}

@article{murai2024_mast3rslam,
        title={{MASt3R-SLAM}: Real-Time Dense {SLAM} with {3D} Reconstruction Priors},
        author={Murai, Riku and Dexheimer, Eric and Davison, Andrew J.},
        journal={arXiv preprint},
        year={2024},
    }

@InProceedings{wang2023vggsfm,
  author    = {Jianyuan Wang and Nikita Karaev and Christian Rupprecht and David Novotny},
  title     = {VGGSfM: Visual Geometry Grounded Deep Structure From Motion},
  year      = {2023}
}

@INPROCEEDINGS{Navier-stokesinpainting,
  author={Bertalmio, M. and Bertozzi, A.L. and Sapiro, G.},
  booktitle={Proceedings of the 2001 IEEE Computer Society Conference on Computer Vision and Pattern Recognition. CVPR 2001}, 
  title={Navier-stokes, fluid dynamics, and image and video inpainting}, 
  year={2001},
  volume={1},
  number={},
  pages={I-I},
  doi={10.1109/CVPR.2001.990497}}

@inproceedings{Segal2009GeneralizedICP,
  title={Generalized-ICP},
  author={Aleksandr V. Segal and Dirk H{\"a}hnel and Sebastian Thrun},
  booktitle={Robotics: Science and Systems},
  year={2009},
  url={https://api.semanticscholar.org/CorpusID:231748613}
}

@article{wei2024gsfusion,
  title={Gsfusion: Online rgb-d mapping where gaussian splatting meets tsdf fusion},
  author={Wei, Jiaxin and Leutenegger, Stefan},
  journal={IEEE Robotics and Automation Letters},
  year={2024},
  publisher={IEEE}
}

@InProceedings{Hu2025VTGSSLAM,
                title = {VTGaussian-SLAM: RGBD SLAM for Large Scale Scenes with Splatting View-Tied 3D Gaussians},
                author = {Hu, Pengchong and Han, Zhizhong},
                booktitle = {Proceedings of the 42nd International Conference on Machine Learning},
                year = {2025}
}

@inproceedings{szymanowicz24splatter,
      title={Splatter Image: Ultra-Fast Single-View 3D Reconstruction},
      author={Stanislaw Szymanowicz and Christian Rupprecht and Andrea Vedaldi},
      year={2024},
      booktitle={The IEEE/CVF Conference on Computer Vision and Pattern Recognition (CVPR)},
}

@article{zhang2024gaussiancube,
  title={GaussianCube: Structuring Gaussian Splatting using Optimal Transport for 3D Generative Modeling},
  author={Zhang, Bowen and Cheng, Yiji and Yang, Jiaolong and Wang, Chunyu and Zhao, Feng and Tang, Yansong and Chen, Dong and Guo, Baining},
  journal={arXiv preprint arXiv:2403.19655},
  year={2024}
}

@article{maggio2025vggtslam,
  title={VGGT-SLAM: Dense RGB SLAM Optimized on the SL (4) Manifold},
  author={Maggio, Dominic and Lim, Hyungtae and Carlone, Luca},
  journal={arXiv preprint arXiv:2505.12549},
  year={2025}
}

@InProceedings{Liu2026shortersplatting,
  title = {Speeding Up the Learning of 3D Gaussians with Much Shorter Gaussian Lists},
  author = {Liu, Jiaqi and Han, Zhizhong},
  booktitle = {Proceedings of the IEEE/CVF Conference on Computer Vision and Pattern Recognition},
  year = {2026}
}

@misc{bai2025learningcompactlatentspace,
      title={Learning Compact Latent Space for Representing Neural Signed Distance Functions with High-fidelity Geometry Details}, 
      author={Qiang Bai and Bojian Wu and Xi Yang and Zhizhong Han},
      year={2025},
      eprint={2511.14539},
      archivePrefix={arXiv},
      primaryClass={cs.CV},
      url={https://arxiv.org/abs/2511.14539}, 
}

@ARTICLE{udfstudio,
author={Zhou, Junsheng and Zhang, Weiqi and Ma, Baorui and Shi, Kanle and Liu, Yu-Shen and Han, Zhizhong},
journal={ IEEE Transactions on Pattern Analysis \& Machine Intelligence },
title={{ UDFStudio: A Unified Framework of Datasets, Benchmarks and Generative Models for Unsigned Distance Functions }},
year={5555},
volume={},
number={01},
ISSN={1939-3539},
pages={1-18},
doi={10.1109/TPAMI.2026.3668763},
url = {https://doi.ieeecomputersociety.org/10.1109/TPAMI.2026.3668763},
publisher={IEEE Computer Society},
address={Los Alamitos, CA, USA},
month=feb}

@misc{zhang2026vrpudfunbiasedlearningunsigned,
      title={VRP-UDF: Towards Unbiased Learning of Unsigned Distance Functions from Multi-view Images with Volume Rendering Priors}, 
      author={Wenyuan Zhang and Chunsheng Wang and Kanle Shi and Yu-Shen Liu and Zhizhong Han},
      year={2026},
      eprint={2407.16396},
      archivePrefix={arXiv},
      primaryClass={cs.CV},
      url={https://arxiv.org/abs/2407.16396}, 
}

@misc{zhou2025ucanunsupervisedpointcloud,
      title={U-CAN: Unsupervised Point Cloud Denoising with Consistency-Aware Noise2Noise Matching}, 
      author={Junsheng Zhou and Xingyu Shi and Haichuan Song and Yi Fang and Yu-Shen Liu and Zhizhong Han},
      year={2025},
      eprint={2510.25210},
      archivePrefix={arXiv},
      primaryClass={cs.CV},
      url={https://arxiv.org/abs/2510.25210}, 
}

@misc{zhang2025materialrefgsreflectivegaussiansplatting,
      title={MaterialRefGS: Reflective Gaussian Splatting with Multi-view Consistent Material Inference}, 
      author={Wenyuan Zhang and Jimin Tang and Weiqi Zhang and Yi Fang and Yu-Shen Liu and Zhizhong Han},
      year={2025},
      eprint={2510.11387},
      archivePrefix={arXiv},
      primaryClass={cs.CV},
      url={https://arxiv.org/abs/2510.11387}, 
}

@misc{han2025sparsereconneuralimplicitsurface,
      title={SparseRecon: Neural Implicit Surface Reconstruction from Sparse Views with Feature and Depth Consistencies}, 
      author={Liang Han and Xu Zhang and Haichuan Song and Kanle Shi and Yu-Shen Liu and Zhizhong Han},
      year={2025},
      eprint={2508.00366},
      archivePrefix={arXiv},
      primaryClass={cs.CV},
      url={https://arxiv.org/abs/2508.00366}, 
}

@misc{noda2025learningbijectivesurfaceparameterization,
      title={Learning Bijective Surface Parameterization for Inferring Signed Distance Functions from Sparse Point Clouds with Grid Deformation}, 
      author={Takeshi Noda and Chao Chen and Junsheng Zhou and Weiqi Zhang and Yu-Shen Liu and Zhizhong Han},
      year={2025},
      eprint={2503.23670},
      archivePrefix={arXiv},
      primaryClass={cs.CV},
      url={https://arxiv.org/abs/2503.23670}, 
}

@misc{li2025gaussianudfinferringunsigneddistance,
      title={GaussianUDF: Inferring Unsigned Distance Functions through 3D Gaussian Splatting}, 
      author={Shujuan Li and Yu-Shen Liu and Zhizhong Han},
      year={2025},
      eprint={2503.19458},
      archivePrefix={arXiv},
      primaryClass={cs.CV},
      url={https://arxiv.org/abs/2503.19458}, 
}

@misc{zhang2025nerfpriorlearningneuralradiance,
      title={NeRFPrior: Learning Neural Radiance Field as a Prior for Indoor Scene Reconstruction}, 
      author={Wenyuan Zhang and Emily Yue-ting Jia and Junsheng Zhou and Baorui Ma and Kanle Shi and Yu-Shen Liu and Zhizhong Han},
      year={2025},
      eprint={2503.18361},
      archivePrefix={arXiv},
      primaryClass={cs.CV},
      url={https://arxiv.org/abs/2503.18361}, 
}

@misc{zhang2025monoinstanceenhancingmonocularpriors,
      title={MonoInstance: Enhancing Monocular Priors via Multi-view Instance Alignment for Neural Rendering and Reconstruction}, 
      author={Wenyuan Zhang and Yixiao Yang and Han Huang and Liang Han and Kanle Shi and Yu-Shen Liu and Zhizhong Han},
      year={2025},
      eprint={2503.18363},
      archivePrefix={arXiv},
      primaryClass={cs.CV},
      url={https://arxiv.org/abs/2503.18363}, 
}

@misc{jiang2024sensingsurfacepatchesvolume,
      title={Sensing Surface Patches in Volume Rendering for Inferring Signed Distance Functions}, 
      author={Sijia Jiang and Tong Wu and Jing Hua and Zhizhong Han},
      year={2024},
      eprint={2412.16467},
      archivePrefix={arXiv},
      primaryClass={cs.CV},
      url={https://arxiv.org/abs/2412.16467}, 
}

@misc{chen2024learninglocalpatternmodularization,
      title={Learning Local Pattern Modularization for Point Cloud Reconstruction from Unseen Classes}, 
      author={Chao Chen and Yu-Shen Liu and Zhizhong Han},
      year={2024},
      eprint={2408.14279},
      archivePrefix={arXiv},
      primaryClass={cs.CV},
      url={https://arxiv.org/abs/2408.14279}, 
}

@inproceedings{zhang2024learning,
      title={Learning unsigned distance functions from multi-view images with volume rendering priors},
      author={Zhang, Wenyuan and Shi, Kanle and Liu, Yu-Shen and Han, Zhizhong},
      booktitle={European Conference on Computer Vision},
      pages={397--415},
      year={2024},
      organization={Springer}
    }

@misc{jiang2024queryquantizedneuralslam,
      title={Query Quantized Neural SLAM}, 
      author={Sijia Jiang and Jing Hua and Zhizhong Han},
      year={2024},
      eprint={2412.16476},
      archivePrefix={arXiv},
      primaryClass={cs.CV},
      url={https://arxiv.org/abs/2412.16476}, 
}

@misc{chen2024sharpeningneuralimplicitfunctions,
      title={Sharpening Neural Implicit Functions with Frequency Consolidation Priors}, 
      author={Chao Chen and Yu-Shen Liu and Zhizhong Han},
      year={2024},
      eprint={2412.19720},
      archivePrefix={arXiv},
      primaryClass={cs.CV},
      url={https://arxiv.org/abs/2412.19720}, 
}

@misc{zhou2024zeroshotscenereconstructionsingle,
      title={Zero-Shot Scene Reconstruction from Single Images with Deep Prior Assembly}, 
      author={Junsheng Zhou and Yu-Shen Liu and Zhizhong Han},
      year={2024},
      eprint={2410.15971},
      archivePrefix={arXiv},
      primaryClass={cs.CV},
      url={https://arxiv.org/abs/2410.15971}, 
}

@misc{han2024binocularguided3dgaussiansplatting,
      title={Binocular-Guided 3D Gaussian Splatting with View Consistency for Sparse View Synthesis}, 
      author={Liang Han and Junsheng Zhou and Yu-Shen Liu and Zhizhong Han},
      year={2024},
      eprint={2410.18822},
      archivePrefix={arXiv},
      primaryClass={cs.CV},
      url={https://arxiv.org/abs/2410.18822}, 
}

@misc{chen2024inferringneuralsigneddistance,
      title={Inferring Neural Signed Distance Functions by Overfitting on Single Noisy Point Clouds through Finetuning Data-Driven based Priors}, 
      author={Chao Chen and Yu-Shen Liu and Zhizhong Han},
      year={2024},
      eprint={2410.19680},
      archivePrefix={arXiv},
      primaryClass={cs.CV},
      url={https://arxiv.org/abs/2410.19680}, 
}

@misc{noda2024multipulldetailingsigneddistance,
      title={MultiPull: Detailing Signed Distance Functions by Pulling Multi-Level Queries at Multi-Step}, 
      author={Takeshi Noda and Chao Chen and Weiqi Zhang and Xinhai Liu and Yu-Shen Liu and Zhizhong Han},
      year={2024},
      eprint={2411.01208},
      archivePrefix={arXiv},
      primaryClass={cs.CV},
      url={https://arxiv.org/abs/2411.01208}, 
}

@misc{pak2025g2sicpslamgeometryawaregaussian,
      title={G2S-ICP SLAM: Geometry-aware Gaussian Splatting ICP SLAM}, 
      author={Gyuhyeon Pak and Hae Min Cho and Euntai Kim},
      year={2025},
      eprint={2507.18344},
      archivePrefix={arXiv},
      primaryClass={cs.RO},
      url={https://arxiv.org/abs/2507.18344}, 
}
}

\newpage

% \clearpage
\twocolumn[
  \begin{center}
    {\Large \bf Supplementary Material for SGAD-SLAM: Splatting Gaussians at Adjusted Depth for Better Radiance Fields in RGBD SLAM \par}
    \vspace{1.5em}
    {\large Pengchong Hu \hspace{3em} Zhizhong Han \par}
    % \vspace{1em}
    {\large Machine Perception Lab, Wayne State University, Detroit, USA \par}
    {\tt\small pchu@wayne.edu, h312h@wayne.edu \par}
    \vspace{2.5em}
  \end{center}
]

\setcounter{section}{0}

This supplementary material will include more details on the implementation and numerical results for each scene. Additionally, we also include more visual results.

\section{Further Implementation Details}
We implemented SGAD-SLAM in Python using the
PyTorch framework, and ran all experiments on NVIDIA
RTX4090 GPUs. In the mapping process, the number of mapping iterations is set to 100 for Replica~\cite{replica}, TUM-RGBD~\cite{6385773ATERMSE_tumrgbd}, and ScanNet~\cite{scannet}, 500 for ScanNet++~\cite{yeshwanthliu2023scannetpp}. During tracking, the downsampling ratio $R$ is set to 10 for Replica~\cite{replica} and ScanNet++~\cite{yeshwanthliu2023scannetpp}, 5 for TUM-RGBD~\cite{6385773ATERMSE_tumrgbd} and ScanNet~\cite{scannet}. When calculating the covariance matrix $c_i^j$  of a 3D point $d_i^j$ from its $K_c$ nearest neighbors, we choose $K_c = 10$ for Replica~\cite{replica} and ScanNet++~\cite{yeshwanthliu2023scannetpp}, $K_c = 23$ for TUM-RGBD~\cite{6385773ATERMSE_tumrgbd}, and $K_c = 15$ for ScanNet~\cite{scannet}. To balance each term in the loss function Eq.~1 in the main paper, 
% ~\ref{Eq:mappingrendering1}, 
we set $\rho = 0.8$, $\tau = 0.2$, and $\sigma=1.0$ for all datasets. At the beginning of mapping, we need to initialize Gaussians from the input RGBD images. The radius of each Gaussian, $r$, is initialized by utilizing the following formula as introduced in~\cite{keetha2024splatam}:
\begin{equation}
    r = \frac{D_{gt}}{F},
\end{equation}
where $D_{gt}$ is the ground-truth depth, and $F$ is the focal length. We set the learning rates as follows: $lr_{color} = 0.0025$ for color, $lr_{radius}=0.005$ for radius, $lr_{opacity}=0.05$ for opacity, and $lr_{offset}=0.01$ for offset. During mapping, at least 40$\%$ of the mapping iterations will be spent on the current view. In camera tracking, we initialize camera pose for Generalized ICP (GICP)~\cite{Segal2009GeneralizedICP} by using the constant speed assumption on ScanNet~\cite{scannet}. Due to the high quality of RGBD images and smooth motion of the camera, we adopt the latest camera initialization strategy on Replica~\cite{replica} and TUM-RGBD~\cite{6385773ATERMSE_tumrgbd}. Note that ScanNet++~\cite{yeshwanthliu2023scannetpp} is not specifically designed for SLAM tasks, and its DSLR-captured sequences include occasional sudden large motions. Therefore, following previous methods~\cite{yugay2023gaussianslam,zhu2024_loopsplat}, we utilize multi-scale RGBD odometry~\cite{colorptreg_odo} to help the pose initialization and only employ the first 250 frames of each scene in evaluations, which present smoother trajectories.

\section{More Results}
In this section, to demonstrate the state-of-the-art performance of our method, we report more detailed results on each scene on Replica~\cite{replica}, TUM-RGBD~\cite{6385773ATERMSE_tumrgbd}, ScanNet~\cite{scannet}, and ScanNet++~\cite{yeshwanthliu2023scannetpp}.

\noindent\textbf{Datasets. }Replica~\cite{replica} is a synthetic dataset that provides high-fidelity 3D reconstructions of indoor scenes. For evaluation, we utilize the widely used RGB-D sequences from eight scenes captured by Sucar~\cite{sucar2021imap}, which includes precise ground-truth trajectories. TUM-RGBD~\cite{6385773ATERMSE_tumrgbd}, ScanNet~\cite{scannet}, and ScanNet++~\cite{yeshwanthliu2023scannetpp} are real-world datasets, offering diverse and challenging environments for evaluations. The poses in TUM-RGBD were captured using an external motion capture system, whereas ScanNet derives its poses from BundleFusion~\cite{dai2017bundlefusion}. Additionally, ScanNet++ employs laser scanning to register images and obtain accurate camera poses. Similar to previous methods~\cite{keetha2024splatam,yugay2023gaussianslam,MatsukiCVPR2024_monogs,zhu2024_loopsplat,wei2024gsfusiononlinergbdmapping,li2024sgsslamsemanticgaussiansplatting}, to evaluate the superior performance of our method, we conduct experiments on eight scenes from Replica~\cite{replica}, three scenes from TUM-RGBD~\cite{6385773ATERMSE_tumrgbd}, six scenes from ScanNet~\cite{scannet}, and five scenes from ScanNet++~\cite{yeshwanthliu2023scannetpp} ((a) \texttt{b20a261fdf}, (b) \texttt{8b5caf3398}, (c) \texttt{fb05e13ad1}, (d) \texttt{2e74812d00}, (e) \texttt{281bc17764}).

\begin{figure*}
  \centering
% \vspace{-0.4in}
   \includegraphics[width=\linewidth]{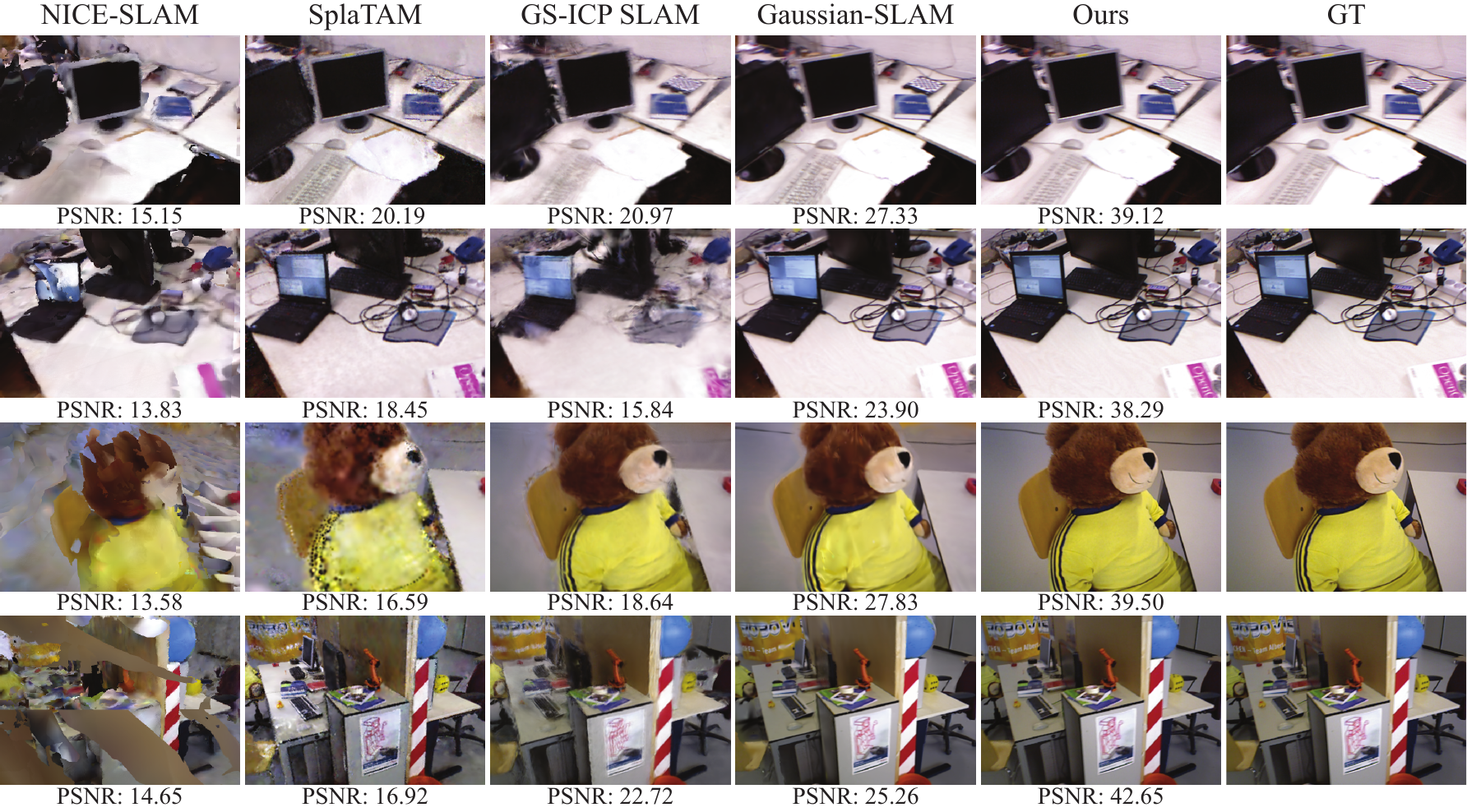}
   % \vspace{-0.3in}
   \caption{Visual comparisons in rendering on TUM-RGBD~\cite{6385773ATERMSE_tumrgbd}. Please watch our video for more comparisons in rendering.}
   \label{fig:suppmapping_tum}
% \vspace{-0.2in}
\end{figure*}

\begin{figure*}
  \centering
% \vspace{-0.4in}
   \includegraphics[width=\linewidth]{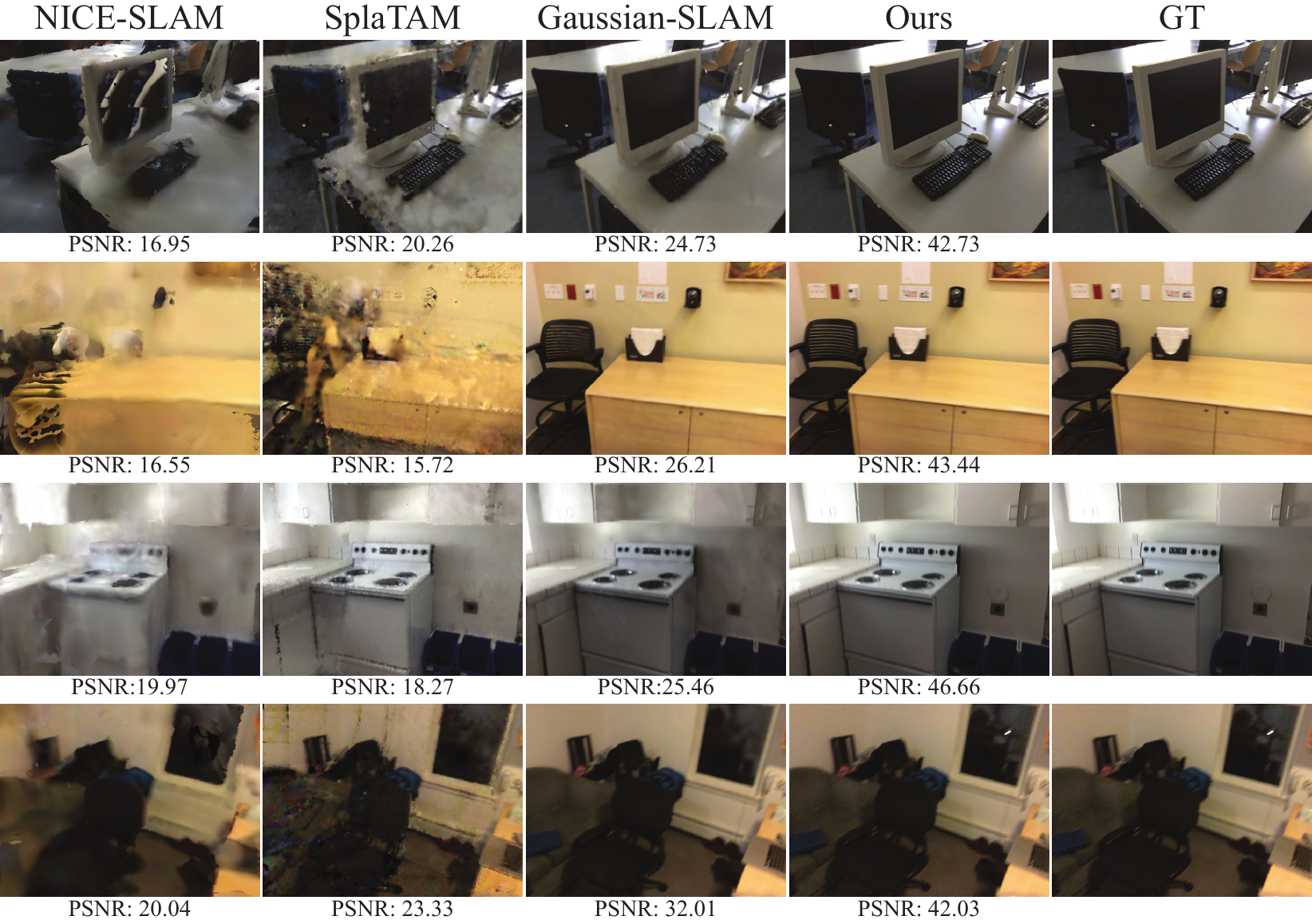}
   % \vspace{-0.2in}
   \caption{Visual comparisons in rendering on ScanNet~\cite{scannet}. Please watch our video for more comparisons in rendering.}
   \label{fig:suppmapping_scannet}
% \vspace{-0.2in}
\end{figure*}

\begin{figure*}
  \centering
% \vspace{-0.4in}
   \includegraphics[width=\linewidth]{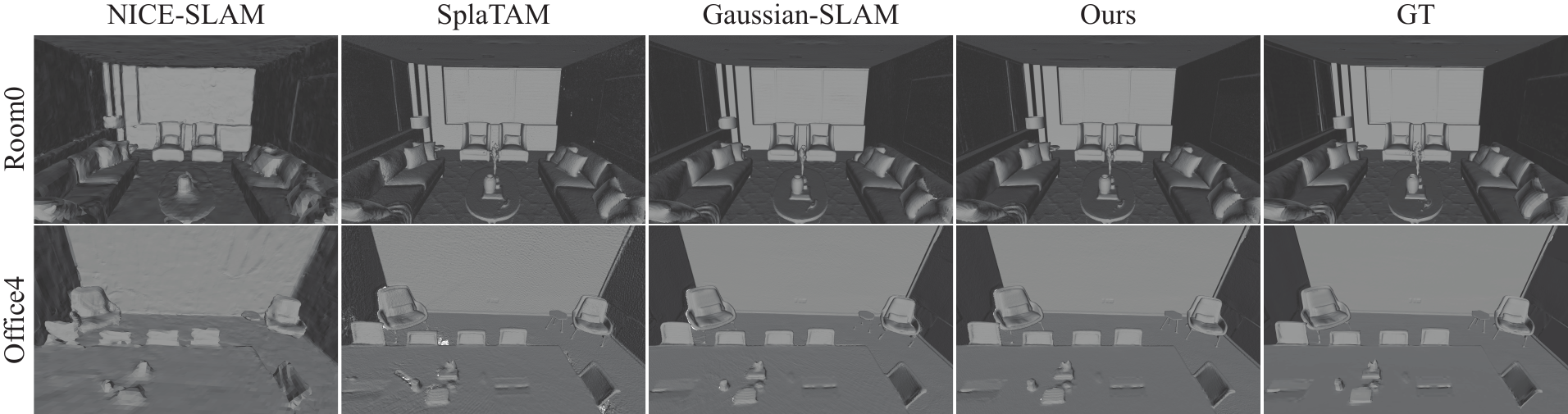}
   % \vspace{-0.2in}
   \caption{Visual comparisons in reconstruction on Replica~\cite{replica}.}
   \label{fig:suppreplicameshvis}
% \vspace{-0.2in}
\end{figure*}

\begin{figure*}
  \centering
% \vspace{-0.4in}
   \includegraphics[width=\linewidth]{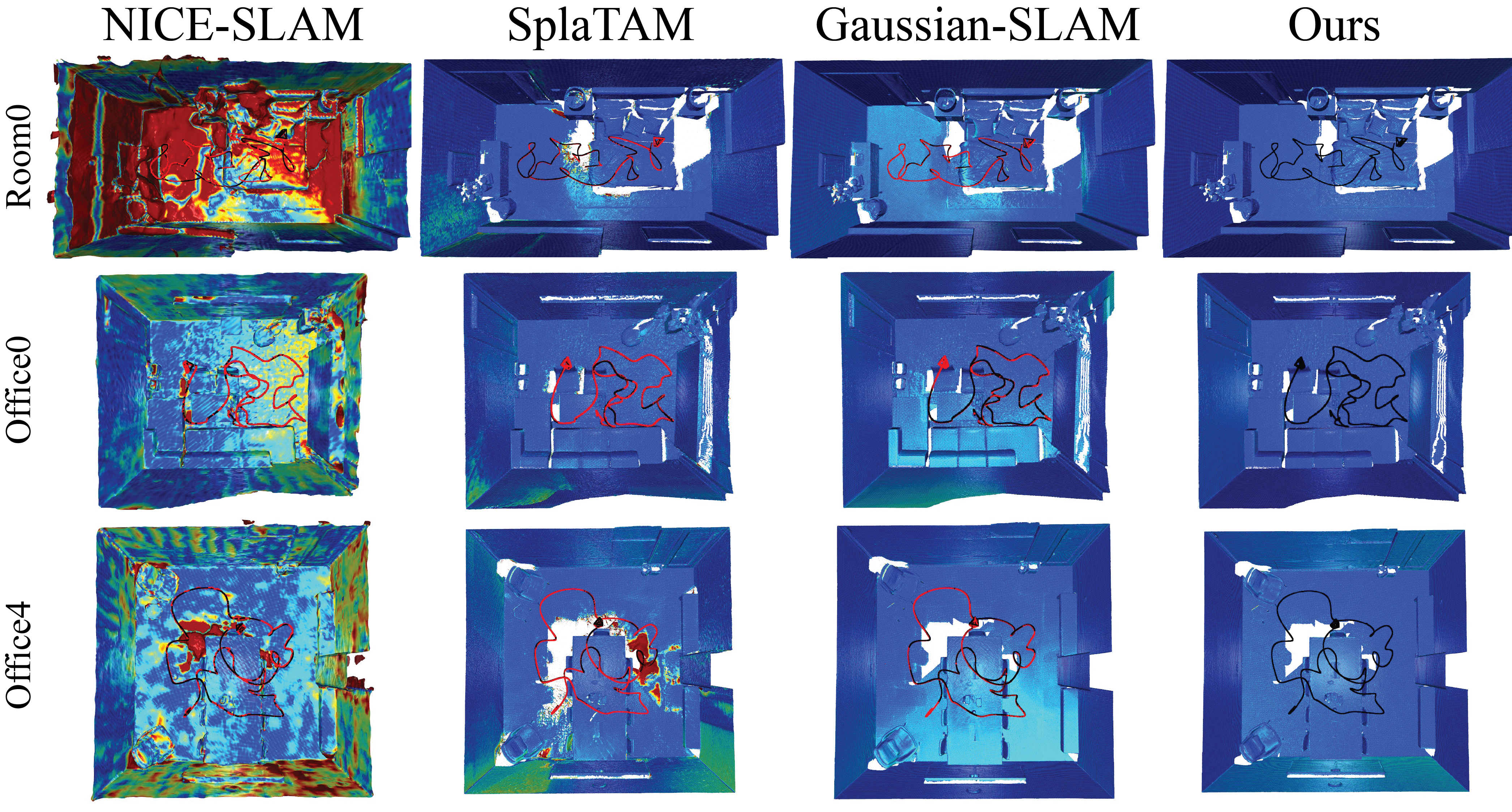}
   % \vspace{-0.2in}
   \caption{Visual comparisons in reconstruction and camera tracking on Replica~\cite{replica}. Please watch our video for more comprehensive results.}
   \label{fig:suppreplicameshviserrormap}
% \vspace{-0.2in}
\end{figure*}

\begin{figure*}
  \centering
% \vspace{-0.4in}
   \includegraphics[width=\linewidth]{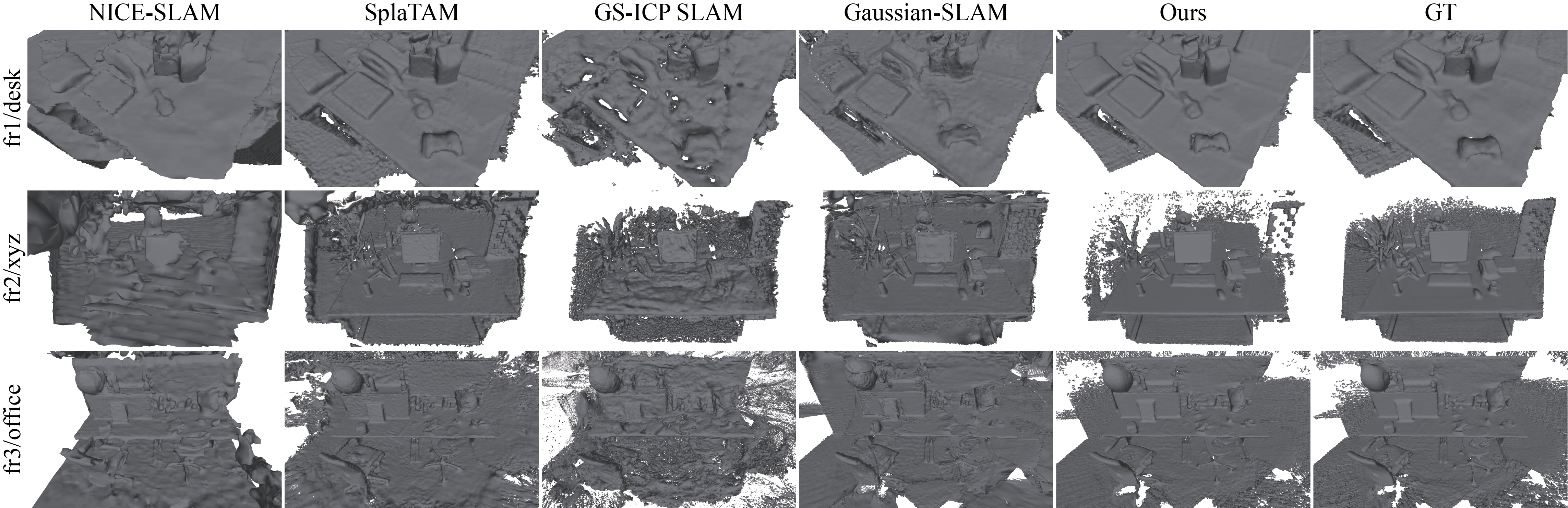}
   % \vspace{-0.25in}
   \caption{Visual comparisons in reconstruction on TUM-RGBD~\cite{6385773ATERMSE_tumrgbd}.}
   \label{fig:supptummeshvis}
% \vspace{-0.15in}
\end{figure*}

\noindent\textbf{Metrics.}
We evaluate both the accuracy of estimated camera poses at each frame and the rendering quality from both observed and unobserved viewpoints. To evaluate tracking performance, we employ the root mean square absolute trajectory error (ATE RMSE)~\cite{6385773ATERMSE_tumrgbd} as a metric. To assess rendering performance, we measure PSNR, SSIM~\cite{1284395_ssim}, and LPIPS~\cite{zhang2018unreasonableeffectivenessdeepfeatures_lpips}. Consistent with prior works~\cite{Sandström2023ICCVpointslam, liso2024loopyslam, zhu2024_loopsplat, yugay2023gaussianslam}, all rendering metrics are computed by rendering full-resolution images along the estimated trajectory. Additionally, we reconstruct scene meshes using the marching cubes algorithm~\cite{Lorensen87marchingcubes}, following the approach in \cite{Sandström2023ICCVpointslam}. The reconstruction quality is assessed using the F1-score, the harmonic mean of Precision (P) and Recall (R), with a distance threshold of 1 cm for all evaluations. Furthermore, we employ the depth L1 to evaluate the rendered mesh depth error at sampled novel views as in~\cite{Zhu2021NICESLAM}. 

\noindent\textbf{Numerical Results. }To ensure statistical significance, all reported numerical results are averaged over five runs. For camera tracking performance, per-scene results are reported in Tab.~\ref{Tab:suppcamreplicaperscene} - Tab.~\ref{Tab:suppcamscannetppperscene} for Replica~\cite{replica}, TUM-RGBD~\cite{6385773ATERMSE_tumrgbd}, ScanNet~\cite{scannet}, and ScanNet++~\cite{yeshwanthliu2023scannetpp}. We present comparisons in rendering performance for each scene in Replica~\cite{replica} in Tab.~\ref{Tab:supprenderreplicaperscene}, in TUM-RGBD~\cite{6385773ATERMSE_tumrgbd} in Tab.~\ref{Tab:supprendertumperscene}, in ScanNet~\cite{scannet} in Tab.~\ref{Tab:supprenderscannetperscene}, and ScanNet++~\cite{yeshwanthliu2023scannetpp} in Tab.~\ref{Tab:supprenderscannetppperscene}. Additionally, we report the novel view synthesis (NVS) results on ScanNet++, where the test views are distant from training views. To evaluate PSNR on each novel view, we follow previous methods~\cite{yugay2023gaussianslam,zhu2024_loopsplat} to finetune the merged global map with 10$K$ iterations and obtain the renderings in novel views.
Our per scene results in Tab.~\ref{Tab:suppnvsrenderscannetppperscene} show that our method yields the best NVS performance.

\begin{figure*}
  \centering
% \vspace{-0.4in}
   \includegraphics[width=\linewidth]{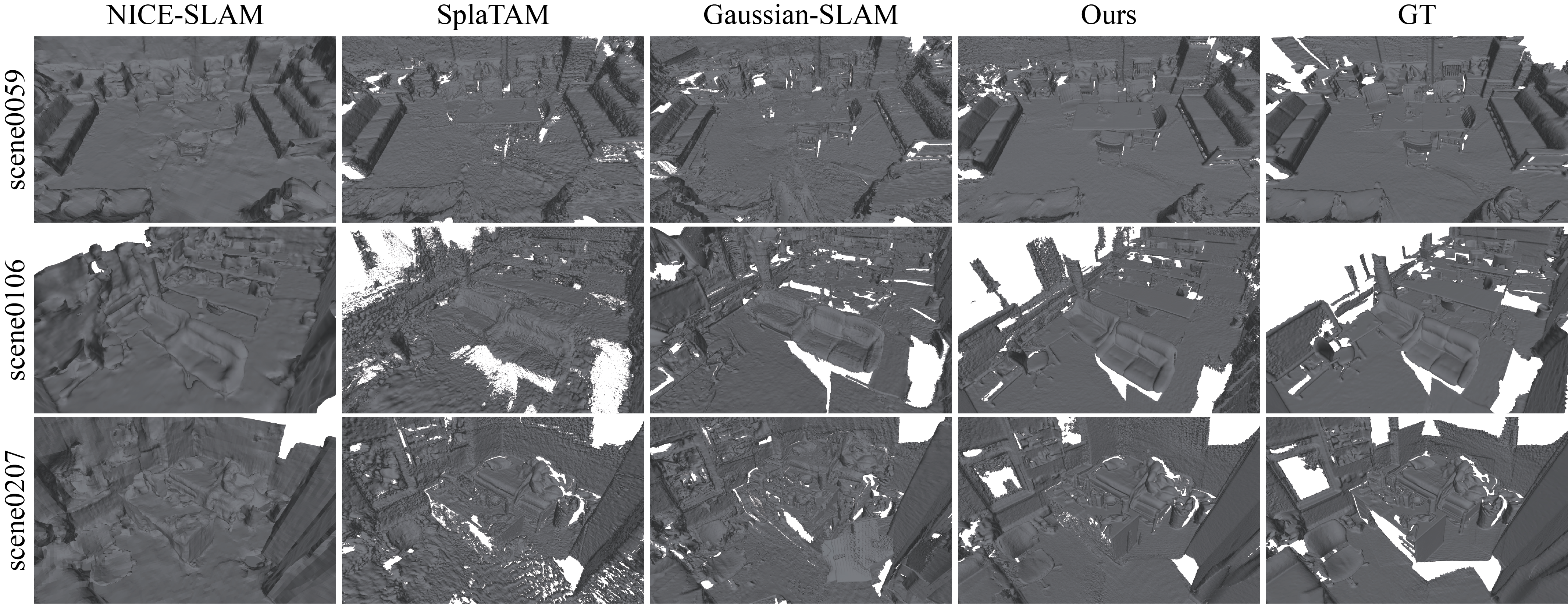}
   % \vspace{-0.25in}
   \caption{Visual comparisons in reconstruction on ScanNet~\cite{scannet}.}
   \label{fig:suppscannetmeshvis}
% \vspace{-0.15in}
\end{figure*}

\begin{figure*}
  \centering
% \vspace{-0.4in}
   \includegraphics[width=\linewidth]{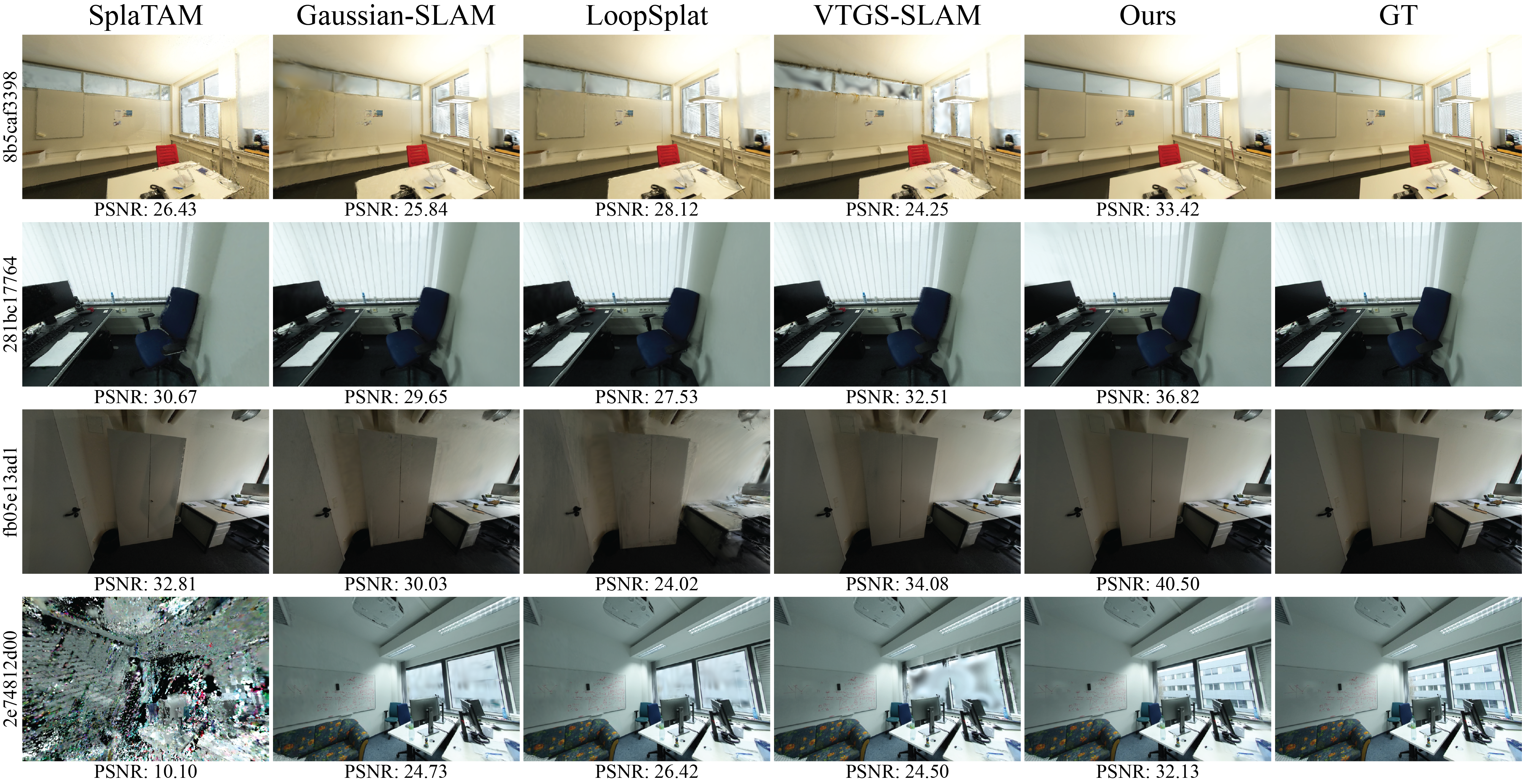}
   % \vspace{-0.2in}
   \caption{Visual comparisons in training views rendering on ScanNet++~\cite{yeshwanthliu2023scannetpp}. Please watch our video for more comparisons in rendering.}
   \label{fig:suppmapping_scannetpp}
% \vspace{-0.2in}
\end{figure*}

\noindent\textbf{Visual Results. }Moreover, we provide more visual comparisons in rendering and reconstruction. In Replica~\cite{replica}, we report reconstruction visual comparisons in Fig.~\ref{fig:suppreplicameshvis}, and present each scene reconstruction comparisons in Tab.~\ref{Tab:suppreconreplicaperscene}. Meanwhile, we show a visual comparison in scene reconstruction with camera tracking in Fig.~\ref{fig:suppreplicameshviserrormap} for Replica, where the error map is obtained from the reconstructed mesh for better visualization. Here we employ depth L1 and F1-score as metrics to evaluate the mesh obtained by marching cubes~\cite{Lorensen87marchingcubes} following \cite{Sandström2023ICCVpointslam}. The comparisons show that our method can acquire more accurate reconstruction, although Point-SLAM~\cite{Sandström2023ICCVpointslam} requires ground truth depth images as input
to guide sampling when rendering, which is an unfair
advantage over other NeRF-based methods. In TUM-RGBD~\cite{6385773ATERMSE_tumrgbd}, we provide more rendering and reconstruction results in Fig.~\ref{fig:suppmapping_tum} and Fig.~\ref{fig:supptummeshvis} separately. Compared to the latest methods~\cite{Zhu2021NICESLAM,keetha2024splatam,ha2024rgbdgsicpslam,yugay2023gaussianslam,Hu2025VTGSSLAM}, our method shows superior rendering performance and reconstruction quality. In addition, we also present more rendering results and reconstruction results for ScanNet~\cite{scannet} in Fig.~\ref{fig:suppmapping_scannet} and Fig.~\ref{fig:suppscannetmeshvis}, and more rendering results for ScanNet++~\cite{yeshwanthliu2023scannetpp} in Fig.~\ref{fig:suppmapping_scannetpp}. All of these visual comparisons clearly show our high fidelity rendering.

\noindent\textbf{Compared to VTGS-SLAM. }Since VTGS-SLAM~\cite{Hu2025VTGSSLAM} still relies on rendering for tracking, it requires complex strategies to balance the memory limit and the number of Gaussians, such as splitting a video into sections, adding dense Gaussians merely on the first frame in a section while incrementally adding sparse Gaussians on the others, and adopting different tracking strategies in the same section. Instead, we do not need section splitting, can use extremely dense Gaussians on each frame, and run appearance mapping on multiple frames in parallel, leading to much simpler tracking, better appearance modeling, and faster geometry mapping. Tab.~\ref{tab:tracking_tum_vtgs} shows VTGS-SLAM fails in tracking with dense Gaussians on each frame like us.

\noindent\textbf{Performance in Structureless Environments. }We report an evaluation on \texttt{fr3/nostructure\_texture\_far} in Tab.~\ref{tab:tracking_fr3}, which is a scene with minimal geometric structure. Our point-to-surface metric and rendering-based initialization contribute to our superior performance in this challenging scenario.

\section{More Analysis.} Our pixel-aligned Gaussians can enhance our ability to handle large scenes and the efficiency during mapping. For each frame, we only need to splat its Gaussians with the ones associated to its $N$ neighboring frames, rather than all Gaussians in the scene as in previous methods~\cite{sandstrom2024splatslam,yugay2023gaussianslam,zhu2024_loopsplat}. This design not only saves time on every rendering by reducing the number of Gaussians, but also enables us to speed up the mapping by splatting Gaussians in each frame in parallel in some scenarios. Additionally, unlike previous methods that require a set of keyframes to maintain the Gaussians' consistency to all previous frames, we do not need to minimize rendering errors to all keyframes, but just the latest frame as shown in Eq.~1 in the main paper. 

\section{Limitations}
In real-world applications, the high-quality depth image are often difficult to acquire, which increases the time cost in the mapping process and degrades the rendering and tracking performance, although our movable pixel-aligned Gaussians can mitigate the effect of noisy depth images.

\section{Code}
Please refer to our project page for code at \href{https://machineperceptionlab.github.io/SGAD-SLAM-Project/}{https://machineperceptionlab.github.io/SGAD-SLAM-Project}.

\section{Video}
Our accompanying video provides additional visualizations, including more comprehensive visual comparisons. Please watch our video for more details.

\begin{table*}
\caption{Rendering performance comparisons in PSNR $\uparrow$, SSIM $\uparrow$, and LPIPS $\downarrow$ on Replica~\cite{replica}. $*$ indicates methods relying on pre-trained data-driven priors.}
\vskip 0.05in
  \centering
  \resizebox{0.8\linewidth}{!}{
\begin{tabular}{llccccccccc}
\toprule
Method & Metric & \texttt{Rm0} & \texttt{Rm1} & \texttt{Rm2} & \texttt{Off0} & \texttt{Off1} & \texttt{Off2} & \texttt{Off3} & \texttt{Off4} & Avg. \\
\midrule
\rowcolor{gray!20}
\multicolumn{11}{l}{\textit{Neural Implicit Fields}} \\
\multirow{3}{*}{NICE-SLAM~\cite{Zhu2021NICESLAM}} 
& PSNR$\uparrow$  & 22.12 & 22.47 & 24.52 & 29.07 & 30.34 & 19.66 & 22.23 & 24.94 & 24.42 \\
& SSIM$\uparrow$  & 0.689 & 0.757 & 0.814 & 0.874 & 0.886 & 0.797 & 0.801 & 0.856 & 0.809 \\
& LPIPS$\downarrow$  & 0.330 & 0.271 & 0.208 & 0.229 & 0.181 & 0.235 & 0.209 & 0.198 & 0.233 \\
\midrule
\multirow{3}{*}{Vox-Fusion~\cite{Yang_Li_Zhai_Ming_Liu_Zhang_2022_voxfusion}} 
& PSNR$\uparrow$  & 22.39 & 22.36 & 23.92 & 27.79 & 29.83 & 20.33 & 23.47 & 25.21 & 24.41 \\
& SSIM$\uparrow$  & 0.683 & 0.751 & 0.798 & 0.857 & 0.876 & 0.794 & 0.803 & 0.847 & 0.801 \\
& LPIPS$\downarrow$  & 0.303 & 0.269 & 0.234 & 0.241 & 0.184 & 0.243 & 0.213 & 0.199 & 0.236 \\
\midrule
\multirow{3}{*}{ESLAM~\cite{johari-et-al-2023-ESLAM}} 
& PSNR$\uparrow$  & 25.25 & 27.39 & 28.09 & 30.33 & 27.04 & 27.99 & 29.27 & 29.15 & 28.06 \\
& SSIM$\uparrow$  & 0.874 & 0.89 & 0.935 & 0.934 & 0.910 & 0.942 & 0.953 & 0.948 & 0.923 \\
& LPIPS$\downarrow$  & 0.315 & 0.296 & 0.245 & 0.213 & 0.254 & 0.238 & 0.186 & 0.210 & 0.245 \\
\midrule
\multirow{3}{*}{Point-SLAM~\cite{Sandström2023ICCVpointslam}} 
& PSNR$\uparrow$  & 32.40 & 34.08 & 35.50 & 38.26 & 39.16 & 33.99 & 33.48 & 33.49 & 35.17 \\
& SSIM$\uparrow$  & 0.974 & 0.977 & 0.982 & 0.983 & 0.986 & 0.960 & 0.960 & 0.979 & 0.975 \\
& LPIPS$\downarrow$  & 0.113 & 0.116 & 0.111 & 0.100 & 0.118 & 0.156 & 0.132 & 0.142 & 0.124 \\
\hdashline
\multirow{3}{*}{Loopy-SLAM$*$~\cite{liso2024loopyslam}} 
& PSNR$\uparrow$  & - & - & - & - & - & - & - & - & 35.47  \\
& SSIM$\uparrow$  & - & - & - & - & - & - & - & - & 0.981  \\
& LPIPS$\downarrow$  & - & - & - & - & - & - & - & - & 0.109 \\
\midrule
\midrule
\rowcolor{gray!20}
\multicolumn{11}{l}{\textit{3D Gaussian Splatting}} \\
\multirow{3}{*}{SplaTAM~\cite{keetha2024splatam}} 
& PSNR$\uparrow$  & 32.86 & 33.89 & 35.25 & 38.26 & 39.17 & 31.97 & 29.70 & 31.81 & 34.11 \\
& SSIM$\uparrow$  & 0.98 & 0.97 & 0.98 & 0.98 & 0.98 & 0.97 & 0.95 & 0.95 & 0.97 \\
& LPIPS$\downarrow$  & 0.07 & 0.10 & 0.08 & 0.09 & 0.09 & 0.10 & 0.12 & 0.15 & 0.10 \\
\midrule
\multirow{3}{*}{SGS-SLAM~\cite{li2024sgsslamsemanticgaussiansplatting}} 
& PSNR$\uparrow$  & 32.50 & 34.25 & 35.10 & 38.54 & 39.20 & 32.90 & 32.05 & 32.75 & 34.66 \\
& SSIM$\uparrow$  & 0.976 & 0.978 & 0.981 & 0.984 & 0.980 & 0.967 & 0.966 & 0.949 & 0.973 \\
& LPIPS$\downarrow$  & 0.070 & 0.094 & 0.070 & 0.086 & 0.087 & 0.101 & 0.115 & 0.148 & 0.096 \\
\midrule
\multirow{3}{*}{GS-SLAM~\cite{yan2023gs}} 
& PSNR$\uparrow$  & 31.56 & 32.86 & 32.59 & 38.70 & 41.17 & 32.36 & 32.03 & 32.92 & 34.27 \\
& SSIM$\uparrow$  & 0.968 & 0.973 & 0.971 & 0.986 & 0.993 & 0.978 & 0.970 & 0.968 & 0.975 \\
& LPIPS$\downarrow$  & 0.094 & 0.075 & 0.093 & 0.050 & 0.033 & 0.094 & 0.110 & 0.112 & 0.082 \\
\midrule
\multirow{3}{*}{MonoGS~\cite{MatsukiCVPR2024_monogs}} 
& PSNR$\uparrow$  & 34.83 & 36.43 & 37.49 & 39.95 & 42.09 & 36.24 & 36.70 & 36.07 & 37.50 \\
& SSIM$\uparrow$  & 0.954 & 0.959 & 0.965 & 0.971 & 0.977 & 0.964 & 0.963 & 0.957 & 0.960 \\
& LPIPS$\downarrow$  & 0.068 & 0.076 & 0.075 & 0.072 & 0.055 & 0.078 & 0.065 & 0.099 & 0.070 \\
\midrule
\multirow{3}{*}{Gaussian-SLAM~\cite{yugay2023gaussianslam}} 
& PSNR$\uparrow$ & 38.88 & 41.80 & 42.44 & 46.40 & 45.29 & 40.10 & 39.06 & 42.65 & 42.08 \\
& SSIM$\uparrow$  & 0.993 & 0.996 & 0.996 & 0.998 & 0.997 & 0.997 & 0.997 & 0.997 & 0.996 \\
& LPIPS$\downarrow$  & 0.017 & 0.018 & 0.019 & 0.015 & 0.016 & 0.020 & 0.020 & 0.020 & 0.018 \\
\midrule
\multirow{3}{*}{VTGS-SLAM~\cite{Hu2025VTGSSLAM}} 
& PSNR$\uparrow$  & 39.95 & 43.06 & 43.13 & 46.88 & 47.20 & 42.14 & 40.99 & 43.35 & 43.34 \\
& SSIM$\uparrow$  & 0.992 & 0.996 & 0.996 & 0.998 & 0.997 & 0.996 & 0.996 & 0.996 & 0.996 \\
& LPIPS$\downarrow$ & \textbf{0.014} & \textbf{0.013} & \textbf{0.014} & \textbf{0.009} & \textbf{0.009} & \textbf{0.012} & \textbf{0.013} & \textbf{0.015} & \textbf{0.012} \\
\hdashline
\multirow{3}{*}{LoopSplat$*$~\cite{zhu2024_loopsplat}} 
& PSNR$\uparrow$  & 33.07 & 35.32 & 36.16 & 40.82 & 40.21 & 34.67 & 35.67 & 37.10 & 36.63 \\
& SSIM$\uparrow$  & 0.973 & 0.978 & 0.985 & 0.992 & 0.990 & 0.985 & 0.990 & 0.989 & 0.985 \\
& LPIPS$\downarrow$  & 0.116 & 0.122 & 0.111 & 0.085 & 0.123 & 0.140 & 0.096 & 0.106 & 0.112 \\
\hdashline
\multirow{3}{*}{CG-SLAM$*$~\cite{hu2024cg}} 
& PSNR$\uparrow$  & 33.27 & - & - & - & - & - & 34.60 & - & - \\
& SSIM$\uparrow$  & - & - & - & - & - & - & - & - & - \\
& LPIPS$\downarrow$  & - & - & - & - & - & - & - & - & - \\
\midrule
\midrule
\multirow{3}{*}{Ours} 
& PSNR$\uparrow$  & \textbf{40.85} & \textbf{43.94} & \textbf{44.52} & \textbf{48.55} & \textbf{48.41} & \textbf{44.35} & \textbf{42.62} & \textbf{45.71} & \textbf{44.87} \\
& SSIM$\uparrow$  & \textbf{0.997} & \textbf{0.998} & \textbf{0.998} & \textbf{0.999} & \textbf{0.998}& \textbf{0.998} &\textbf{0.998} & \textbf{0.998} & \textbf{0.998} \\
& LPIPS$\downarrow$ & 0.020 & 0.020 & 0.022 & 0.018 & 0.024 & 0.017 & 0.022 & 0.022 & 0.021 \\

\bottomrule
\end{tabular}}
% \caption{Rendering performance comparisons in PSNR $\uparrow$, SSIM $\uparrow$, and LPIPS $\downarrow$ on Replica~\cite{replica}. $*$ indicates methods relying on pre-trained data-driven priors.}
\label{Tab:supprenderreplicaperscene}
\end{table*}

\begin{table*}
\caption{Reconstruction performance comparison in Depth L1 [cm]$\downarrow$ and F1 $[\mathrm{\%}] \uparrow$ on Replica~\cite{replica}. $*$ indicates methods relying on pre-trained data-driven priors.}
% \caption{Reconstruction performance comparison in Depth L1 [cm]$\downarrow$ and F1 $[\mathrm{\%}] \uparrow$ on Replica~\cite{replica}. $*$ indicates methods relying on pre-trained data-driven priors.}
\vskip 0.05in
  \centering
  \resizebox{\linewidth}{!}{
\begin{tabular}{llccccccccc}
\toprule
Method & Metric & \texttt{Rm0} & \texttt{Rm1} & \texttt{Rm2} & \texttt{Off0} & \texttt{Off1} & \texttt{Off2} & \texttt{Off3} & \texttt{Off4} & Avg. \\
\midrule
\rowcolor{gray!20}
\multicolumn{11}{l}{\textit{Neural Implicit Fields}} \\
\multirow{2}{*}{NICE-SLAM~\cite{Zhu2021NICESLAM}} 
& Depth L1 [cm]$\downarrow$ & 1.81 & 1.44 & 2.04 & 1.39 & 1.76 & 8.33 & 4.99 & 2.01 & 2.97 \\
& F1 $[\mathrm{\%}] \uparrow$ & 45.0 & 44.8 & 43.6 & 50.0 & 51.9 & 39.2 & 39.9 & 36.5 & 43.9 \\
\midrule
\multirow{2}{*}{Vox-Fusion~\cite{Yang_Li_Zhai_Ming_Liu_Zhang_2022_voxfusion}} 
& Depth L1 [cm]$\downarrow$ & 1.09 & 1.90 & 2.21 & 2.32 & 3.40 & 4.19 & 2.96 & 1.61 & 2.46 \\
& F1 $[\mathrm{\%}] \uparrow$ & 69.9 & 34.4 & 59.7 & 46.5 & 40.8 & 51.0 & 64.6 & 50.7 & 52.2 \\
\midrule
\multirow{2}{*}{ESLAM~\cite{johari-et-al-2023-ESLAM}} 
& Depth L1 [cm]$\downarrow$ & 0.97 & 1.07 & 1.28 & 0.86 & 1.26 & 1.71 & 1.43 & 1.06 & 1.18 \\
& F1 $[\mathrm{\%}] \uparrow$ & 81.0 & 82.2 & 83.9 & 78.4 & 75.5 & 77.1 & 75.5 & 79.1 & 79.1 \\
\midrule
\multirow{2}{*}{Co-SLAM~\cite{wang2023coslam}} 
& Depth L1 [cm]$\downarrow$ & 0.99 & 0.82 & 2.28 & 1.24 & 1.61 & 7.70 & 4.65 & 1.43 & 2.59\\
& F1 $[\mathrm{\%}] \uparrow$ & 77.7 & 74.2 & 69.3 & 75.2 & 75.2 & 54.3 & 56.8 & 75.3 & 69.7 \\
\midrule
\multirow{2}{*}{Point-SLAM~\cite{Sandström2023ICCVpointslam}} 
& Depth L1 [cm]$\downarrow$ & 0.53 & 0.22 & 0.46 & 0.30 & 0.57 & 0.49 & 0.51 & 0.46 & 0.44 \\
& F1 $[\mathrm{\%}] \uparrow$ & 86.9 & \textbf{92.3} & 90.8 & 93.8 & \textbf{91.6} & 89.0 & 88.2 & 85.6 & 89.8 \\
\hdashline
\multirow{2}{*}{Loopy-SLAM$*$~\cite{liso2024loopyslam}} 
& Depth L1 [cm]$\downarrow$ & 0.30 & 0.20 & 0.42 & 0.23 & 0.46 & 0.60 & 0.37 & 0.24 & 0.35 \\
& F1 $[\mathrm{\%}] \uparrow$ & 91.6 & 92.4 & 90.6 & 93.9 & 91.6 & 88.5 & 89.0 & 88.7 & 90.8 \\
\midrule
\midrule
\rowcolor{gray!20}
\multicolumn{11}{l}{\textit{3D Gaussian Splatting}} \\
\multirow{2}{*}{SplaTAM~\cite{keetha2024splatam}} 
& Depth L1 [cm]$\downarrow$ & 0.43 & 0.38 & 0.54 & 0.44 & 0.66 & 1.05 & 1.60 & 0.68 & 0.72 \\
& F1 $[\mathrm{\%}] \uparrow$ & 89.3 & 88.2 & 88.0 & 91.7 & 90.0 & 85.1 & 77.1 & 80.1 & 86.1 \\
\midrule
\multirow{2}{*}{GS-SLAM~\cite{yan2023gs}} 
& Depth L1 [cm]$\downarrow$ & 1.31 & 0.82 & 1.26 & 0.81 & 0.96 & 1.41 & 1.53 & 1.08 & 1.16 \\
& F1 $[\mathrm{\%}] \uparrow$ & 62.9 & 79.9 & 66.8 & 80.0 & 81.6 & 66.0 & 59.2 & 65.0 & 70.2 \\
\midrule
\multirow{2}{*}{Gaussian-SLAM~\cite{yugay2023gaussianslam}} 
& Depth L1 [cm]$\downarrow$ & 0.61 & 0.25 & 0.54 & 0.50 & 0.52 & 0.98 & 1.63 & 0.42 & 0.68 \\
& F1 $[\mathrm{\%}] \uparrow$ & 88.8 & 91.4 & 90.5 & 91.7 & 90.1 & 87.3 & 84.2 & 87.4 & 88.9 \\
\midrule
\multirow{2}{*}{VTGS-SLAM~\cite{Hu2025VTGSSLAM}} 
& Depth L1 [cm]$\downarrow$ & 0.48 & 0.28 & 0.61 & 0.41 & 0.48 & 0.62 & 0.86 & 0.53 & 0.53 \\
& F1 $[\mathrm{\%}] \uparrow$ & 90.7 & 91.7 & 90.7 & 93.0 & 90.8 & 88.3 & 87.5 & 87.0 & 90.0 \\
\hdashline
\multirow{2}{*}{LoopSplat$*$~\cite{zhu2024_loopsplat}} 
& Depth L1 [cm]$\downarrow$ & 0.39 & 0.23 & 0.52 & 0.32 & 0.51 & 0.63 & 1.09 & 0.40 & 0.51 \\
& F1 $[\mathrm{\%}] \uparrow$ & 90.6 & 91.9 & 91.1 & 93.3 & 90.4 & 88.9 & 88.7 & 88.3 & 90.4 \\
\midrule
\midrule
\multirow{2}{*}{Ours} 
& Depth L1 [cm]$\downarrow$ & \textbf{0.27} & \textbf{0.17} & \textbf{0.36} & \textbf{0.22} & \textbf{0.38} & \textbf{0.37} & \textbf{0.45} & \textbf{0.21} & \textbf{0.30}\\
& F1 $[\mathrm{\%}] \uparrow$ & \textbf{91.6} & \textbf{92.3} & \textbf{91.4} & \textbf{93.9} & 91.2 & \textbf{89.3} & \textbf{88.9} & \textbf{88.7} & \textbf{90.9}\\
\bottomrule

\end{tabular}}
% \caption{Reconstruction performance comparison in Depth L1 [cm]$\downarrow$ and F1 $[\mathrm{\%}] \uparrow$ on Replica~\cite{replica}. $*$ indicates methods relying on pre-trained data-driven priors.}
\label{Tab:suppreconreplicaperscene}
\end{table*}

\begin{table*}
\caption{Rendering performance comparison in PSNR $\uparrow$, SSIM $\uparrow$, and LPIPS $\downarrow$ on TUM-RGBD~\cite{6385773ATERMSE_tumrgbd}. $*$ indicates methods relying on pre-trained data-driven priors.}
\vskip 0.05in
  \centering
  \resizebox{0.8\linewidth}{!}{
\begin{tabular}{llcccc}
\toprule
Method & Metric & \texttt{fr1/desk} & \texttt{fr2/xyz} & \texttt{fr3/office} & Avg. \\
\midrule
\rowcolor{gray!20}
\multicolumn{6}{l}{\textit{Neural Implicit Fields}} \\
\multirow{3}{*}{NICE-SLAM~\cite{Zhu2021NICESLAM}} 
& PSNR$\uparrow$  & 13.83 & 17.87 & 12.89 & 14.86 \\
& SSIM$\uparrow$ & 0.569 & 0.718 & 0.554 & 0.614 \\
& LPIPS$\downarrow$ & 0.482 & 0.344 & 0.498 & 0.441 \\
\midrule
\multirow{3}{*}{Vox-Fusion~\cite{Yang_Li_Zhai_Ming_Liu_Zhang_2022_voxfusion}} 
& PSNR$\uparrow$ & 15.79 & 16.32 & 17.27 & 16.46 \\
& SSIM$\uparrow$ & 0.647 & 0.706 & 0.677 & 0.677 \\
& LPIPS$\downarrow$ & 0.523 & 0.433 & 0.456 & 0.471 \\
\midrule
\multirow{3}{*}{ESLAM~\cite{johari-et-al-2023-ESLAM}} 
& PSNR$\uparrow$ & 11.29 & 17.46 & 17.02 & 15.26 \\
& SSIM$\uparrow$ & 0.666 & 0.310 & 0.457 & 0.478 \\
& LPIPS$\downarrow$ & 0.358 & 0.698 & 0.652 & 0.569 \\
\midrule
\multirow{3}{*}{Point-SLAM~\cite{Sandström2023ICCVpointslam}} 
& PSNR$\uparrow$ & 13.87 & 17.56 & 18.43 & 16.62 \\
& SSIM$\uparrow$ & 0.627 & 0.708 & 0.754 & 0.696 \\
& LPIPS$\downarrow$ & 0.544 & 0.585 & 0.448 & 0.526 \\
\hdashline
\multirow{3}{*}{Loopy-SLAM$*$~\cite{liso2024loopyslam}} 
& PSNR$\uparrow$ & - & - & - & 12.94 \\
& SSIM$\uparrow$ & - & - & - & 0.489 \\
& LPIPS$\downarrow$ & - & - & - & 0.645 \\

\midrule
\midrule
\rowcolor{gray!20}
\multicolumn{6}{l}{\textit{3D Gaussian Splatting}} \\
\multirow{3}{*}{SplaTAM~\cite{keetha2024splatam}} 
& PSNR$\uparrow$ & 22.00 & 24.50 & 21.90 & 22.80 \\
& SSIM$\uparrow$ & 0.857 & 0.947 & 0.876 & 0.893 \\
& LPIPS$\downarrow$ & 0.232 & 0.100 & 0.202 & 0.178 \\
\midrule
\multirow{3}{*}{Gaussian-SLAM~\cite{yugay2023gaussianslam}} 
& PSNR$\uparrow$  & 24.01 & 25.02 & 26.13 & 25.05 \\
& SSIM$\uparrow$ & 0.924 & 0.924 & 0.939 & 0.929 \\
& LPIPS$\downarrow$  & 0.178 & 0.186 & 0.141 & 0.168 \\
\midrule
\multirow{3}{*}{VTGS-SLAM~\cite{Hu2025VTGSSLAM}} 
& PSNR$\uparrow$ & 27.09 & 33.01 & 30.50 & 30.20 \\
& SSIM$\uparrow$ & 0.959 & 0.982 & 0.974 & 0.972 \\
& LPIPS$\downarrow$ & 0.085 & 0.038 & 0.063 & 0.062 \\
\hdashline
\multirow{3}{*}{LoopSplat$*$~\cite{zhu2024_loopsplat}} 
& PSNR$\uparrow$ & 22.03 & 22.68 & 23.47 & 22.72 \\
& SSIM$\uparrow$ & 0.849 & 0.892 & 0.879 & 0.873 \\
& LPIPS$\downarrow$ & 0.307 & 0.217 & 0.253 & 0.259 \\
\midrule
\midrule
\multirow{3}{*}{Ours} 
& PSNR$\uparrow$ &  \textbf{39.21} & \textbf{36.74} & \textbf{39.86} & \textbf{38.60}\\
& SSIM$\uparrow$ & \textbf{0.998} & \textbf{0.996} & \textbf{0.997} & \textbf{0.997} \\
& LPIPS$\downarrow$ & \textbf{0.009} & \textbf{0.014} & \textbf{0.012} & \textbf{0.012} \\

\bottomrule
\end{tabular}}
% \caption{Rendering performance comparison in PSNR $\uparrow$, SSIM $\uparrow$, and LPIPS $\downarrow$ on TUM-RGBD~\cite{6385773ATERMSE_tumrgbd}. $*$ indicates methods relying on pre-trained data-driven priors.}
\label{Tab:supprendertumperscene}
\end{table*}

\begin{table*}
\caption{Rendering performance comparison in PSNR $\uparrow$, SSIM $\uparrow$, and LPIPS $\downarrow$ on ScanNet~\cite{scannet}. $*$ indicates methods relying on pre-trained data-driven priors.}
\vskip 0.05in
  \centering
  \resizebox{0.8\linewidth}{!}{
\begin{tabular}{llccccccc}
\toprule
Method & Metric & \texttt{0000} & \texttt{0059} & \texttt{0106} & \texttt{0169} & \texttt{0181} & \texttt{0207} & Avg. \\
\midrule
\rowcolor{gray!20}
\multicolumn{9}{l}{\textit{Neural Implicit Fields}} \\
\multirow{3}{*}{NICE-SLAM~\cite{Zhu2021NICESLAM}} 
& PSNR$\uparrow$ & 18.71 & 16.55 & 17.29 & 18.75 & 15.56 & 18.38 & 17.54 \\
& SSIM$\uparrow$ & 0.641 & 0.605 & 0.646 & 0.629 & 0.562 & 0.646 & 0.621 \\
& LPIPS$\downarrow$ & 0.561 & 0.534 & 0.510 & 0.534 & 0.602 & 0.552 & 0.548 \\
\midrule
\multirow{3}{*}{Vox-Fusion~\cite{Yang_Li_Zhai_Ming_Liu_Zhang_2022_voxfusion}} 
& PSNR$\uparrow$ & 19.06 & 16.38 & 18.46 & 18.69 & 16.75 & 19.66 & 18.17 \\
& SSIM$\uparrow$ & 0.662 & 0.615 & 0.753 & 0.650 & 0.666 & 0.696 & 0.673 \\
& LPIPS$\downarrow$ & 0.515 & 0.528 & 0.439 & 0.513 & 0.532 & 0.500 & 0.504 \\
\midrule
\multirow{3}{*}{ESLAM~\cite{johari-et-al-2023-ESLAM}} 
& PSNR$\uparrow$ & 15.70 & 14.48 & 15.44 & 14.56 & 14.22 & 17.32 & 15.29 \\
& SSIM$\uparrow$ & 0.687 & 0.632 & 0.628 & 0.656 & 0.696 & 0.653 & 0.658 \\
& LPIPS$\downarrow$ & 0.449 & 0.450 & 0.529 & 0.486 & 0.482 & 0.534 & 0.488 \\
\midrule
\multirow{3}{*}{Point-SLAM~\cite{Sandström2023ICCVpointslam}} 
& PSNR$\uparrow$ & 21.30 & 19.48 & 16.80 & 18.53 & 22.27 & 20.56 & 19.82 \\
& SSIM$\uparrow$ & 0.806 & 0.765 & 0.676 & 0.686 & 0.823 & 0.750 & 0.751 \\
& LPIPS$\downarrow$ & 0.485 & 0.499 & 0.544 & 0.542 & {0.471} & 0.544 & 0.514 \\
\hdashline
\multirow{3}{*}{LoopySLAM$*$~\cite{liso2024loopyslam}} 
& PSNR$\uparrow$ & - & - & - & - & - & - & 15.23 \\
& SSIM$\uparrow$ & - & - & - & - & - & - & 0.629 \\
& LPIPS$\downarrow$ & - & - & - & - & - & - & 0.671 \\
\midrule
\midrule
\rowcolor{gray!20}
\multicolumn{9}{l}{\textit{3D Gaussian Splatting}} \\
\multirow{3}{*}{SplaTAM~\cite{keetha2024splatam}} 
& PSNR$\uparrow$ & 19.33 & 19.27 & 17.73 & 21.97 & 16.76 & 19.8 & 19.14 \\
& SSIM$\uparrow$ & 0.660 & 0.792 & 0.690 & 0.776 & 0.683 & 0.696 & 0.716 \\
& LPIPS$\downarrow$ & 0.438 & 0.289 & 0.376 & 0.281 & 0.420 & 0.341 & 0.358 \\
\midrule
\multirow{3}{*}{Gaussian-SLAM~\cite{yugay2023gaussianslam}} 
% & PSNR & 28.539 & 26.208 & 26.258 & 28.604 & 27.789 & 28.627 & 27.67 \\
& PSNR$\uparrow$ & 28.54 & 26.21 & 26.26 & 28.60 & 27.79 & 28.63 & 27.70 \\
& SSIM$\uparrow$ & 0.926 & 0.934 & 0.926 & 0.917 & 0.922 & 0.914 & 0.923  \\
& LPIPS$\downarrow$ & 0.271 & 0.211 & 0.217 & 0.226 & 0.277 & 0.288 & 0.248 \\
\midrule
\multirow{3}{*}{VTGS-SLAM~\cite{Hu2025VTGSSLAM}} 
& PSNR$\uparrow$ & 31.51 & 30.60 & 31.27 & 32.02 & 29.60 & 31.58 &  31.10 \\
& SSIM$\uparrow$ & 0.957 & 0.974 & 0.975 & 0.962 & 0.954 & 0.946 & 0.961 \\
& LPIPS$\downarrow$ & 0.131 & 0.080 & 0.074 & 0.091 & 0.145 & 0.124 & 0.108 \\
\hdashline
\multirow{3}{*}{LoopSplat$*$~\cite{zhu2024_loopsplat}} 
& PSNR$\uparrow$ & 24.99 & 23.23 & 23.35 & 26.80 & 24.82 & 26.33 & 24.92 \\
& SSIM$\uparrow$ & 0.840 & 0.831 & 0.846 & 0.877 & 0.824 & 0.854 & 0.845 \\
& LPIPS$\downarrow$ & 0.450 & 0.400 & 0.409 & 0.346 & 0.514 & 0.430 & 0.425 \\
\midrule
\midrule
\multirow{3}{*}{Ours} 
& PSNR$\uparrow$ & \textbf{40.85} & \textbf{41.10} & \textbf{42.91} & \textbf{42.76} & \textbf{43.51} & \textbf{42.71} & \textbf{42.31} \\
& SSIM$\uparrow$ & \textbf{0.996} & \textbf{0.997} & \textbf{0.998} & \textbf{0.997} & \textbf{0.997} & \textbf{0.996} & \textbf{0.997}  \\
& LPIPS$\downarrow$ & \textbf{0.056} & \textbf{0.051} & \textbf{0.041} & \textbf{0.041} & \textbf{0.057} & \textbf{0.046} & \textbf{0.049} \\

\bottomrule
\end{tabular}}
% \caption{Rendering performance comparison in PSNR $\uparrow$, SSIM $\uparrow$, and LPIPS $\downarrow$ on ScanNet~\cite{scannet}. $*$ indicates methods relying on pre-trained data-driven priors.}
\label{Tab:supprenderscannetperscene}
\end{table*}

\begin{table*} %{r}{0.5\linewidth}
% \vspace{-0.23in}
\caption{Tracking performance comparisons in ATE RMSE $\downarrow [\mathrm{cm}]$ on Replica~\cite{replica}. $*$ methods relying on pre-trained data-driven priors.} %$\dagger$ method using only RGB as input.}
% \vskip 0.05in
\setlength{\tabcolsep}{2pt}
  \centering
  \resizebox{0.8\linewidth}{!}{
\begin{tabular}{lccccccccc}
\toprule
Method & \texttt{Rm0} & \texttt{Rm1} & \texttt{Rm2} & \texttt{Off0} & \texttt{Off1} & \texttt{Off2} & \texttt{Off3} & \texttt{Off4} & Avg. \\
\midrule
\rowcolor{gray!20}
\multicolumn{10}{l}{\textit{Neural Implicit Fields}} \\
NICE-SLAM~\cite{Zhu2021NICESLAM} & 1.69 & 2.04 & 1.55 & 0.99 & 0.90 & 1.39 & 3.97 & 3.08 & 1.95 \\
DF-Prior~\cite{Hu2023LNI-ADFP} & 1.39 & 1.55 & 2.60 & 1.09 & 1.23 & 1.61 & 3.61 & 1.42 & 1.81 \\
Vox-Fusion~\cite{Yang_Li_Zhai_Ming_Liu_Zhang_2022_voxfusion} & 0.27 & 1.33 & 0.47 & 0.70 & 1.11 & 0.46 & 0.26 & 0.58 & 0.65 \\
ESLAM~\cite{johari-et-al-2023-ESLAM} & 0.71 & 0.70 & 0.52 & 0.57 & 0.55 & 0.58 & 0.72 & 0.63 & 0.63 \\
Point-SLAM~\cite{Sandström2023ICCVpointslam} & 0.61 & 0.41 & 0.37 & 0.38 & 0.48 & 0.54 & 0.72 & 0.63 & 0.52 \\
\hdashline
Loopy-SLAM$*$~\cite{liso2024loopyslam} & 0.24 & 0.24 & 0.28 & 0.26 & 0.40 & 0.29 & 0.22 & 0.35 & 0.29 \\
\midrule
\midrule
\rowcolor{gray!20}
\multicolumn{10}{l}{\textit{3D Gaussian Splatting}} \\
SplaTAM~\cite{keetha2024splatam} & 0.31 & 0.40 & 0.29 & 0.47 & 0.27 & 0.29 & 0.32 & 0.55 & 0.36 \\

GS-SLAM~\cite{yan2023gs} & 0.48 & 0.53 & 0.33 & 0.52 & 0.41 & 0.59 & 0.46 & 0.70 & 0.50 \\
Gaussian-SLAM~\cite{yugay2023gaussianslam} & 0.29 & 0.29 & 0.22 & 0.37 & 0.23 & 0.41 & 0.30 & 0.35 & 0.31 \\
VTGS-SLAM~\cite{Hu2025VTGSSLAM} & 0.22 & 0.26 & 0.19 & 0.28 & 0.26 & 0.34 & 0.25 & 0.43 & 0.28 \\
GS-ICP SLAM~\cite{ha2024rgbdgsicpslam} & \textbf{0.15} & \textbf{0.16} & 0.11 & 0.18 & \textbf{0.12} & 0.17 & \textbf{0.16} & 0.21 & \textbf{0.16} \\
\hdashline
LoopSplat$*$~\cite{zhu2024_loopsplat} & 0.28 & 0.22 & 0.17 & 0.22 & 0.16 & 0.49 & 0.20 & 0.30 & 0.26 \\
CG-SLAM$*$~\cite{hu2024cg} & 0.29 & 0.27 & 0.25 & 0.33 & 0.14 & 0.28 & 0.31 & 0.29 & 0.27 \\
\midrule
\midrule
Ours & \textbf{0.15} & 0.17 & \textbf{0.10} & \textbf{0.16} & \textbf{0.12} & \textbf{0.16} & 0.25 & \textbf{0.20} & \textbf{0.16} \\

\bottomrule
\end{tabular}}
% \vspace{-0.1in}
% \caption{Tracking performance comparisons in ATE RMSE $\downarrow [\mathrm{cm}]$ on Replica~\cite{replica}. $*$ methods relying on pre-trained data-driven priors. $\dagger$ method using only RGB as input.}
\label{Tab:suppcamreplicaperscene}
% \vspace{-0.3in}
\end{table*}

\begin{table*} %{r}{0.5\linewidth}
% \vskip -0.25in
\caption{Tracking performance in ATE RMSE $\downarrow [\mathrm{cm}]$ on TUM-RGBD~\cite{6385773ATERMSE_tumrgbd}. $*$ methods using pre-trained data-driven priors.}
% \vskip 0.05in
  \centering
  \resizebox{0.8\linewidth}{!}{
\begin{tabular}{lcccc}
\toprule
Method & \texttt{fr1/desk} & \texttt{fr2/xyz} & \texttt{fr3/office} & Avg. \\
\midrule
\rowcolor{gray!20}
\multicolumn{5}{l}{\textit{Neural Implicit Fields}} \\
NICE-SLAM~\cite{Zhu2021NICESLAM} & 4.3 & 31.7 & 3.9 & 13.3 \\
Vox-Fusion~\cite{Yang_Li_Zhai_Ming_Liu_Zhang_2022_voxfusion} & 3.5 & 1.5 & 26.0 & 10.3 \\
Point-SLAM~\cite{Sandström2023ICCVpointslam} & 4.3 & 1.3 & 3.5 & 3.0 \\
\hdashline
Loopy-SLAM$*$~\cite{liso2024loopyslam} & 3.8 & 1.6 & 3.4 & 2.9 \\
\midrule
\midrule
\rowcolor{gray!20}
\multicolumn{5}{l}{\textit{3D Gaussian Splatting}} \\
SplaTAM~\cite{keetha2024splatam} & 3.4 & 1.2 & 5.2 & 3.3 \\ 
GS-SLAM~\cite{yan2023gs} & 3.3 & 1.3 & 6.6 & 3.7 \\
Gaussian-SLAM~\cite{yugay2023gaussianslam} & 2.6 & 1.3 & 4.6 & 2.9 \\
VTGS-SLAM~\cite{Hu2025VTGSSLAM} & 2.4 & \textbf{1.1} & 4.4 & 2.6 \\
GS-ICP SLAM~\cite{ha2024rgbdgsicpslam} & 2.7 & 1.8 & 2.7 & 2.4 \\
\hdashline
LoopSplat$*$~\cite{zhu2024_loopsplat} & 2.1 & 1.6 & 3.2 & 2.3 \\
CG-SLAM$*$~\cite{hu2024cg} & 2.4 & 1.2 & 2.5 & 2.0 \\
\midrule
\midrule
Ours & \textbf{2.2} & 1.7 & \textbf{2.0} & \textbf{2.0} \\ % rendering init
\bottomrule
\end{tabular}}
% \vspace{-0.1in}
% \caption{Tracking performance in ATE RMSE $\downarrow [\mathrm{cm}]$ on TUM-RGBD~\cite{6385773ATERMSE_tumrgbd}. $*$ methods using pre-trained data-driven priors.}
% \vspace{-0.15in}
\label{Tab:suppcamtumperscene}
\end{table*}

\begin{table*}
\caption{Tracking performance in ATE RMSE $\downarrow [\mathrm{cm}]$ on TUM-RGBD~\cite{6385773ATERMSE_tumrgbd}. $*$ indicates VTGS-SLAM~\cite{Hu2025VTGSSLAM} treats each frame as a section, initializing dense Gaussians on each frame similar to our approach.}
% } \renewcommand{\arraystretch}{0.9}
% \vspace{-0.15in}
% \setlength{\tabcolsep}{0.65mm}
    % \renewcommand{\arraystretch}{1.2}
	\centering
    % \small
	\resizebox{0.8\linewidth}{!}{
   \begin{tabular}{lcccc}
    \toprule
    % \hline
    Method & \texttt{fr1/desk} & \texttt{fr2/xyz} & \texttt{fr3/office} & Avg. \\
\midrule
% \hline
   VTGS-SLAM~\cite{Hu2025VTGSSLAM} & 382.4 & 3462.5 & 400.5 & 1415.1  \\
   Ours & \textbf{2.2} & \textbf{1.7} & \textbf{2.0} & \textbf{2.0} \\
    \bottomrule
    % \hline
    \end{tabular}
    }
\label{tab:tracking_tum_vtgs}
% \vspace{-0.175in}
\end{table*}

\begin{table*}
% \vspace{-0.4in}
\caption{Tracking performance in ATE RMSE $\downarrow [\mathrm{cm}]$ on \texttt{fr3/nostructure\_texture\_far} in TUM-RGBD~\cite{6385773ATERMSE_tumrgbd}. $*$ methods using pre-trained data-driven priors.} 
% \renewcommand{\arraystretch}{0.9}
% \vspace{-0.15in}
% \setlength{\tabcolsep}{0.65mm}
    % \renewcommand{\arraystretch}{1.2}
	\centering
    % \small
	\resizebox{\linewidth}{!}{
   \begin{tabular}{lcccccc}
    \toprule
    % \hline
    Method & SplaTAM~\cite{keetha2024splatam} & LoopSplat$*$~\cite{zhu2024_loopsplat} & VTGS-SLAM~\cite{Hu2025VTGSSLAM} & GS-ICP SLAM~\cite{ha2024rgbdgsicpslam} & Ours(w/o init.) & Ours(w/ init.) \\
\midrule
% \hline
  ATE RMSE$\downarrow$$ [\mathrm{cm}]$ & 11.3 & 7.3 & 9.7 & 115.8 & 122.4  & \textbf{4.7} \\
    \bottomrule
    % \hline
    \end{tabular}
    }
\label{tab:tracking_fr3}
% \vspace{-0.35in}
\end{table*}

\begin{table*}
\caption{Tracking performance in ATE RMSE $\downarrow [\mathrm{cm}]$ on ScanNet~\cite{scannet}. $*$ methods using pre-trained data-driven priors.}
\centering
    \setlength{\tabcolsep}{2pt}
  \centering
  \resizebox{0.8\linewidth}{!}{
\begin{tabular}{lccccccc}
\toprule
Method & \texttt{0000} & \texttt{0059} & \texttt{0106} & \texttt{0169} & \texttt{0181} & \texttt{0207} & Avg. \\
\midrule
\rowcolor{gray!20}
\multicolumn{8}{l}{\textit{Neural Implicit Fields}} \\
NICE-SLAM~\cite{Zhu2021NICESLAM} & 12.0 & 14.0 & 7.9 & 10.9 & 13.4 & 6.2 & 10.7 \\
Vox-Fusion~\cite{Yang_Li_Zhai_Ming_Liu_Zhang_2022_voxfusion} & 68.8 & 24.2 & 8.4 & 27.3 & 23.3 & 9.4 & 26.9 \\
Point-SLAM~\cite{Sandström2023ICCVpointslam} & \textbf{10.2} & 7.8 & 8.7 & 22.2 & 14.8 & 9.5 & 12.2 \\
\hdashline
Loopy-SLAM$*$~\cite{liso2024loopyslam} & 4.2 & 7.5 & 8.3 & 7.5 & 10.6 & 7.9 & 7.7 \\
\midrule
\midrule
\rowcolor{gray!20}
\multicolumn{8}{l}{\textit{3D Gaussian Splatting}} \\
SplaTAM~\cite{keetha2024splatam} & 12.8 & 10.1 & 17.7 & 12.1 & 11.1 & 7.5 & 11.9  \\ 
Gaussian-SLAM~\cite{yugay2023gaussianslam} & 24.8 & 8.6 & 11.3 & 14.6 & 18.7 & 14.4 & 15.4 \\
% GS-ICP SLAM~\cite{ha2024rgbdgsicpslam}  &  &  &  &  &  &  &  \\
VTGS-SLAM~\cite{Hu2025VTGSSLAM} & 17.8 & 8.7 & 11.8 & 10.5 & 10.6 & 8.6 & 11.3\\
\hdashline
LoopSplat$*$~\cite{zhu2024_loopsplat} & 6.2 & 7.1 & 7.4 & 10.6 & 8.5 & 6.6 & 7.7 \\
CG-SLAM$*$~\cite{hu2024cg} & 7.1 & 7.5 & 8.9 & 8.2 & 11.6 & 5.3 & 8.1 \\
\midrule
\midrule
Ours & 11.9 & \textbf{6.4} & \textbf{5.3} & \textbf{8.5} & \textbf{10.3} & \textbf{4.7} & \textbf{7.9} \\

\bottomrule
\end{tabular}}
% \vspace{-0.2in}
% \caption{Tracking performance in ATE RMSE $\downarrow [\mathrm{cm}]$ on ScanNet~\cite{scannet}. $*$ methods using pre-trained data-driven priors.}
\label{Tab:suppcamscannetperscene}
\end{table*}

\begin{table*}
\caption{Tracking performance in ATE RMSE $\downarrow [\mathrm{cm}]$ on ScanNet++~\cite{yeshwanthliu2023scannetpp}. $*$ methods relying on pre-trained data-driven priors.}
    \centering
    \setlength{\tabcolsep}{2pt}
  \centering
  \resizebox{0.8\linewidth}{!}{
\begin{tabular}{lcccccc}
\toprule
Method & \texttt{a} & \texttt{b} & \texttt{c} & \texttt{d} & \texttt{e} & Avg. \\
\midrule
\rowcolor{gray!20}
\multicolumn{7}{l}{\textit{Neural Implicit Fields}} \\
Point-SLAM~\cite{Sandström2023ICCVpointslam} & 246.16 & 632.99 & 830.79 & 271.42 & 574.86 & 511.24 \\
ESLAM~\cite{johari-et-al-2023-ESLAM} & 25.15 & 2.15 & 27.02 & 20.89 & 35.47 & 22.14 \\
\hdashline
Loopy-SLAM$*$~\cite{liso2024loopyslam} & - & - & 25.16 & 234.25 & 81.48 & 113.63 \\
\midrule
\midrule
\rowcolor{gray!20}
\multicolumn{7}{l}{\textit{3D Gaussian Splatting}} \\
SplaTAM~\cite{keetha2024splatam} & 1.50 & \textbf{0.57} & 0.31 & 443.10 & 1.58 & 89.41  \\ 
Gaussian-SLAM~\cite{yugay2023gaussianslam} & 1.37 & 5.97 & 2.70 & 2.35 & 1.02 & 2.68 \\
VTGS-SLAM~\cite{Hu2025VTGSSLAM} & 2.80 & 1.50 & 1.00 & 1.20 & 1.30 & 1.60 \\
\hdashline
LoopSplat$*$~\cite{zhu2024_loopsplat} & 1.14 & 3.16 & 3.16 & 1.68 & 0.91 & 2.05 \\
\midrule
\midrule
Ours(w/o Initialization) & 5.57 & 16.70 & 1.70 & 4.50 & 4.20 & 6.50 \\
Ours & \textbf{0.80} & 0.71 & \textbf{0.05} & \textbf{0.63} & \textbf{0.74} & \textbf{0.59} \\

\bottomrule
\end{tabular}}
% \vspace{-0.2in}
% \caption{Tracking performance in ATE RMSE $\downarrow [\mathrm{cm}]$ on ScanNet++~\cite{yeshwanthliu2023scannetpp}. $*$ methods relying on pre-trained data-driven priors.}
% \vspace{-0.1in}
\label{Tab:suppcamscannetppperscene}
\end{table*}

\begin{table*}
\caption{Rendering performance comparison in PSNR $\uparrow$ on ScanNet++~\cite{yeshwanthliu2023scannetpp}. $*$ indicates methods relying on pre-trained data-driven priors.}
\vskip 0.05in
  \centering
  \resizebox{0.8\linewidth}{!}{
\begin{tabular}{lcccccc}
\toprule
Method & \texttt{a} & \texttt{b} & \texttt{c} & \texttt{d} & \texttt{e} & Avg. \\
\midrule
\rowcolor{gray!20}
\multicolumn{7}{l}{\textit{3D Gaussian Splatting}} \\
SplaTAM~\cite{keetha2024splatam} & 28.02 & 27.93 & 29.48 & 19.65 & 28.48 & 26.71 \\
Gaussian-SLAM~\cite{yugay2023gaussianslam} & 30.06 & 30.02 & 31.15 & 28.75 & 31.94 & 30.38  \\
VTGS-SLAM~\cite{Hu2025VTGSSLAM} & 32.84 & 31.02 & 32.44 & 31.43 & 33.38 & 32.22 \\
\hdashline
LoopSplat$*$~\cite{zhu2024_loopsplat} & 30.15 & 30.08 & 30.04 & 28.94 & 31.78 & 30.20 \\
\midrule
Ours & \textbf{35.95} & \textbf{34.84} & \textbf{35.81} & \textbf{35.71} & \textbf{41.32} & \textbf{36.73} \\

\bottomrule
\end{tabular}}
% \caption{Rendering performance comparison in PSNR $\uparrow$ on ScanNet++~\cite{yeshwanthliu2023scannetpp}. $*$ indicates methods relying on pre-trained data-driven priors.}
\label{Tab:supprenderscannetppperscene}
\end{table*}

\begin{table*}
\caption{Novel View Synthesis performance comparison in PSNR $\uparrow$ on ScanNet++~\cite{yeshwanthliu2023scannetpp}. $*$ indicates methods relying on pre-trained data-driven priors. We calculate PSNR including all pixels, regardless of whether they have a valid depth input.}
\vskip 0.05in
  \centering
  \resizebox{0.8\linewidth}{!}{
\begin{tabular}{lcccccc}
\toprule
Method & \texttt{a} & \texttt{b} & \texttt{c} & \texttt{d} & \texttt{e} & Avg. \\
\midrule
\rowcolor{gray!20}
\multicolumn{7}{l}{\textit{3D Gaussian Splatting}} \\
SplaTAM~\cite{keetha2024splatam} & 23.95 & 22.66 & 13.95 & 8.47 & 20.06 & 17.82 \\
Gaussian-SLAM~\cite{yugay2023gaussianslam} & 26.66 & 24.42 & 15.01 & 18.35 & 21.91 & 21.27 \\
VTGS-SLAM~\cite{Hu2025VTGSSLAM}  & 25.55 & 24.25 & \textbf{16.94} & 18.59 & \textbf{21.95} & 21.46 \\
\hdashline
LoopSplat$*$~\cite{zhu2024_loopsplat} & 25.60 & 23.65 & 15.87 & 18.86 & 22.51 & 21.30 \\
\midrule
Ours & \textbf{26.81} & \textbf{26.79} & 15.38 & \textbf{20.89} & 21.71 & \textbf{22.32} \\

\bottomrule
\end{tabular}}
% \caption{Novel View Synthesis performance comparison in PSNR $\uparrow$ on ScanNet++~\cite{yeshwanthliu2023scannetpp}. $*$ indicates methods relying on pre-trained data-driven priors. We calculate PSNR including all pixels, regardless of whether they have a valid depth input.}
\label{Tab:suppnvsrenderscannetppperscene}
\end{table*}


\end{document}